\documentclass{article}

\def\fighome{./}

\usepackage{ijcai19}

\usepackage{url}
\usepackage[hidelinks]{hyperref}
\usepackage{cleveref}
\usepackage{times}
\usepackage[square,sort,comma,numbers]{natbib}
\usepackage[outdir=./]{epstopdf}
\usepackage{graphicx,float,pgfplots,wrapfig,sidecap,lipsum}
\usepackage{tabularx}
\usepackage{booktabs}
\usepackage{paralist}
\usepackage{algorithm}
\usepackage[noend]{algorithmic}
\usepackage{amsfonts}
\usepackage{amsthm}
\usepackage{amsmath}
\usepackage{amssymb}
\usepackage{enumitem}
\usepackage{xcolor}
\usepackage[small]{caption}
\usepackage{mathtools} 
\usepackage{mathrsfs} 

\usepackage{tablefootnote}
\usepackage{subcaption}
\usepackage{psfrag}
\usepackage{tikz}
\usetikzlibrary{fit}
\usetikzlibrary{calc,shapes}
\usetikzlibrary{decorations.pathmorphing} 
\usetikzlibrary{fit}					
\usetikzlibrary{backgrounds}	

\usepackage{multirow}
\usepackage{xspace}
\usepackage{stmaryrd}
\usepackage[utf8]{inputenc}
\usepackage[english]{babel}
\usepackage{soul}
\usepackage{bm}

\newcommand*{\mymatrix}[1]{\mathbf{#1}}
\newcommand*{\myvector}[1]{\mathbf{#1}}

\newcommand{\matrixSub}[2]{\mathbf{#1}_{#2}}	
\newcommand{\matrixSup}[2]{\mathbf{#1}^{(#2)}}

\newcommand{\tensor}[1]{\mathcal{#1}}
\newcommand{\tensorSub}[2]{\mathcal{#1}_{#2}}	
\newcommand{\tensorSup}[2]{\mathcal{#1}^{(#2)}}
\newcommand{\tensorInd}[3]{\mathcal{#1}_{#2}^{(#3)}}

\newcommand{\swapaxes}{\mathsf{swapaxes}}
\newcommand{\flipaxis}{\mathsf{flipaxis}}

\newcommand{\relu}{\mathsf{ReLU}}

\newcommand\R{\mathbb{R}}

\newcommand\vectorize{\mathsf{vec}}

\def\tha{{\mbox{\tiny th}}}
\def\beq{\begin{equation}}

\def\eeq{\end{equation}\noindent}

\newcommand{\bp}{\begin{psfrags}}
\newcommand{\ep}{\end{psfrags}}
\newcommand{\bc}{\begin{center}}
\newcommand{\ec}{\end{center}}

\floatname{algorithm}{Procedure}

\newcommand{\Resnet}{ResNet\xspace}

\newcommand{\tensornet}{tensorial neural networks\xspace}

\newcommand{\tensornetshort}{TNN\xspace}

\newcommand{\convnet}{convolutional neural networks\xspace}

\newcommand{\convnetshort}{CNN\xspace}

\newcommand{\neuralnet}{neural networks\xspace}

\newcommand{\neuralnetshort}{NN\xspace}

\newcommand{\etoe}{E2E\xspace}
\newcommand{\etoeLong}{end-to-end\xspace}

\newcommand{\ourbaseline}{low-rank approximation based compression\xspace}

\newcommand{\ourbaselineshort}{NN-C\xspace}

\newcommand{\SeqTrain}{Sequential\xspace}
\newcommand{\seqTrain}{sequential\xspace}
\newcommand{\seqTune}{Seq\xspace}

\newcommand{\ouralgo}{TNN based compression\xspace}

\newcommand{\ouralgoshort}{TNN-C\xspace}

\newcommand{\SVD}{SVD\xspace}
\newcommand{\CP}{CP\xspace}
\newcommand{\TK}{TK\xspace}
\newcommand{\TT}{TT\xspace}

\newcommand{\CPlong}{CANDECOMP/PARAFAC\xspace}
\newcommand{\TTlong}{Tensor-train\xspace}
\newcommand{\TKlong}{Tucker\xspace}

\newcommand{\nnPrefix}{NN-}
\newcommand{\nSVD}{\nnPrefix\SVD} 
\newcommand{\nCP}{\nnPrefix\CP}
\newcommand{\nTK}{\nnPrefix\TK}
\newcommand{\nTT}{\nnPrefix\TT}

\newcommand{\reshapePrefix}{m}
 
\newcommand{\rCP}{\reshapePrefix\CP}
\newcommand{\rTK}{\reshapePrefix\TK}
\newcommand{\rTT}{\reshapePrefix\TT}

\newcommand{\tnnPrefix}{TNN-}
 
\newcommand{\tnnCP}{\tnnPrefix\rCP}
\newcommand{\tnnTK}{\tnnPrefix\rTK}
\newcommand{\tnnTT}{\tnnPrefix\rTT}

\newcommand{\mytitle}
{Tensorial Neural Networks: Generalization of Neural Networks \\ 
and Application to Model Compression}

\title{\mytitle}

\author{
Jiahao Su$^1$
\and
Jingling Li$^{2}$\and
Bobby Bhattacharjee$^{2}$\and
Furong Huang$^{2}$
\affiliations
$^1$Department of Electrical and Computer Engineering, University of Maryland, College Park \\
$^2$Department of Computer Science, University of Maryland, College Park
\emails
jiahaosu@terpmail.umd.edu, 
\{jingling, bobby, furongh\}@cs.umd.edu
}

\begin{document}
\maketitle

\begin{abstract}
We propose \textit{\tensornet ({\tensornetshort}s)}, a generalization of existing neural networks by extending tensor operations on low order operands to those on high order ones. 
The problem of parameter learning is challenging, as it corresponds to hierarchical nonlinear tensor decomposition.
We propose to solve the learning problem using stochastic gradient descent by deriving nontrivial backpropagation rules in generalized tensor algebra we introduce.
 Our proposed {\tensornetshort}s has three advantages over existing neural networks: 
 (1) {\tensornetshort}s naturally apply to high order input object and thus preserve the multi-dimensional structure in the input, as there is no need to flatten the data. 
 (2) {\tensornetshort}s interpret designs of existing neural network architectures. 
 (3) Mapping a neural network to {\tensornetshort}s with the same expressive power results in a {\tensornetshort} of fewer parameters. 
 \ouralgo of neural network improves existing \ourbaseline methods as {\tensornetshort}s exploit two other types of {\em invariant structures}, periodicity and modulation, in addition to the low rankness.
Experiments on LeNet-5 (MNIST), ResNet-32 (CIFAR10) and ResNet-50 (ImageNet) demonstrate that our \ouralgo outperforms (5\% test accuracy improvement universally on CIFAR10) the state-of-the-art \ourbaseline methods under the same compression rate, besides achieving orders of magnitude faster convergence rates due to the efficiency of {\tensornetshort}s. 
\end{abstract}

\section{Introduction}
\label{sec:introduction}

Modern neural networks~\cite{krizhevsky2012imagenet, simonyan2014very, szegedy2015going, he2016deep, szegedy2017inception, huang2017densely} achieve unprecedented accuracy over many difficult learning problems at the cost of deeper and wider architectures with overwhelming number of model parameters. 
The large number of model parameters causes repeated high-cost in predictions, which becomes a practical bottleneck when these neural networks are deployed on constrained devices, such as smartphones and Internet of Things (IoT) devices.

One fundamental problem in deep learning research is to design neural network models with compact architectures, but still maintain comparable expressive power as large models.
To achieve this goal, two complementary approaches are adopted in the community: one approach is to compress well-trained neural networks while preserving their predictive performance as much as possible~\cite{cheng2017survey}.
Another approach is to find better architectures for neural network designs such as grouping the filters into inception modules~\cite{szegedy2015going, szegedy2017inception} or bottleneck layers~\cite{lin2013network, he2016identity}. 

To address this aforementioned fundamental problem, we propose \textit{\tensornet} ({\tensornetshort}s), which allows not only compression of well-trained networks, but also exploration of better designs in network architectures.
\tensornetshort is a generalization of existing \neuralnet ({\neuralnetshort}s) where matrix-vector multiplication (in fully connected layer and recurrent layers) and convolution (in convolutional layer) are extended to \textit{generalized tensor operations}. To achieve this, we introduce new tensor algebra to extend existing operations with low order operands to those with high order operands (see Section~\ref{sec:preliminary} and Appendix~\ref{app:operations} for details).

\begin{figure}[!htbp]
	\centering
	\psfrag{c1}[][][0.8]{$\bm{\mathcal{G}}^{p}$}
	\psfrag{c2}[][][0.8]{$\bm{\mathcal{H}}^{p}$}
	\psfrag{c3}[][][0.8]{$\bm{\mathcal{G}}^{q}$}
	\psfrag{c4}[][][0.8]{$\bm{\mathcal{H}}^{q}$}
	\psfrag{f}[][][0.8]{$f$}
	\psfrag{g1}[][][0.8]{$g^{q}$}
	\psfrag{g2}[][][0.8]{$g^p$}
	\psfrag{h1}[][][0.8]{$h^{q}$}
	\psfrag{h2}[][][0.8]{$h^p$}
	\psfrag{c1:compressedNN}[][][0.8]{$\bm{\mathcal{G}}^{p}$:compressed NN}
	\psfrag{c2:compressedTNN}[][][0.8]{$\bm{\mathcal{H}}^{p}$:compressed TNN}
	\psfrag{c3:NN}[][][0.8]{$\bm{\mathcal{G}}^{q}$: NN}
	\psfrag{c4:TNN}[][][0.8]{$\bm{\mathcal{H}}^{q}$: TNN}
	\includegraphics[width=0.4\textwidth]{\fighome/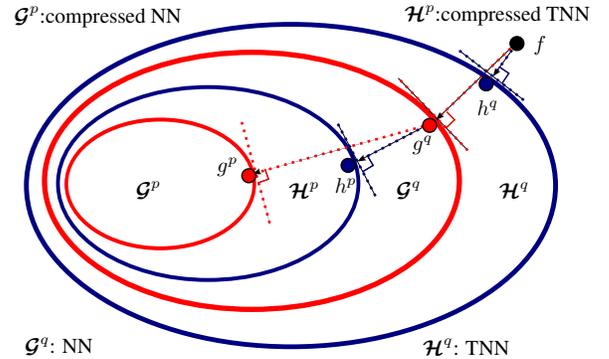}
\caption{Relationship between {\neuralnetshort}s and {\tensornetshort}s. In the figure, $f$ is the target concept. (1) \textbf{Learning} of a \neuralnetshort with $q$ parameters results in $g^{q}$ that is closest to $f$ in $\mathcal{G}^{q}$, while learning of a \tensornetshort with $q$ number of parameters results in $h^{q}$ that is closest to $f$ in $\mathcal{H}^{q}$. Apparently, $h^{q}$ is closer to $f$ than $g^{q}$.  (2) \textbf{Compression} of a pre-trained {\neuralnetshort} $g^{q}$ to {\neuralnetshort}s with $p$ parameters ($p \le q$) results in $g^{p}$ that is closest to $g^{q}$ in $\mathcal{G}^p$, while compressing of $g^{q}$ to {\tensornetshort}s with $p$ parameters results in $h^p$ that is closest to $g^{q}$ in $\mathcal{H}^p$.  Apparently, the compressed {\tensornetshort} $h^{p}$ is closer to the pre-trained {} $g^{q}$ than the compressed NN $g^{p}$. }
\label{fig:framework}
\end{figure}

Figure~\ref{fig:framework} illustrates the relationship between existing \neuralnet and our proposed \tensornet. 
Let $\mathcal{G}^{q}$ and $\mathcal{H}^{q}$ denote the sets of functions that can be represented by existing {\neuralnetshort}s and our {\tensornetshort}s, both with at most $q$ parameters.
Since existing {\neuralnetshort}s are special cases of {\tensornetshort}s, we have the following properties: 
(1) for any $q > 0$, $\mathcal{G}^{q} \subseteq \mathcal{H}^{q}$ and 
(2) there exists $p \leq q$ such that $ \mathcal{H}^{p} \subseteq \mathcal{G}^{q}$. 
The first property indicates that {\tensornetshort}s are generalization of {\neuralnetshort}s while the second property guarantees that {\tensornetshort}s can be used for compression of {\neuralnetshort}s.

The input to a {\tensornetshort} is a tensor of any order $m$, and the \tensornetshort reduces to a normal {\neuralnetshort} if the input is a vector ($m = 1$) or a matrix ($m = 2$).
\textbf{Prediction} in {\tensornetshort}s is similar to traditional {\neuralnetshort}s: the input is passed through the layers of a \tensornetshort in a feedforward manner, where each layer is a generalized tensor (multilinear) operation between the {\em high-order} input and the {\em high-order} weight kernels followed by a nonlinear activation function such as $\relu$.
\textbf{Learning} parameters in a {\tensornetshort} is equivalent to hierarchical nonlinear tensor decomposition, which is hard for arbitrary \tensornetshort architectures. We introduce a suite of generalized tensor algebra, which allows easy derivation of backpropagation rules and a class of TNN architectures (detailed in Appendices~\ref{app:derivatives}, \ref{app:dense-tensorized} and \ref{app:convolutional-tensorized}). With these backpropagation rules, {\tensornetshort}s can be efficiently learned by standard {\em stochastic gradient descent}. 

{\tensornetshort}s can be used for compression of traditional \neuralnet, since our proposed {\tensornetshort}s naturally identify \emph{invariant structures} in \neuralnet (justified in section~\ref{sec:invariant}).
Given a pre-trained {\neuralnetshort} $g^{q} \in \mathcal{G}^{q}$, compressing it to a {\tensornetshort} with $p$ parameters results in $h^{p}$ that is closest to $g^{q}$ in $\mathcal{H}^p$ as depicted in Figure~\ref{fig:framework}. It proceeds in two steps: (1) \textbf{data tensorization}: the input is reshaped into an $m$-order tensor; and (2) \textbf{knowledge distillation}: mapping from a \neuralnetshort to a \tensornetshort using layer-wise data reconstruction. 

We demonstrate the effectiveness of compression using \tensornet by conducting a set of experiments on several benchmark computer vision datasets.
 We compress \Resnet-32 on the CIFAR10 dataset by 10x with a degrading of only 1.92\% (achieving an accuracy of 91.28\%). 
Experiments on LeNet-5 (MNIST), ResNet-32 (CIFAR10) and ResNet-50 (ImageNet) demonstrate that our \tensornetshort compression outperforms (5\% test accuracy improvement universally on CIFAR10) the state-of-the-art low-rank approximation techniques under same compression rate, besides achieving orders of magnitude faster convergence rates due to the efficiency of {\tensornetshort}s. 

\paragraph{Contributions of this paper}
\begin{compactenum}
\item We propose a new framework of {\tensornet} that extends traditional {\neuralnet}, which naturally preserve multi-dimensional structures of the input data (such as videos).
\item We introduce a system of {\em generalized tensor algebra} for efficient learning and prediction in {\tensornetshort}s. In particular, we are the first to derive and analyze backpropagation rules for generalized tensor operations.  
\item We apply {\tensornet} to effectively compress existing neural networks by exploiting additional invariant structures in both data and parameter spaces, therefore reduce the complexity of the model (the number of parameters).
\item We provide interpretations of famous neural network architectures in computer vision using our proposed {\tensornetshort}s. Understanding of why some existing neural network architectures are successful in practice could lead to more insightful neural network designs.
\end{compactenum}

\subsection{Related Works}
\label{sec:related}
\paragraph{Relation to Tensor Networks. }
Tensor networks are widely used in quantum physics~\cite{orus2014practical},  numerical analysis~\cite{grasedyck2013literature} and recently machine learning~\cite{cichocki2016low, cichocki2017tensor}. Different from existing tensor networks, (1) our {\tensornetshort}s have nonlinearity on the edges between the tensors; and (2) {\tensornetshort}s are constructed as deep compositions of interleaving {\em generalized tensor operations} and nonlinear transformations, similar to feedforward neural networks. Furthermore, while tensor networks are still multi-linear and thus can be learned by algorithms such as power iteration~\cite{wang2017tensor} or alternative least squares~\cite{comon2009tensor}, {\tensornetshort}s require decomposition of nonlinear (no longer multi-linear) hierarchical tensor networks.

\paragraph{Compression of Neural Networks. }
A recent survey~\cite{cheng2017survey} reviews state-of-the-art techniques for compressing neural networks. These techniques can be grouped into two categories: (1) compressing an existing model and (2) novel compact designs. The first category includes \textit{low-rank approximations}~\cite{jaderberg2014speeding, denton2014exploiting, lebedev2014speeding, kim2015compression}, \textit{knowledge distillation}~\cite{romero2014fitnets, ba2014deep, hinton2015distilling, furlanello2018born} and \textit{quantization}~\cite{han2015deep, courbariaux2015binaryconnect, zhu2016trained, rastegari2016xnor, hubara2017quantized}; while the second category includes \textit{compact designs of filters}~\cite{cheng2015exploration, yang2015deep, sindhwani2015structured} and \textit{compact designs of architectures}~\cite{szegedy2015going, he2016deep, huang2017densely, chollet2016xception, szegedy2017inception}. When our {\tensornetshort}s are used for compression, we project an existing neural network to the class of {\tensornetshort}s with fewer parameters, and in Section~\ref{sec:interpretation}, we demonstrate this projection naturally corresponds novel compact architecture.

\section{Generalized Tensor Algebra}
\label{sec:preliminary}

\begin{figure}[!htbp]
\begin{minipage}{0.45\textwidth}
	\begin{subfigure}[b]{0.2\textwidth}
	\centering
	\psfrag{n1}{$a$}
	\includegraphics[width=0.4\textwidth]{\fighome/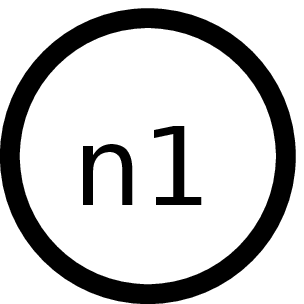}
	\caption{Scalar}
	\label{fig:diagram-scalar}
	\end{subfigure}
	\hfill
	\begin{subfigure}[b]{0.24\textwidth}
	\centering
	\psfrag{n1}{$\myvector{v}$}
	\psfrag{S1}{$I$}
	\includegraphics[width=0.66\textwidth]{\fighome/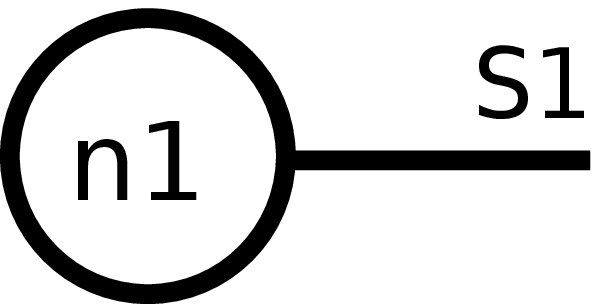}
	\caption{Vector}
	\label{fig:diagram-vector}
	\end{subfigure}
	\hfill
	\begin{subfigure}[b]{0.24\textwidth}
	\centering
	\psfrag{n1}{$\mymatrix{M}$}
	\psfrag{S1}{$I$}
	\psfrag{S2}{$J$}
	\includegraphics[width=\textwidth]{\fighome/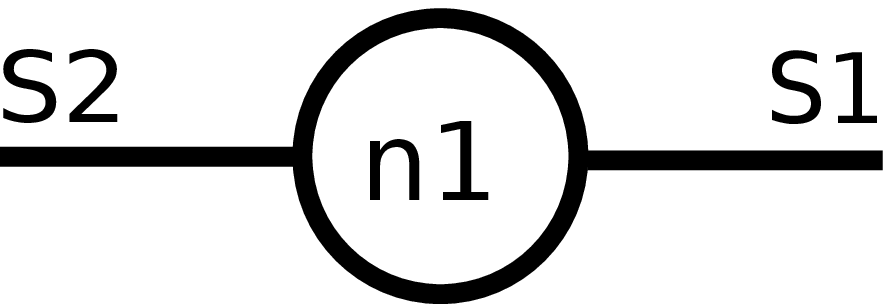}
	\caption{Matrix}
	\label{fig:diagram-matrix}
	\end{subfigure}
	\hfill
	\begin{subfigure}[b]{0.24\textwidth}
	\centering
	\psfrag{n1}{$\tensor{T}$}
	\psfrag{S1}{$I$}
	\psfrag{S2}{$J$}
	\psfrag{S3}{$K$}
	\includegraphics[width=\textwidth]{\fighome/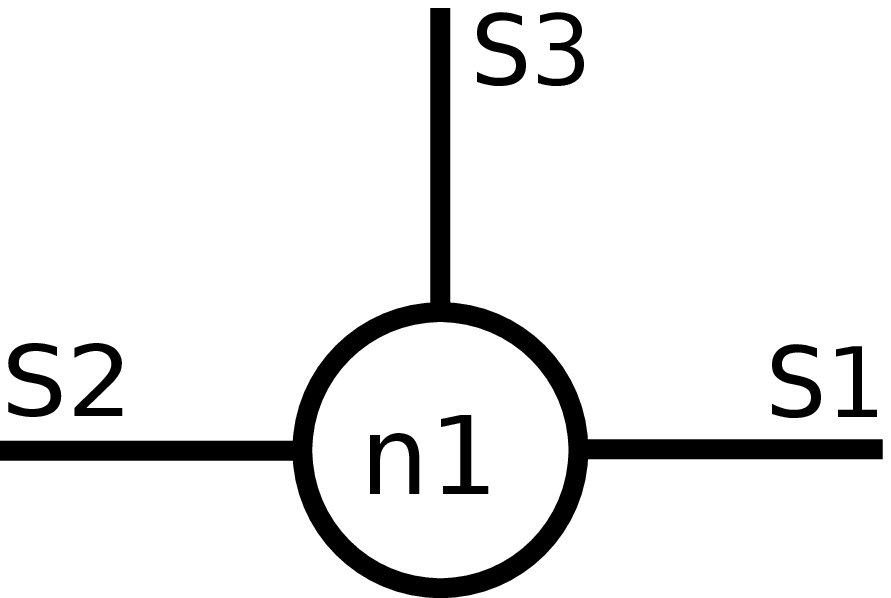}
	\caption{Tensor}
	\label{fig:diagram-tensor}
	\end{subfigure}
\caption{\small{\textbf{Tensor Diagrams} of 
a scalar $a\! \in\! \R$,
a vector $\myvector{v}\! \in\! \R^{I}$,
a matrix $\mymatrix{M} \in \R^{I \times J}$, 
and a $3$-order tensor $\tensor{T} \in \R^{I \times J \times K}$.}}
\label{fig:diagrams}
\end{minipage}
\end{figure}

\paragraph{Notations.} 
An $m$-dimensional array $\tensor{T}$ is defined as an $m$-{\em order} tensor 
$\tensor{T} \in \R^{I_0 \times \cdots \times I_{m-1}}$.
Its $(i_0, \cdots, i_{n-1}, i_{n+1}, \cdots, i_{m-1})^{\tha}$ {\em mode}-$n$ {\em fiber}, a vector along the $n^{\tha}$ axis, is denoted as $\tensorSub{T}{i_0,\cdots, i_{n-1}, \mathbf{:}, i_{n+1},\cdots, i_{m-1}}$.

\paragraph{Tensor Diagrams.}
Following the convention in quantum physics~\cite{orus2014practical, grasedyck2013literature}, Figure~\ref{fig:diagrams} introduces {\em tensor diagrams}, graphical representations for multi-dimensional objects. In {tensor diagrams}, an array (scalar/vector/matrix/tensor) is represented as a {\em node} in the graph, and its {\em order} is denoted by the number of {\em edges} extending from the node, where each edge corresponds to one {\em mode} (whose {\em dimension} is denoted by the number associated to the edge). 


\begin{figure*}[!htbp]
\centering
\begin{minipage}{\textwidth}
\begin{subfigure}[b]{0.48\textwidth}
\centering
\psfrag{n1}[][][0.8]{$\tensor{X}$}
\psfrag{n2}[][][0.8]{$\tensor{Y}$}
\psfrag{n3}[][][0.6]{$\tensor{T}^{1}$}
\psfrag{S1}[][][0.6]{$I_0$}
\psfrag{S2}[][][0.6]{$I_1$}
\psfrag{S3}[][][0.6]{$I_2$}
\psfrag{T1}[][][0.6]{$J_0$}
\psfrag{T2}[][][0.6]{$J_1$}
\psfrag{T3}[][][0.6]{$J_2$}
\psfrag{S1=T2}[][][0.6]{$I_0 = J_1$}
\includegraphics[height=0.65in]{\fighome/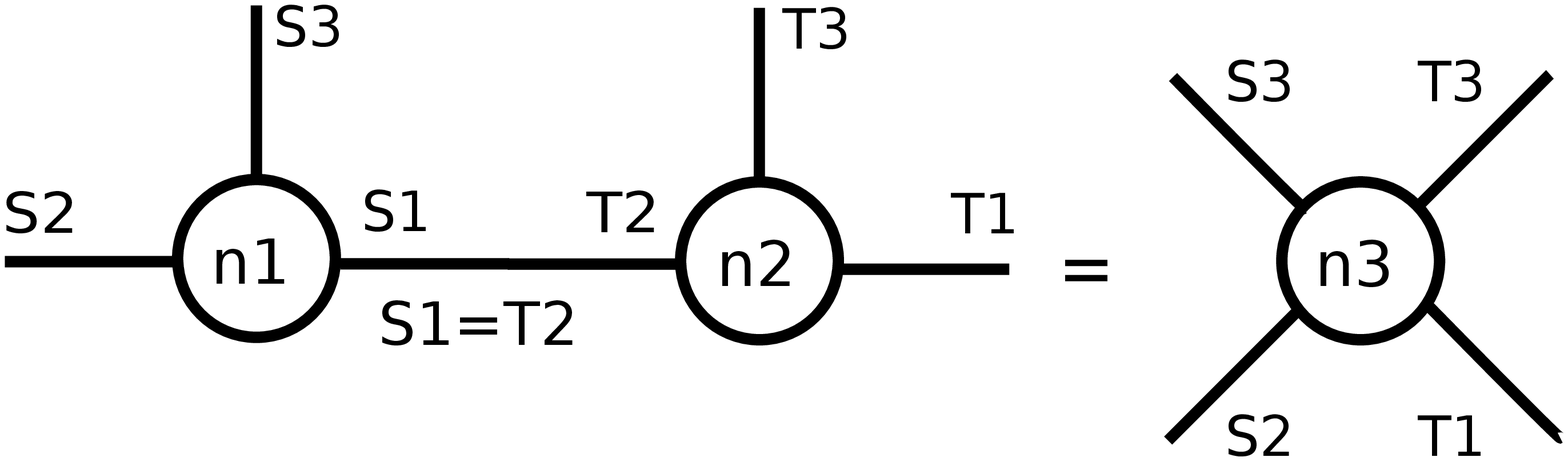}
\caption{\begin{tabular}{c} Mode-(0,1) tensor contraction $\tensor{X} \times^{0}_{1} \tensor{Y} \rightarrow \tensor{T}^{1}$ \\
$\tensorSub{T}{i_1, i_2, j_0, j_2}^{1} = \sum_{r} \tensorSub{X}{r, i_1, i_2} \tensorSub{Y}{j_0, r, j_2}$ \end{tabular}}
\label{fig:operation-contraction}
\end{subfigure}
\hfill
\begin{subfigure}[b]{0.48\textwidth}
\centering
\psfrag{n1}[][][0.8]{$\tensor{X}$}
\psfrag{n2}[][][0.8]{$\mymatrix{M}$}
\psfrag{n3}[][][0.6]{$\tensor{T}^{2}$}
\psfrag{S1}[][][0.6]{$I_0$}
\psfrag{S2}[][][0.6]{$I_1$}
\psfrag{S3}[][][0.6]{$I_2$}
\psfrag{T1}[][][0.6]{$J_1$}
\psfrag{T2}[][][0.6]{$J_0$}
\psfrag{S1=T2}[][][0.6]{$I_0 = J_0$}
{$\ $\vspace{0.183in}$\ $
\includegraphics[height=0.467in]{\fighome/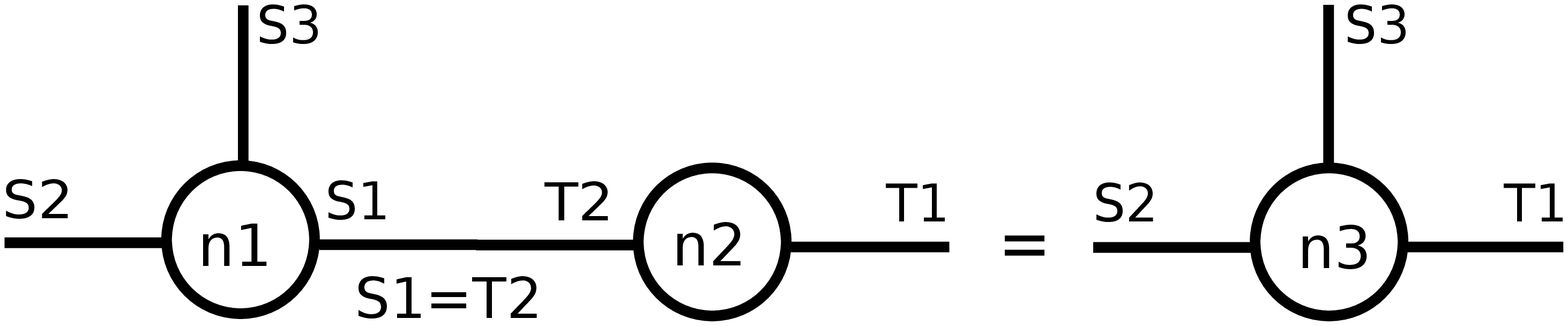}}
\caption{\begin{tabular}{c} Mode-0 tensor product/multiplication \\
$\tensor{X} \times_{0} \mymatrix{M}  \rightarrow \tensor{T}^{2}$: $\tensorSub{T}{j_0, i_1, i_2}^{2} = \sum_{r}\tensorSub{X}{r, i_1, i_2} \matrixSub{M}{j_0, r}$ \end{tabular}}
\label{fig:operation-multiplication}
\end{subfigure}
\vskip \baselineskip
\begin{subfigure}[b]{0.48\textwidth}
\centering
\psfrag{n1}[][][0.8]{$\tensor{X}$}
\psfrag{n2}[][][0.8]{$\tensor{Y}$}
\psfrag{n3}[][][0.6]{$\tensor{T}^{3}$}
\psfrag{S1}[][][0.6]{$I_0$}
\psfrag{S2}[][][0.6]{$I_1$}
\psfrag{S3}[][][0.6]{$I_2$}
\psfrag{T1}[][][0.6]{$J_0$}
\psfrag{T2}[][][0.6]{$J_1$}
\psfrag{T3}[][][0.6]{$J_2$}
\psfrag{N}[][][0.6]{$I_0^{\prime}$}
\includegraphics[height=0.65in]{\fighome/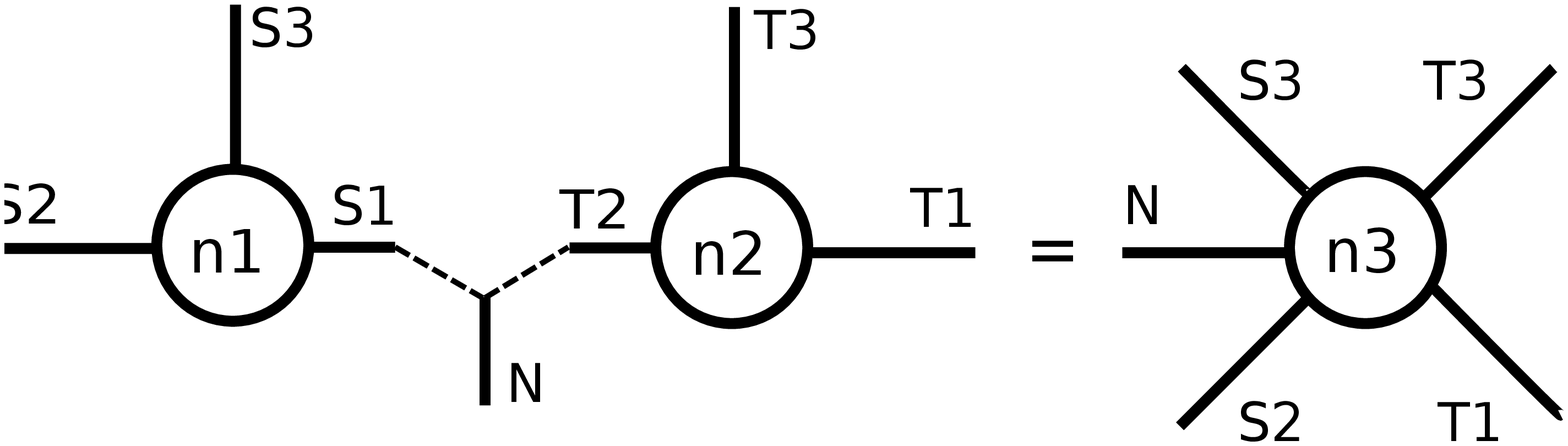}
\caption{\begin{tabular}{c}Mode-(0,1) tensor convolution $\tensor{X} \ast^{0}_{1} \tensor{Y}  \rightarrow \tensor{T}^{3}$ \\
$\tensorSub{T}{\mathbf{:}, i_1, i_2, j_0, j_2}^{3} = \tensorSub{X}{\mathbf{:}, i_1, i_2} \ast \tensorSub{Y}{j_0, \mathbf{:}, j_2}$ \end{tabular}}
\label{fig:operation-convolution}
\end{subfigure}
\hfill
\begin{subfigure}[b]{0.49\textwidth}
\centering
\psfrag{n1}[][][0.8]{$\tensor{X}$}
\psfrag{n2}[][][0.8]{$\tensor{Y}$}
\psfrag{n3}[][][0.6]{$\tensor{T}^{4}$}
\psfrag{S1}[][][0.6]{$I_0$}
\psfrag{S2}[][][0.6]{$I_1$}
\psfrag{S3}[][][0.6]{$I_2$}
\psfrag{T1}[][][0.6]{$J_0$}
\psfrag{T2}[][][0.6]{$J_1$}
\psfrag{T3}[][][0.6]{$J_2$}
\psfrag{N}[][][0.6]{$I_0$}
\psfrag{I0=J1 }[][][0.6]{$I_0=J_1$}
\includegraphics[height=0.65in]{\fighome/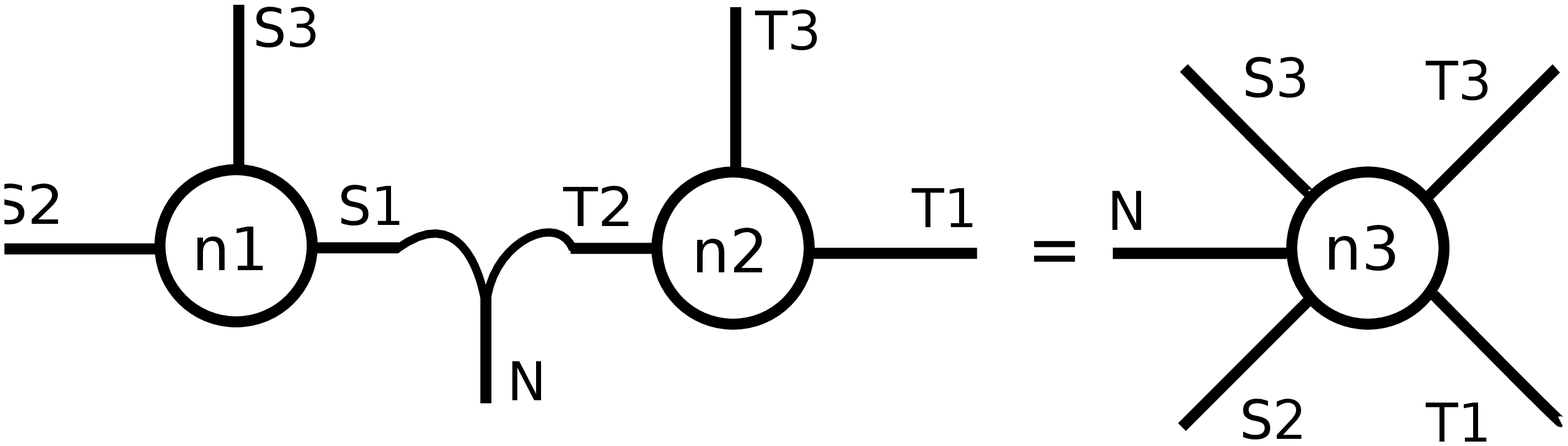}
\caption{\begin{tabular}{c} Mode-(0,1) tensor partial outer product  \\
$\tensor{X} \otimes^{0}_{1} \tensor{Y}  \rightarrow \tensor{T}^{4}$: $\tensorSub{T}{r, i_1, i_2, j_0, j_2}^{4} = \tensorSub{X}{r, i_1, i_2} \tensorSub{Y}{j_0, r, j_2}$ \end{tabular}}
\label{fig:diagram-outer-product}
\end{subfigure}
\caption{\textbf{Generalized tensor operations.}
Examples of tensor operations in which $\mymatrix{M} \in \R^{J_0 \times J_1}$, $\tensor{X} \in \R^{I_0 \times I_1 \times I_2}$ and $\tensor{Y} \in \R^{J_0 \times J_1 \times J_2}$ are input matrix/tensors, and $\tensor{T}^{1} \in \R^{I_1 \times I_2 \times J_0 \times J_2}$, $\tensor{T}^{2} \in \R^{J_0 \times I_1 \times I_2}$, $\tensor{T}^{3} \in \R^{I_0^{\prime} \times I_1 \times I_2 \times J_0 \times J_2}$ and $\tensor{T}^{4} \in \R^{I_0 \times I_1 \times I_2 \times J_0 \times J_2}$ are output tensors of corresponding operations. Similar definitions apply to general mode-$(i, j)$ tensor operations. Existing tensor operations are only defined on lower-order $\tensor{X}$ and $\tensor{Y}$ such as matrices and vectors.
}
\label{fig:operations}
\end{minipage}
\end{figure*}

\paragraph{Generalized Tensor Operations.}
We introduce {\em generalized tensor operations} on {\em high-order tensor operands} that consist of four primitive operations in Figure~\ref{fig:operations}, extending the usual operations with vector or matrix operands.
In tensor diagrams, an operation is represented by linking edges from the input tensors, where the type of operation is denoted by the shape of line that connects the nodes: solid line stands for {\em tensor contraction} or {\em tensor multiplication}, dashed line represents {\em tensor convolution}, and curved line is for {\em tensor partial outer product}. 
We illustrate our generalized tensor operations via simple examples where third-order operands $\tensor{X}$ and $\tensor{Y}$ are considered in Figure~\ref{fig:operations}. However, our generalized tensor operations can extend to higher-order tensors as rigorously defined in Appendix~\ref{app:operations}.

\section{Tensorial Neural Networks} 
\label{sec:tnn}

\begin{figure}[!htbp]
\begin{subfigure}[b]{0.08\textwidth}
	\psfrag{x}[][][0.5]{$\tensor{U}$}
	\psfrag{nl}[][][0.5]{$\relu$}
	\psfrag{n1}[][][0.5]{$\tensorSup{K}0$}
	\psfrag{n2}[][][0.5]{$\tensorSup{K}1$}
	\psfrag{y}[][][0.5]{$\tensorSup{K}2$}
	\includegraphics[height=3.2in]{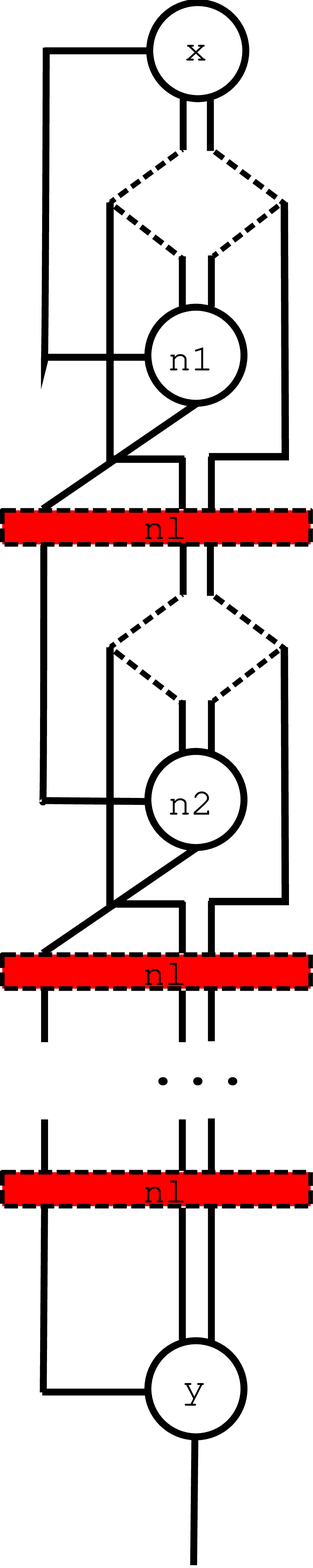}
\caption{\convnetshort}
\label{fig:nn}
\end{subfigure}
\hfill
\begin{subfigure}[b]{0.4\textwidth}
\centering
	\psfrag{x}[][][0.4]{$\tensor{U^\prime}$}
	\psfrag{a1}[][][0.4]{$\tensorInd{K}{0}{0}$}
	\psfrag{a2}[][][0.4]{$\tensorInd{K}{1}{0}$}
	\psfrag{a3}[][][0.4]{$\tensorInd{K}{2}{0}$}
	\psfrag{a4}[][][0.4]{$\tensorInd{K}{m-2}{0}$}
	\psfrag{b1}[][][0.4]{$\tensorInd{K}{0}{0}$}
	\psfrag{b2}[][][0.4]{$\tensorInd{K}{1}{1}$}
	\psfrag{b3}[][][0.4]{$\tensorInd{K}{2}{1}$}
	\psfrag{b4}[][][0.4]{$\tensorInd{K}{m-2}{1}$}
	\psfrag{c1}[][][0.4]{$\tensorInd{K}{0}{1}$}
	\psfrag{c2}[][][0.4]{$\tensorInd{K}{1}{2}$}
	\psfrag{c3}[][][0.4]{$\tensorInd{K}{2}{2}$}
	\psfrag{c4}[][][0.4]{$\tensorInd{K}{m-2}{2}$}
	\psfrag{y}[][][0.5]{$\tensorSup{K}{3}$}
	\psfrag{nl}[][][0.5]{$\relu$}
	\includegraphics[height=3.2in]{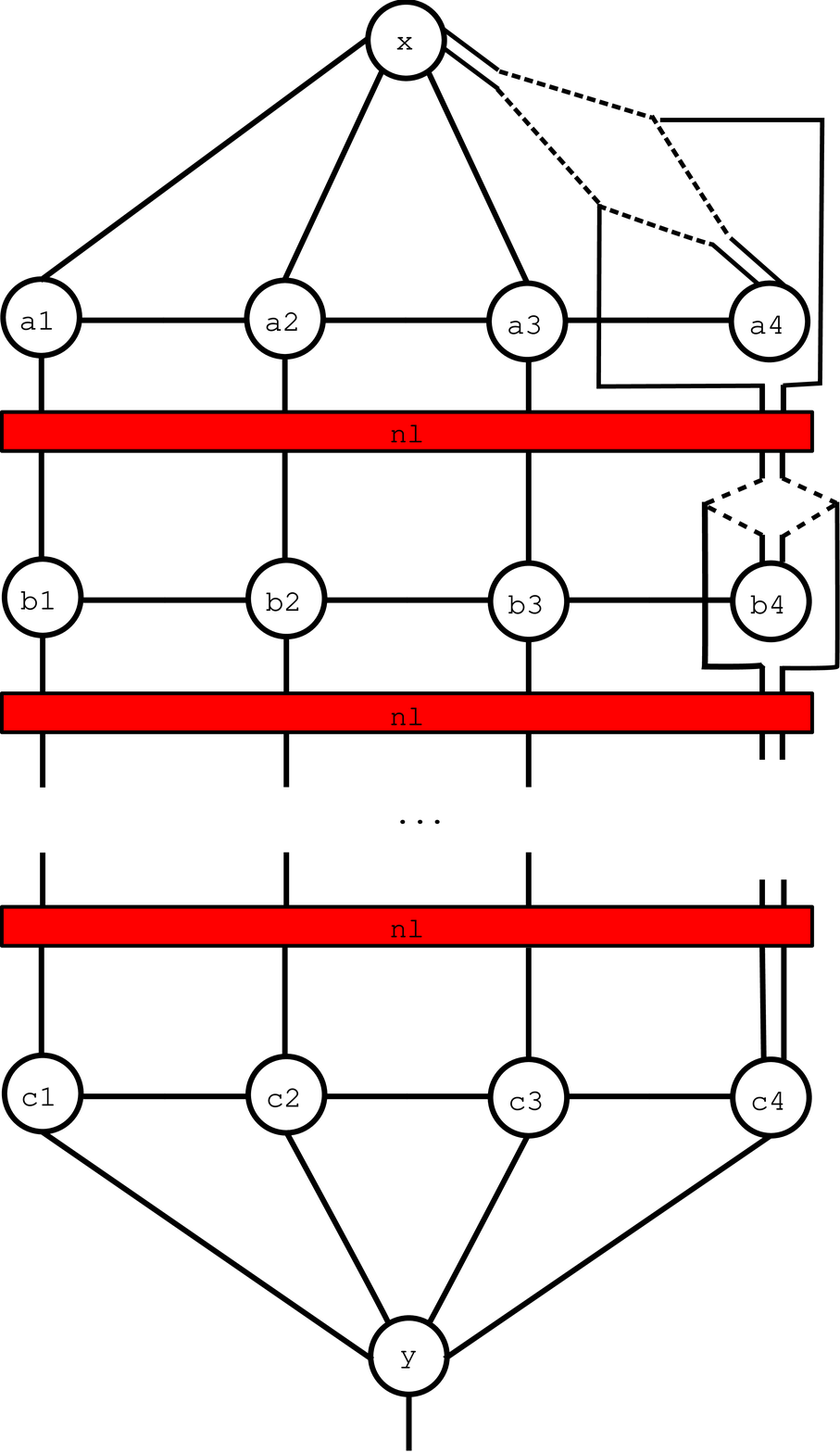}
\caption{\tensornetshort}
\label{fig:tnn}
\end{subfigure}
\begin{subfigure}[b]{0.15\textwidth}
\centering
	\psfrag{n1}[][][0.6]{$\tensor{U}$}
	\psfrag{n2}[][][0.6]{$\tensor{K}$}
	\psfrag{X}[][][0.6]{$X$}
	\psfrag{Y}[][][0.6]{$Y$}
	\psfrag{H}[][][0.6]{$H$}
	\psfrag{W}[][][0.6]{$W$}
	\psfrag{S}[][][0.6]{$S$}
	\psfrag{T}[][][0.6]{$T$}
	\psfrag{M}[][][0.6]{$X^{\prime}$}
	\psfrag{N}[][][0.6]{$Y^{\prime}$}
	\includegraphics[width=0.667\textwidth]{\fighome/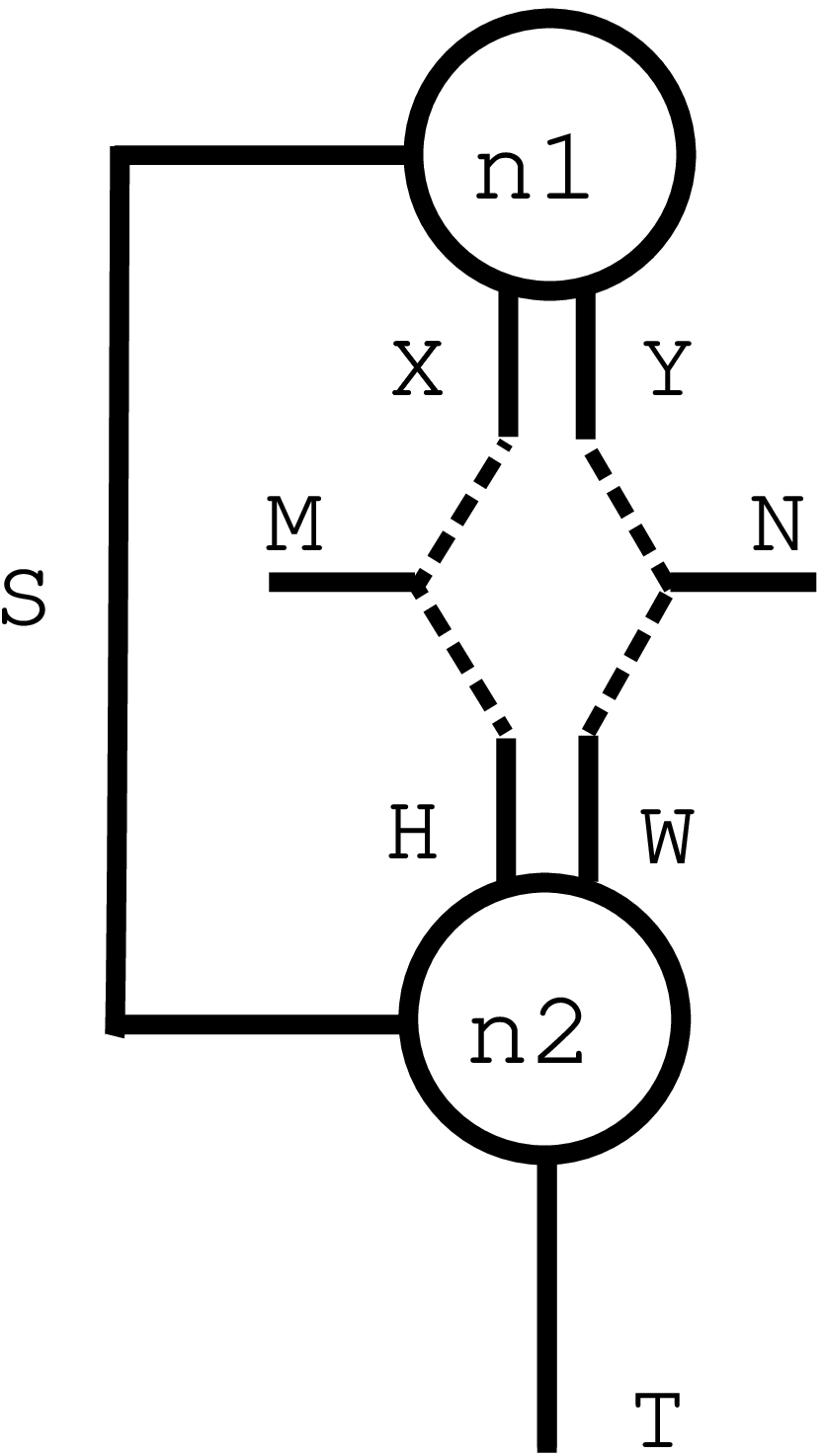}
\caption{1 layer of \convnetshort}
\label{fig:process-original}
\end{subfigure}
\hfill
\begin{subfigure}[b]{0.33\textwidth}
\centering
	\psfrag{n1}[][][0.6]{$\tensor{U}^{\prime}$}
	\psfrag{n2}[][][0.6]{$\tensorSub{K}{0}$}
	\psfrag{n3}[][][0.6]{$\tensorSub{K}{1}$}
	\psfrag{n4}[][][0.6]{$\tensorSub{K}{2}$}
	\psfrag{n5}[][][0.4]{$\tensorSub{K}{m-2}$}
	\psfrag{X}[][][0.6]{$X$}
	\psfrag{Y}[][][0.6]{$Y$}
	\psfrag{H}[][][0.6]{$H$}
	\psfrag{W}[][][0.6]{$W$}
	\psfrag{S1}[][][0.6]{$S_0$}
	\psfrag{S2}[][][0.6]{$S_1$}
	\psfrag{S3}[][][0.6]{$S_2$}
	\psfrag{T1}[][][0.6]{$T_0$}
	\psfrag{T2}[][][0.6]{$T_1$}
	\psfrag{T3}[][][0.6]{$T_2$}
	\psfrag{R1}[][][0.6]{$R_0$}
	\psfrag{R2}[][][0.6]{$R_1$}
	\psfrag{R3}[][][0.6]{$R_2$}
	\psfrag{M}[][][0.6]{$Y^{\prime}$}
	\psfrag{N}[][][0.6]{$X^{\prime}$}
	\includegraphics[width=\textwidth]{\fighome/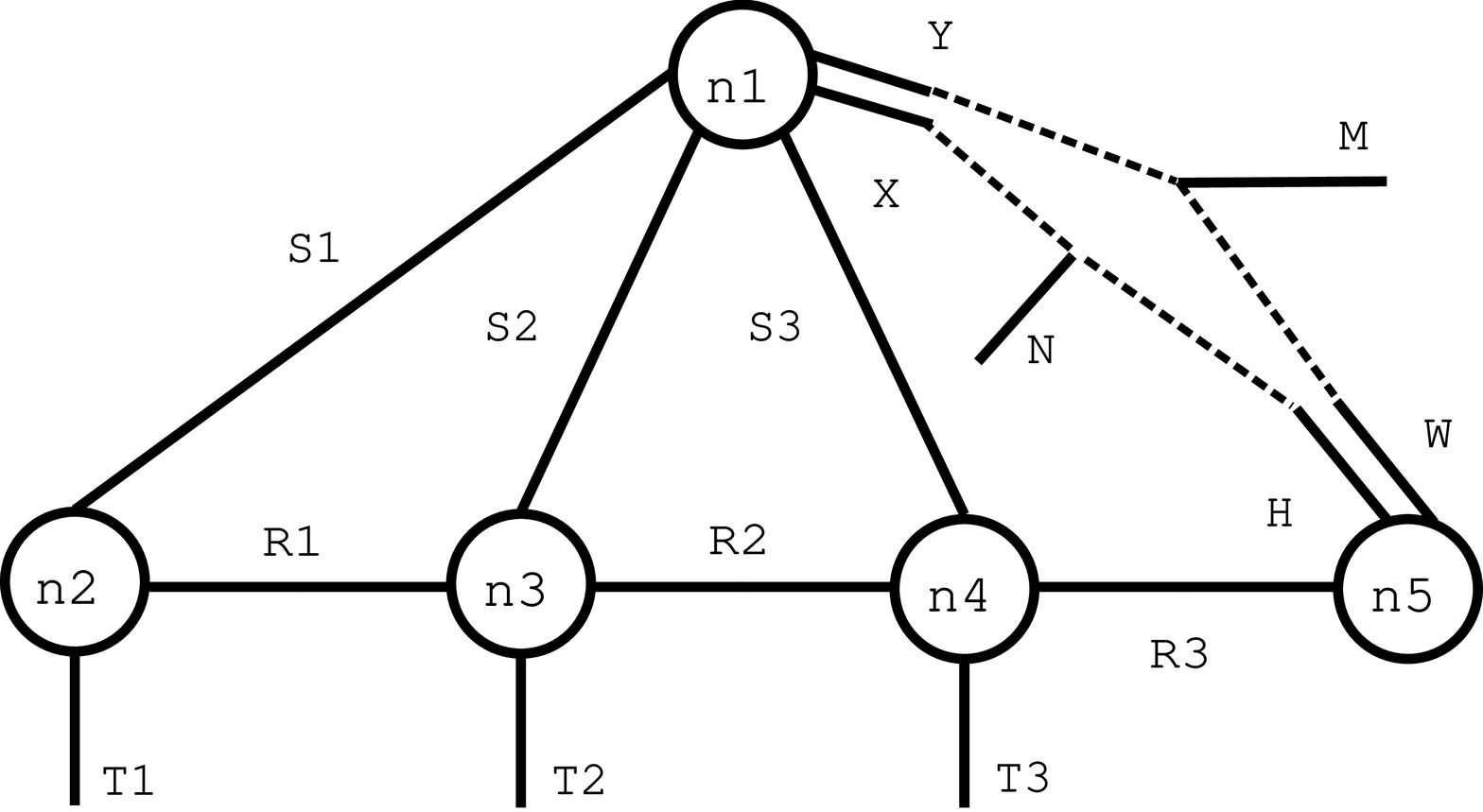}
\caption{One layer of \tensornetshort}
\label{fig:process-rtt}
\end{subfigure}
\caption{\textbf{Tensor diagram} of {\convnetshort}s and {\tensornetshort}s.}
\end{figure}

A traditional convolutional neural network using tensor diagram is depicted in Figure~\ref{fig:nn}. We propose a richer class of functions, \tensornet ({\tensornetshort}s), whose the input are tensors of any order (denoted $m$) and operations are generalized tensor operations.
{\tensornetshort} are generalization of traditional {\neuralnet}: if the input is a vector and operation is matrix-vector multiplication, \tensornetshort reduces to {\em multi-layer perceptrons (MLP)}; and if the input is a feature map and operation is convolution, \tensornetshort reduces to {\em \convnet (\convnetshort)}. 
The order of the input tensor and the type of the generalized tensor operation are hyperparameters that define the {\em architecture} of a \tensornetshort. 
In this paper, we design a number of successful architectures of {\tensornet} (see Appendices~\ref{app:dense-tensorized} and \ref{app:convolutional-tensorized}),
and an example of \textit{\TTlong \tensornetshort} is illustrated in Figure~\ref{fig:tnn}.


\subsection{One-layer of \tensornetshort vs One-layer of \convnetshort}
In \convnetshort, each linear layer is a special convolution operation between a {\em low-order} input and another {\em low-order} weights kernel.
In contrast, each layer in \tensornetshort is characterized by a generalized tensor operation between a {\em high-order} input and another {\em high-order} weights kernel, allowing for  higher expressive power. 

\paragraph{One-layer of CNN}
A convolutional layer in traditional \neuralnet is parameterized by a $4$-order kernel $\tensor{K} \in \R^{H \times W \times S \times T}$, 
where $H, W$ are height/width of the filters, and $S, T$ are the numbers of input/output channels. 
The layer maps a $3$-order input tensor $\tensor{U} \in \R^{X \times Y \times S}$ to another $3$-order output tensor $\tensor{V} \in \R^{X^{\prime} \times Y^{\prime} \times T}$, where $X, Y$ and $X^{\prime}, Y^{\prime}$ are heights/widths of the input and output feature maps. 
\beq
\tensor{V} = \tensor{U} ~ (\ast^{0}_{0} \circ \ast^{1}_{1} \circ \times^{2}_{2}) ~ \tensor{K}
\label{def:convolutional-main}
\eeq
We define such an operation as \textit{compound operation}, where multiple edges are linked simultaneously (see Figure~\ref{fig:process-original} for the tensor diagram of a convolutional layer).
General compound operations are discussed systematically in Appendices~\ref{app:operations} and~\ref{app:derivatives}, and in general {\tensornetshort}s allow arbitrary compound operation to be used at each linear layer. 

\paragraph{One-layer of \tensornetshort} 
We illustrate one layer of the {\TTlong \tensornetshort} in Figure~\ref{fig:process-rtt} (with input tensor of order $m = 5$), where the parameters are characterized by $(m-1)$ kernels $\{ \tensorSub{K}{i} \}_{i = 0}^{m - 2}$.
The multi-dimensional structure lying in the higher-order input tensor $\tensor{U}^{\prime}$ is effectively preserved:
each mode $i$ of the input tensor $\tensor{U}^{\prime}$ contracts with the corresponding kernel $\tensorSub{K}{i}$. {\TTlong \tensornetshort} allows interactions between the modes by the contraction edges between adjacent kernels $\tensorSub{K}{i}$ and $\tensorSub{K}{i+1}$. These edges are crucial for modeling general multi-dimensional transformations while preserving the structures of the input data. 
Effectively, each layer of {\TTlong \tensornetshort} implements a multi-dimensional propagation of the high-order input.

\paragraph{Relationship between {\tensornetshort}s and {\convnetshort}s}
(1) \textit{{\tensornetshort} generalizes {\neuralnetshort}}. Formally, for any $q > 0$, $\mathcal{G}^{q} \subseteq \mathcal{H}^{q}$ holds. 
Consider a special case of Figure~\ref{fig:process-rtt} where $\tensor{K}_0$ and $\tensorSub{K}{1}$ are identical mappings and contraction of $\tensorSub{K}{2}$ and $\tensorSub{K}{m-2}$ equals to a kernel $\tensor{K}$ as in Figure~\ref{fig:process-original}, the \tensornetshort reduces to a \convnetshort.  
(2) \textit{\neuralnetshort can be mapped to \tensornetshort with fewer parameters}. Formally, there exists $p \leq q$ such that $\mathcal{H}^{p} \subseteq \mathcal{G}^{q}$.
Given an $h^{p} \in \mathcal{H}^{p}$, we could factorize $\tensor{K}$ into $\{\tensorSub{K}{i} \}_{i = 0}^{m - 2}$ using generalized tensor decomposition (as detailed in Section~\ref{sec:invariant} and Appendix~\ref{app:decompositions}), therefore find $q \leq p$ and $g^{q} \in \mathcal{G}^q$ ($\{\tensorSub{K}{i} \}_{i = 0}^{m - 2}$ altogether have fewer parameters than $\tensor{K}$).

\paragraph{Why are {\tensornetshort}s more flexible than {\neuralnetshort}s?}
(1) {\tensornetshort}s are more flexible than {\neuralnetshort}s as the input can be tensors of arbitrary order, which naturally deals with data represented by multi-dimensional array without breaking their internal structures. 
(2) Even in the scenario that the input data is a vector or a matrix, one can reshape the input into higher-order tensor in order to exploit the invariant structures within the data discussed in Section~\ref{sec:invariant}.

\subsection{Prediction in \tensornetshort}
Prediction with a \tensornetshort proceeds similarly as a normal neural network: the input is passed through the layers in \tensornetshort in a feedforward manner, where each layer is a generalized tensor operation (multilinear operation) between the input and the model parameters followed by a nonlinear activation function such as $\relu$.  
When the number of operands in a generalized tensor operation is greater than two (i.e. the layer is characterized by more than one kernel), it is generally NP-hard to find the best order to evaluate the operation. We identify efficient strategies for all \tensornetshort architectures in this paper. For example, each layer in {\TTlong \tensornetshort} can be efficiently evaluated as:
\begin{subequations}
\begin{gather}
\tensorSub{U}{i+1} = \tensorSup{U}{i} ~ \left( \times^{2}_{1} \circ \times^{-1}_{0} \right) ~ \tensorSub{K}{i} \\
\tensor{V}^{\prime} = \tensorSub{U}{m-1} ~ \left( \ast^{0}_{1} \circ \ast^{1}_{2} \circ \times^{-1}_{0} \right) ~ \tensorSub{K}{m}
\end{gather}
\end{subequations}
where $\tensorSub{U}{0} = \tensor{U}^{\prime}$ and $\tensor{V}^{\prime}$ are the input/output of the layer, $\tensorSub{U}{i}$ is the intermediate result after interacting with $\tensorSub{K}{l}$. The forward pass for other architectures are derived in Appendix~\ref{app:dense-tensorized} and~\ref{app:convolutional-tensorized}, with their complexities summarized in Table~\ref{table:convolutional-main}. 

\subsection{Learning in \tensornetshort}
Learning in a \tensornetshort corresponds to {\em hierarchical nonlinear general tensor decomposition}, which recovers all tensors in the \tensornetshort. 
Solving the decomposition problem in closed form is difficult, therefore we derive the backpropagation rules for general tensor operations (in Appendix~\ref{app:derivatives}) such that the tensors can be recovered by {\em stochastic gradient descent} approximately.
For each layer in \tensornetshort, backpropagation requires computations of the derivatives of some loss function $\mathcal{L}$ w.r.t. the input ${\partial \mathcal{L}}/{\partial \tensor{U}^{\prime}}$ and kernel factors $\{ {\partial \mathcal{L}}/{\partial \tensor{K}_{i}}\}_{i = 0}^{m-2}$ given ${\partial \mathcal{L}}/ \partial \tensor{V}^{\prime}$.
As in forward pass, identifying the best strategy to compute these backpropagation equations simultaneously is NP-hard, and we need to develop efficient algorithm for each architecture. For the case of {\TTlong \tensornetshort}, the algorithm proceeds as follows:
{\small
\begin{subequations}
\begin{gather}
\frac{\partial \mathcal{L}}{\partial \tensorSub{U}{m-2}} = \frac{\partial \mathcal{L}}{\partial \tensor{V}^{\prime}} \left( (\ast^{0}_{1})^{\top} \circ (\ast^{2}_{1})^{\top} \right) \tensorSub{K}{m-2} \\
\frac{\partial \mathcal{L}}{\partial \tensorSub{K}{m-2}} = \frac{\partial \mathcal{L}}{\partial \tensor{V}^\prime} \left( (\ast^{0}_{0})^{\top} \circ (\ast^{1}_{1})^{\top} \circ \times^{2}_{2} \cdots \times^{m-1}_{m-1} \right) \tensorSub{U}{m-2} \\
\frac{\partial \mathcal{L}}{\partial \tensorSub{U}{i}} = \swapaxes \big( \tensorSub{K}{i} \left( \times^{2}_{-2} \circ \times^{3}_{-1} \right) \frac{\partial \mathcal{L}}{\partial \tensorSub{U}{i+1}} \big) \\
\frac{\partial \mathcal{L}}{\partial \tensorSub{K}{i}} = \swapaxes \big( \tensorSub{U}{i} \left( \times^{0}_{0} \circ \times^{1}_{1} \circ \times^{3}_{2} \cdots \times^{m-1}_{m-2} \right) \frac{\partial \mathcal{L}}{\partial \tensorSub{U}{i+1}} \big) 
\end{gather}
\end{subequations}
}
where $\swapaxes(\cdot)$ permutes the order as needed (explained in Appendix~\ref{app:notations}). The algorithms for other architectures are derived in appendices~\ref{app:dense-tensorized} and~\ref{app:convolutional-tensorized}, whose time complexities are summarized in Table~\ref{table:convolutional-main}. 


\begin{table*}[!htbp]
\centering
\begin{tabular}{ c | c | c | c }
Architect. & $O$(\# of parameters) & $O$(\# of forward ops.) & $O$(\# of backward ops.) \\ 
\hline
original & $k^2 N^2$ & $k^2 N^2 D^2$ & $N^2 D^4$ \\
\hline 
\tnnCP & $(k^2 + m N^{\frac{2}{m}}) R$ & $(m N^{1 + \frac{1}{m}} + k^2 N) R D^2$ & $(m N^{1 + \frac{1}{m}} + N D^2) R D^2$ \\
\tnnTK & $(k^2 R^{2m-1} + 2mN) R$ & $(k^2 R^{2m-1} + 2mN) R D^2$ & $(R^{2m-1} D^2 + 2mN) R D^2$ \\
\tnnTT & $(m N^{\frac{2}{m}} R + k^2) R$ &
$(m N^{1 + \frac{1}{m}} R + k^2 N) R D^2$ & $(m N^{1 + \frac{1}{m}} R + N D^2) R D^2$ \\
\end{tabular}
\caption{Number of parameters and operations required by one compressed convolutional layer using various types of tensor decompositions as the projection method. This table is for the special case when $X = Y = X^{\prime} = Y^{\prime} = D $, $S = T = N$, $H = W = k$ and $D \gg k$ (see equation~\ref{def:convolutional-main}). 
General settings are summarized in Tables~\ref{table:convolutional} and \ref{table:convolutional-tensorized}.}
\label{table:convolutional-main}
\end{table*}



\section{Compression of Neural Networks via \tensornetshort}
\label{sec:invariant}
Suppose we are given a pre-trained neural network $g^{q} \in \mathcal{G}^{q}$ with $q$ parameters (as described in Figure~\ref{fig:framework}) and we want to compress it to $p$ parameters such that $p \ll q$. 
If we are looking for a compressed model in the class of traditional \neuralnet $\mathcal{G}^{p}$, any compression algorithm can at best find $g^p$, while if we consider a boarder class of compressed models in our proposed {\tensornetshort}s $\mathcal{H}^{p}$, a good enough compression algorithm finds $h^p \in \mathcal{H}^{p} \setminus \mathcal{G}^{p}$.
As is demonstrated in Figure~\ref{fig:framework}, $h^p$ is closer to the uncompressed \neuralnetshort $g^{q}$ than $g^{p}$, therefore outperforms $g^{p}$ in its predictive accuracy.

We introduce a compression algorithm that projects a pre-trained \neuralnetshort $g \in \mathcal{G}^{q}$ to a \tensornetshort $h^{\star} \in \mathcal{H}^{p}$. Superscripts on $g$ and $h$ are omitted for notational simplicity. Suppose the input to $g$ is $\tensor{U}$, our goal is to find a  $h^{\star}$ such that
\beq
h^{\star} = \arg \min_{h \in \mathcal{H}^p} \mathsf{dist}(h(\tensor{U}^{\prime}), g(\tensor{U}))
\label{opt:}
\eeq
where $m$-order $\tensor{U}^\prime$ is the reshaped version of $\tensor{U}$ (to fit the input requirement of $h$), and $ \mathsf{dist}(\cdot, \cdot)$ denotes any distance, such as $\ell_2$ distance, between the outputs of $h$ and $g$.

\subsection{Tensorization: Exploiting Invariant Structures}
\begin{figure}[!htbp]
\centering
	\includegraphics[width=0.46\textwidth]{\fighome/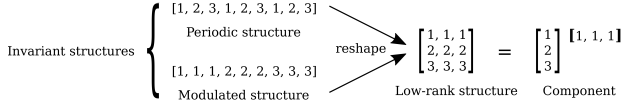}
	\caption{{A toy example of invariant structures. The periodic and modulated structures are picked out by exploiting the low rank structure in the reshaped matrix.}}
\label{fig:demo-invariant}
\end{figure}

We first reshape the input data $\tensor{U}$ to $m$-order tensor $\tensor{U}^{\prime}$. 
Consider this toy example of a vector with periodic structure $[1,2,3,1,2,3,1,2,3]$ or modulated structure $[1,1,1,2,2,2,3,3,3]$ in Figure~\ref{fig:demo-invariant}. The number of parameters needed to represent this vector, naively, is 9. However if we map or \emph{reshape} the vector into a higher order object, for instance, a matrix ($m = 2$)$[1,1,1;2,2,2;3,3,3]$ where the columns of the matrix are repeated, then apparently this reshaped matrix can be decomposed into rank one without losing information. Therefore only 6 parameters are needed to represent the original length-9 vector.

\noindent Now we are ready to learn the mapping from \neuralnetshort to \tensornetshort.

\subsection{Mapping NN to TNN}

\paragraph{End-to-End (\etoe) Learning}
Let the $l$-th layer of \tensornetshort $h$ be parameterized by $(m-1)$ kernels $\{\tensorInd{K}{i}{l}\}_{i=0}^{m-2}$. if $\ell_2$ distance is used, the compression reduces to an optimization problem : 
\beq
\{ {\tensorInd{K}{i}{l}}^{\star} \}_{i, l} = \arg \min_{\{ \tensorInd{K}{i}{l} \}_{i, l}} \lVert h(\tensor{U}^\prime) - g(\tensor{U})\rVert_2^2
\label{eq:end-to-end}
\eeq
Solving the problem by backpropagation is very expensive as it requires end-to-end gradients flow in \tensornetshort, therefore we relax it into a sequence of problems as follows.

 \vspace{-0.5em}
 
\paragraph{\SeqTrain (\seqTune) Learning} Let $\tensorSup{V}{l}$ and $\tensorSup{V^\prime}{l}$ be the outputs of the $l$-th layers of $g$ and $h$. We propose to project each layer of $g$ to a layer of $h$ bottom-up sequentially, i.e. for $l = 0, \cdots, L$:
\beq
\{ {\tensorInd{K}{i}{l}}^{\star} \}_{i} =\arg \min_{\{\tensorInd{K}{i}{l}\}_{i}} \lVert \tensorSup{V}{l} - \tensorSup{V^\prime}{l}\rVert_2^2
\label{eq:sequential}
\eeq
where the input to these layers are fixed to $\tensorSup{V}{l-1}$ and $\tensorSup{V^{\prime}}{l-1}$ respectively. 
 
 \vspace{-0.5em}
 
\paragraph{Generalized Tensor Decomposition}
Since each layer of \tensornetshort should approximate the pre-trained \neuralnetshort, our goal is to find ${\tensorInd{K}{i}{l}}^{\star}$ such that their composition is closed to the uncompressed kernel $\tensorSup{K}{l}$, and in {\TTlong \tensornetshort}:
\beq
\tensorSup{K}{l} \approx {\tensorInd{K}{0}{l}}^{\star} \times^{-1}_{0} {\tensorInd{K}{1}{l}}^{\star} \times^{-1}_{0} \cdots \times^{-1}_{0} {\tensorInd{K}{m-2}{l}}^{\star}
\label{eq:convolutional-rtt-main}
\eeq
The equation above is known as {\em generalized tensor decomposition}, because it reverses the mapping of the general tensor operations proposed earlier: given a set of operations and a tensor, generalized tensor decomposition aims to recover the factors/components such that the operations on these factors result in a tensor approximately equal to the original one.  (See Appendix~\ref{general_decomposition} for details).  

\section{Interpretation of Neural Networks Designs}
\label{sec:interpretation}
Recent advances in novel architecture design of neural networks include {\em Inception}~\cite{szegedy2017inception}, {\em Xception}~\cite{chollet2016xception} and {\em Bottleneck structures}~\cite{lin2013network, he2016identity}. We will show, in this section, that all aforementioned architectures can be naturally derived as special cases of our \tensornetshort framework (with small modifications). 

\begin{figure}[!htbp]
\centering
\begin{subfigure}[b]{0.32\textwidth}
\centering
	\psfrag{Inception}[][][0.6]{Inception}
	\psfrag{Layer 1}[][][0.6]{Layer 1}
	\psfrag{Layer 2}[][][0.6]{Layer 2}
	\psfrag{Permutaion}[][][0.6]{Permutation}
	\includegraphics[width=\textwidth]{\fighome/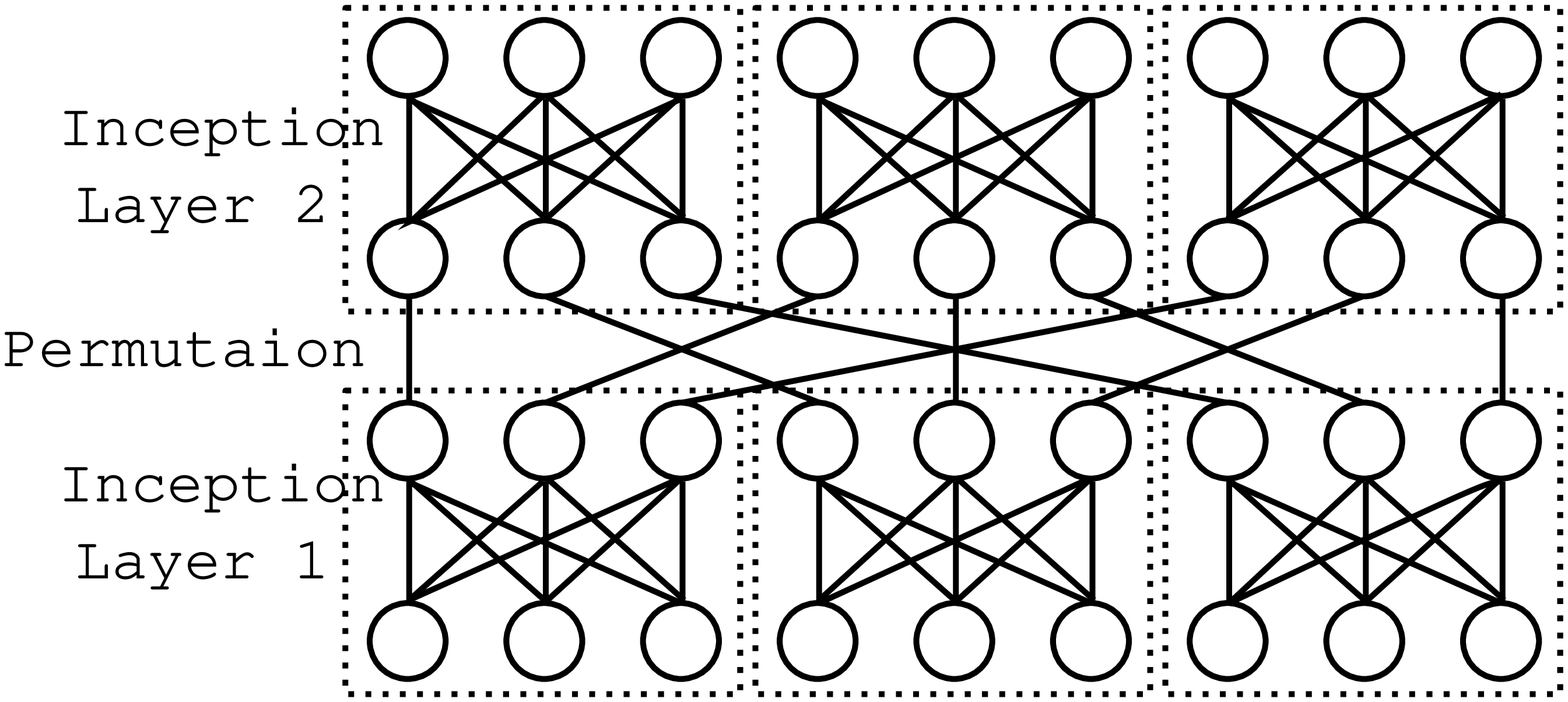}
\caption{Neuron Connections}
\label{fig:inception}
\end{subfigure}
\hfill
\begin{subfigure}[b]{0.14\textwidth}
\centering
	\psfrag{n1}[][][0.6]{$\tensor{U}$}
	\psfrag{n2}[][][0.6]{$\tensor{K}_{1}$}
	\psfrag{n3}[][][0.6]{$\tensor{K}_{2}$}
	\psfrag{1}[][][0.6]{1}
	\psfrag{2}[][][0.6]{2}
	\psfrag{3}[][][0.6]{3}
	\includegraphics[width=0.8\textwidth]{\fighome/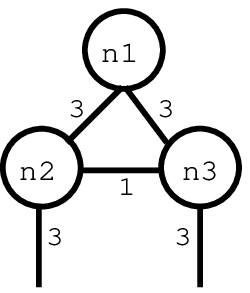}
\caption{Tensor Diagram}
\label{fig:inception-tensordiagram}
\end{subfigure}
\caption{Inception networks.}
\label{fig:inception_all}
\end{figure}

\paragraph{Inception Network.} 
Inception network~\cite{szegedy2017inception} is a special case of {\TTlong \tensornetshort} as shown in Figure~\ref{fig:inception_all}. For instance, an Inception network in Figure~\ref{fig:inception} can be represented using tensor diagram as in Figure~\ref{fig:inception-tensordiagram}, a simplified {\TTlong \tensornetshort} with input tensor of size $3 \times 3$ and the dimension of the connecting edge is $1$. 


\begin{figure}[!htbp]
\centering
\begin{subfigure}[b]{0.32\textwidth}
\centering
	\psfrag{Layer}[][][0.6]{Layer}
	\psfrag{Pointwise}[][][0.6]{Pointwise}
	\psfrag{Depthwise}[][][0.6]{Depthwise}
	\psfrag{Permutaion}[][][0.6]{Permutation}
	\includegraphics[width=\textwidth]{\fighome/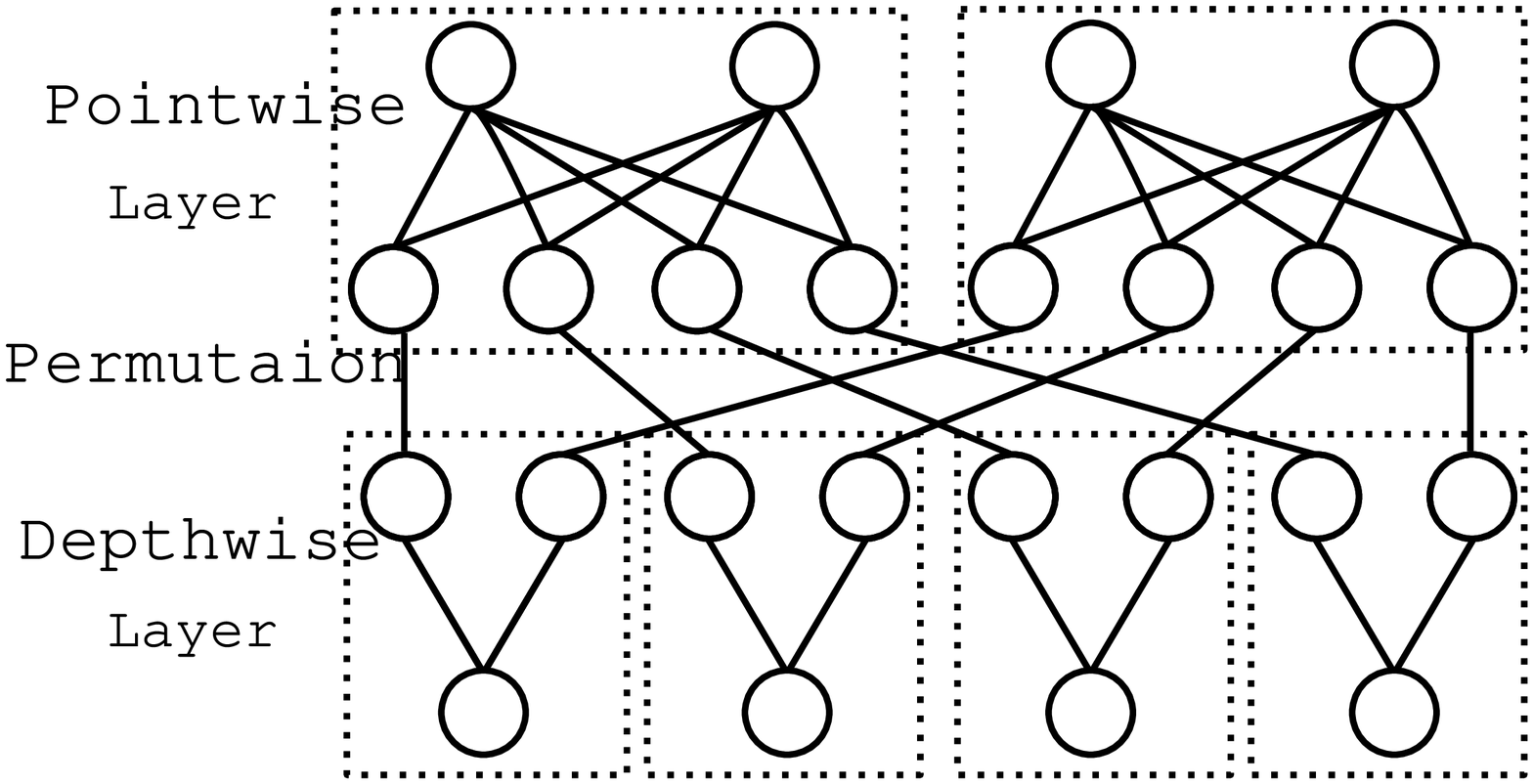}
\caption{}
\label{fig:xception}
\end{subfigure}
\hfill
\begin{subfigure}[b]{0.14\textwidth}
	\psfrag{n1}[][][0.6]{$\tensor{U}$}
	\psfrag{n2}[][][0.6]{$\tensor{K}_{0}$}
	\psfrag{n3}[][][0.6]{$\tensor{K}_{1}$}
	\psfrag{1}[][][0.6]{1}
	\psfrag{2}[][][0.6]{2}
	\psfrag{3}[][][0.6]{3}
	\includegraphics[width=0.8\textwidth]{\fighome/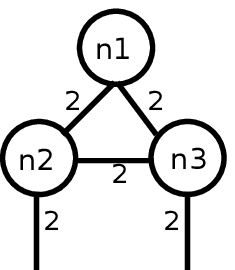}
	\caption{{\small Tensor Diagram}}
	\label{fig:xception-tensordiagram}
\end{subfigure}
\caption{An Xception network.}
\label{fig:xception_all}
\end{figure}

\paragraph{Xception Network.} Xception~\cite{chollet2016xception} is an architecture that further simplifies Inception, where local connection at the \textit{depthwise layers} are further limited to one neuron (or one feature map).
Similar to Inception, the Xception network in Figure~\ref{fig:xception} can be represented using tensor diagram as in Figure~\ref{fig:xception-tensordiagram}, a simple \TTlong \tensornetshort with input tensor of size $2 \times 2$ and the dimension of the connecting edge is $2$. 


\paragraph{Bottleneck Network.} Finally, bottleneck architecture in~\cite{lin2013network, he2016identity} can be interpreted as decomposition of a wider architecture into a tensor network with $3$ kernels.


\begin{figure}[!htbp]
\centering
	\psfrag{kN}[][][0.6]{$kN$}
	\psfrag{N}[][][0.6]{$N$}
	\psfrag{H}[][][0.6]{$H$}
	\psfrag{W}[][][0.6]{$W$}
	\psfrag{n1}[][][0.6]{$\tensor{K}$}
	\psfrag{n2}[][][0.6]{$\tensor{K}_0$}
	\psfrag{n3}[][][0.6]{$\tensor{K}_1$}
	\psfrag{n4}[][][0.6]{$\tensor{K}_2$}
	\psfrag{nl}[][][0.6]{$\relu$}
	\includegraphics[width=0.45\textwidth]{\fighome/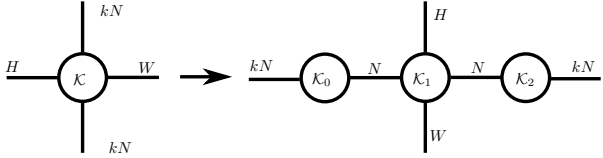}
\caption{Bottleneck structure as tensor decomposition.}
\label{fig:bottleneck}
\end{figure}

\paragraph{Insights for NN designs.} Our \tensornetshort provides some insights on how to design a compact NN architecture in practice: 
(1) First, we can start our design with a traditional neural network (with wide architecture); (2) Second, we factorize the kernel into a tensor network by general tensor decomposition, which converts the original neural network into a \tensornetshort that naturally bear nice compact structures as we see in Figures~\ref{fig:inception} and~\ref{fig:xception}. (3) Since each tensor network corresponds to a compact design, therefore exploration of different designs can be done by designing different tensor networks. 
\section{Experiments}
\label{sec:experiments} 


In this section, we evaluate the effectiveness of our proposed compression algorithm (\ouralgo) proposed in section~\ref{sec:invariant} on several benchmark deep neural networks and datasets. We evaluate fully connected layer compression on MNIST (discussed in Appendix~\ref{app:dense-tensorized}); we evaluate convolutional layer compression on \Resnet-32~\cite{he2016identity} for CIFAR-10; and we evaluate the scalability of our compression algorithm on \Resnet-50~\cite{he2016identity} for ImageNet (2012) dataset.

We conduct experiments on various compression rates and compare the performance of our \ouralgo (\ouralgoshort) methods with that of the state-of-the-art \ourbaseline methods (\ourbaselineshort~\cite{jaderberg2014speeding, lebedev2014speeding, kim2015compression}) under the same compression rate. More details of the state-of-the-art \ourbaseline are provided in appendix~\ref{app:convolutional}. For all experiments, we use the Adam optimizer~\cite{kingma2014adam} with 1e-3 learning rate and decay the learning rate by 10x every 50 epochs. Overall, we show that our \ouralgo maintain high accuracy even when the networks are highly compressed. 

We denote \nCP, \nTK, and \nTT as the compressed neural network obtained by baseline \ourbaseline, which uses classical types of tensor decompositions (\CP, \TK, and \TT) as the layer-wise projecting methods (mentioned in section~\ref{sec:invariant}). While \tnnCP, \tnnTK and \tnnTT are compressed \tensornet obtained by \ouralgo, which uses our proposed modified tensor decomposition (\rCP, \rTK, and \rTT). 

As mentioned in section~\ref{sec:invariant}, we refer to traditional back propogation-based compression of the network as \etoeLong (\etoe) compression, and refer to our strategy of data reconstruction-based sequential compression as \SeqTrain (\seqTune) compression.
\smallskip 

\noindent \textbf{Our algorithm achieves 5\% higher accuracy than baseline on CIFAR10 using ResNet-32.}
The results from table~\ref{baseline} demonstrate that our \ouralgoshort maintains high accuracy even after the networks are highly compressed on CIFAR-10. Given a well-trained 32-layer \Resnet network and a goal of reducing the number of parameters to 10\% of the original size, the \nCP obtained by \ourbaselineshort using \etoeLong compression reduces the original accuracy from 93.2\% to 86.93\%; while the \tnnCP obtained by \ouralgoshort paired with \seqTune compression increases the accuracy to 91.28\% with the same compression rate --- \textbf{a performance loss of 2\% with only 10\% of the number of parameters}. Furthermore, \ouralgoshort achieves further aggressive compression --- \textbf{a performance loss of 6\% with only 2\% of the number of parameters}.
We observe similar trends (higher compression and higher accuracy) are observed for \tnnTT as well.   
The structure of the \TKlong decomposition (see section~\ref{app:convolutional-tensorized}) makes \tnnTK less effective with very high compression, since the {``internal structure''} of the network reduces to very low rank, which may lose necessary information. Increasing the network size to 20\% of the original provides reasonable performance on CIFAR-10 for \tnnTK as well.

\begin{table*}[!htbp]
\centering
\begin{tabular}{l}
\begin{tabular}{c |c  c  c  c || c | c  c  c  c}
\multirow{2}{3.6em}{Architect.} & \multicolumn{4}{c||}{Compression rate} & \multirow{2}{3.6em}{Architect.} & \multicolumn{4}{c}{Compression rate} \\ 
& \multicolumn{1}{c}{5\%} & \multicolumn{1}{c}{10\%} & \multicolumn{1}{c}{20\%} & \multicolumn{1}{c||}{40\%} & \multicolumn{1}{c|}{} & \multicolumn{1}{c}{2\%} & \multicolumn{1}{c}{5\%} & \multicolumn{1}{c}{10\%} & \multicolumn{1}{c}{20\%} \\
\hline
\nSVD~\cite{jaderberg2014speeding} & 83.09 & 87.27 & 89.58 & 90.85 & TNN-TR{$^\dag$}~\cite{wang2018wide} & - & 80.80{$^\dag$} & - & 90.60 \\
\nCP~\cite{lebedev2014speeding}   & 84.02 & 86.93 & 88.75 & 88.75 & \tnnCP & 85.7 & 89.86 & \textbf{91.28} & - \\
\nTK~\cite{kim2015compression}   & 83.57 & 86.00 & 88.03 & 89.35 & \tnnTK & 61.06 & 71.34 & 81.59 & 87.11 \\
\nTT   & 77.44 & 82.92 & 84.13 & 86.64 & \tnnTT & 78.95 & 84.26 & 87.89 & - \\
\end{tabular} \\
\rule{0in}{1.2em}$^\dag$\scriptsize The tensor ring (TR) results are cited from~\cite{wang2018wide}, and the accuracy of 80.8\% is achieved by 6.67\% compression rate.
\end{tabular}
\vspace{-0.8em}
\caption{Percentage test accuracy of baseline \textbf{\ourbaselineshort} with
\etoe compression vs. our \textbf{\ouralgoshort} with  \seqTune compression on CIFAR10.  
The uncompressed \Resnet-32 achieves {93.2\%} accuracy with 0.46M parameters. 
\label{baseline}}
\end{table*}

\begin{table*}[!htbp]
\centering
\begin{tabular}{c | c  c | c  c | c c | c c}
\multirow{3}{3.6em}{Architect.} & \multicolumn{8}{c}{Compression rate} \\ 
& \multicolumn{2}{c|}{5\%} & \multicolumn{2}{c|}{10\%} & \multicolumn{2}{c|}{20\%} & \multicolumn{2}{c}{40\%} \\
& Seq & E2E & Seq & E2E & Seq & E2E & Seq & E2E \\
\hline
\nSVD & 74.04 & \textbf{83.09} & 85.28 & \textbf{87.27} & \textbf{89.74} & 89.58  & \textbf{91.83} & 90.85 \\
\nCP   & 83.19 & \textbf{84.02} & \textbf{88.50} & 86.93 & \textbf{90.72} & 88.75 & \textbf{89.75} & 88.75 \\
\nTK   & 80.11 & \textbf{83.57} & \textbf{86.75} & 86.00 & \textbf{89.55} & 88.03 & \textbf{91.3} & 89.35 \\
\nTT   & \textbf{80.77} & 77.44 & \textbf{87.08} & 82.92 & \textbf{89.14} & 84.13 & \textbf{91.21} & 86.64\\  
\end{tabular}
\caption{Percentage accuracy of \textbf{our Seq} vs. \textbf{baseline E2E} tuning using \ourbaselineshort on CIFAR10.  
\label{seq-e2e-noreshape}}
\end{table*}

\begin{table*}[!htbp]
\centering
\begin{tabular}{c | c c || c | c c}
\multirow{2}{3.6em}{Architect.} & \multicolumn{2}{c||}{Compression rate} & \multirow{2}{3.6em}{Architect.} & \multicolumn{2}{c}{Compression rate}\\  
& \multicolumn{1}{c}{5\%} & \multicolumn{1}{c||}{10\%} & \multicolumn{1}{c|}{} & \multicolumn{1}{c}{5\%} & \multicolumn{1}{c}{10\%} \\ \hline
\nCP & 83.19 & 88.50 & \tnnCP & \textbf{89.86} & \textbf{91.28} \\
\nTK & 80.11 & 86.73 & \tnnTK & 71.34 & 81.59 \\
\nTT & 80.77 & 87.08 & \tnnTT & \textbf{84.26} & \textbf{87.89} \\
\end{tabular}
\caption{Percentage accuracy of \textbf{our \ouralgoshort} vs.  \textbf{baseline \ourbaselineshort} using Seq tuning on CIFAR10.}
\label{reshape-vs-non}
\end{table*}

\begin{table*}[!htbp]
\begin{center}
\begin{tabular}{c|c||c||c||c}
&  &  {Uncompressed}  &{\nTT} (\etoe) &{\tnnTT} (\seqTune)  \\
\# epochs  & \# samples & \# params. = 25M & \# params. = 2.5M & \# params. = 2.5M \\
\hline
0.2 & 0.24M & 4.22& 2.78 & 44.35  \\
0.3 & 0.36M & 6.23& 3.99 & 46.98   \\
0.5 & 0.60M & 9.01& 7.48 & 49.92  \\
1.0 & 1.20M & 17.3& 12.80 & 52.59   \\
2.0 & 2.40M & 30.8& 18.17 & \textbf{54.00}\\
\end{tabular}		
\caption{Convergence of percentage accuracies of \textbf{uncompressed} vs. \textbf{\ourbaselineshort} (\TT decomposition) vs. \textbf{\ouralgoshort} (\tnnTT decomposition) achieving 10\% compression rate for \Resnet-50 ImageNet.
\label{imagenet}}
\end{center}
\end{table*}

\begin{figure}[!htbp]
\begin{minipage}[]{\linewidth}
		\centering
		\psfrag{y-axis}{\scriptsize{Test Error}}
		\psfrag{x-axis}[Bl]{\scriptsize{\# Gradient Updates ($\times 10^{11}$)}}
		\psfrag{cp-seq}[Bl]{\scriptsize{Seq-\CP}}
		\psfrag{cp-e2e}[Bl]{\scriptsize{E2E-\CP}}
		\psfrag{tt-seq}[Bl]{\scriptsize{Seq-\TT}}
		\psfrag{tt-e2e}[Bl]{\scriptsize{E2E-\TT}}
		\psfrag{tk-e2e}[Bl]{\scriptsize{E2E-\TK}}
		\psfrag{tk-seq}[Bl]{\scriptsize{Seq-\TK}}
		\includegraphics[width=0.8\textwidth]{\fighome/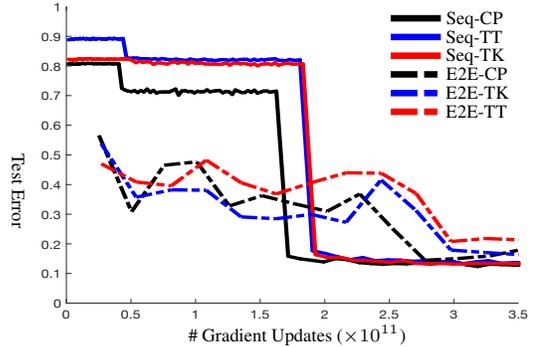}
		\captionof{figure}{\label{convergence}Convergence rate for Seq vs. \etoe compression on CIFAR10.}
	\end{minipage}
\end{figure}

Table~\ref{baseline} shows that \emph{\ouralgoshort with \seqTune compression} outperforms \emph{\ourbaselineshort with \etoeLong compression}.  Now we address the following question: is one factor (\seqTune compression or \ouralgoshort) primarily responsible for increased performance, or is the benefit due to synergy between the two?

\smallskip 

\noindent \textbf{\seqTune compression, \ouralgo, or both?} (1) 
We present the effect of different compression methods on accuracy in Table~\ref{seq-e2e-noreshape}.  
Other than at very high compression rate (5\% column in
Table~\ref{seq-e2e-noreshape}), \seqTune compression (Seq) consistently
outperforms \etoeLong (E2E) compression. In addition, \seqTune compression is also
{\em much\/} faster and leads to more stable convergence compared to \etoeLong compression.
Figure~\ref{convergence} plots the compression error over the number of gradient
updates for various compression methods.
(2) We present the effect of different compression methods on accuracy in Table~\ref{reshape-vs-non}.
Interestingly, as what is demonstrated in Table~\ref{reshape-vs-non}, if \ouralgo method is used, the test accuracy is restored for even very high compression ratios~\footnote{Note that \tnnTK remains an exception for aggressive compression due to the low rank internal structure that we previously discussed.}.  
These results confirm that \tensornetshort is more flexible than NN as \tensornetshort allows exploitation of additional invariant structures in the parameter space of deep neural networks, and such invariant structures are picked up by our proposed \ouralgo (our \ouralgoshort), but not by \ourbaseline (\ourbaselineshort). 
Therefore, our results show that \ouralgoshort and \seqTune compression are
symbiotic, and {\em both\/} are necessary to simultaneously obtain a
high accuracy and a high compression rate.

\paragraph{Scalability}
Finally, we show that our methods scale to state-of-the-art large
networks, by evaluating performance on the ImageNet 2012 dataset on a
50-layer \Resnet (uncompressed with 76.05\%  accuracy).
Table~\ref{imagenet} shows the accuracy of \tnnTT obtained by \ouralgoshort
with \seqTune tuning compared to that of \nTT obtained by \ourbaseline with \etoe tuning, and the accuracy of the uncompressed network (\Resnet-50) with 10\% compression rate.   
Table~\ref{imagenet} shows that \ouralgoshort paired with Seq tuning is much faster than the alternative. This is an important
result because it empirically validates our hypotheses that (1)
our \ouralgo captures the invariant structure of the \Resnet (with few redundancies) better and faster than the baseline \ourbaselineshort compression, 
(2) data reconstruction \seqTune tuning is effective even on the largest networks and datasets, and 
(3) our proposed efficient \ouralgoshort can scale to the state-of-the-art neural networks. 

\section{Conclusion and Perspectives}
\label{sec:conclusion}
We define a new generalized tensor algebra extending existing tensor operations.  We extend vector/matrix operations to their
higher order tensor counterparts, providing systematic notations and libraries for
tensorization of neural networks and higher order tensor
decompositions. 
Using these generalized tensor operations, we introduce \tensornet (\tensornetshort) which extends existing neural networks. 
Our TNN is more flexible than NN and more compact than neural networks allowing same amount expressive power with fewer number of parameters. 
Therefore mapping NN to its closest TNN is a compression of NN as the resulting TNN will carry fewer number of parameters. 
Other compression techniques as mentioned in the related work can naturally be used on the compressed TNN to further compress the TNN.  
 As a future step, we will explore optimizing the order of (parallel) implementations of the tensor algebra.


\bibliographystyle{named}
\bibliography{supp_bib}

\begin{thebibliography}{}

\bibitem[\protect\citeauthoryear{Ba and Caruana}{2014}]{ba2014deep}
Jimmy Ba and Rich Caruana.
\newblock Do deep nets really need to be deep?
\newblock In {\em Advances in neural information processing systems}, pages
  2654--2662, 2014.

\bibitem[\protect\citeauthoryear{Cheng \bgroup \em et al.\egroup
  }{2015}]{cheng2015exploration}
Yu~Cheng, Felix~X Yu, Rogerio~S Feris, Sanjiv Kumar, Alok Choudhary, and Shi-Fu
  Chang.
\newblock An exploration of parameter redundancy in deep networks with
  circulant projections.
\newblock In {\em Proceedings of the IEEE International Conference on Computer
  Vision}, pages 2857--2865, 2015.

\bibitem[\protect\citeauthoryear{Cheng \bgroup \em et al.\egroup
  }{2017}]{cheng2017survey}
Yu~Cheng, Duo Wang, Pan Zhou, and Tao Zhang.
\newblock A survey of model compression and acceleration for deep neural
  networks.
\newblock {\em arXiv preprint arXiv:1710.09282}, 2017.

\bibitem[\protect\citeauthoryear{Chollet}{2016}]{chollet2016xception}
Fran{\c{c}}ois Chollet.
\newblock Xception: Deep learning with depthwise separable convolutions.
\newblock {\em arXiv preprint arXiv:1610.02357}, 2016.

\bibitem[\protect\citeauthoryear{Cichocki \bgroup \em et al.\egroup
  }{2016}]{cichocki2016low}
Andrzej Cichocki, Namgil Lee, Ivan~V Oseledets, Anh~Huy Phan, Qibin Zhao, and
  D~Mandic.
\newblock Low-rank tensor networks for dimensionality reduction and large-scale
  optimization problems: Perspectives and challenges part 1.
\newblock {\em arXiv preprint arXiv:1609.00893}, 2016.

\bibitem[\protect\citeauthoryear{Cichocki \bgroup \em et al.\egroup
  }{2017}]{cichocki2017tensor}
Andrzej Cichocki, Anh-Huy Phan, Qibin Zhao, Namgil Lee, Ivan Oseledets, Masashi
  Sugiyama, Danilo~P Mandic, et~al.
\newblock Tensor networks for dimensionality reduction and large-scale
  optimization: Part 2 applications and future perspectives.
\newblock {\em Foundations and Trends{\textregistered} in Machine Learning},
  9(6):431--673, 2017.

\bibitem[\protect\citeauthoryear{Comon \bgroup \em et al.\egroup
  }{2009}]{comon2009tensor}
Pierre Comon, Xavier Luciani, and Andr{\'e}~LF De~Almeida.
\newblock Tensor decompositions, alternating least squares and other tales.
\newblock {\em Journal of Chemometrics: A Journal of the Chemometrics Society},
  23(7-8):393--405, 2009.

\bibitem[\protect\citeauthoryear{Courbariaux \bgroup \em et al.\egroup
  }{2015}]{courbariaux2015binaryconnect}
Matthieu Courbariaux, Yoshua Bengio, and Jean-Pierre David.
\newblock Binaryconnect: Training deep neural networks with binary weights
  during propagations.
\newblock In {\em Advances in neural information processing systems}, pages
  3123--3131, 2015.

\bibitem[\protect\citeauthoryear{Denton \bgroup \em et al.\egroup
  }{2014}]{denton2014exploiting}
Emily~L Denton, Wojciech Zaremba, Joan Bruna, Yann LeCun, and Rob Fergus.
\newblock Exploiting linear structure within convolutional networks for
  efficient evaluation.
\newblock In {\em Advances in neural information processing systems}, pages
  1269--1277, 2014.

\bibitem[\protect\citeauthoryear{Furlanello \bgroup \em et al.\egroup
  }{2018}]{furlanello2018born}
Tommaso Furlanello, Zachary~C Lipton, Michael Tschannen, Laurent Itti, and
  Anima Anandkumar.
\newblock Born again neural networks.
\newblock {\em arXiv preprint arXiv:1805.04770}, 2018.

\bibitem[\protect\citeauthoryear{Grasedyck \bgroup \em et al.\egroup
  }{2013}]{grasedyck2013literature}
Lars Grasedyck, Daniel Kressner, and Christine Tobler.
\newblock A literature survey of low-rank tensor approximation techniques.
\newblock {\em GAMM-Mitteilungen}, 36(1):53--78, 2013.

\bibitem[\protect\citeauthoryear{Han \bgroup \em et al.\egroup
  }{2015}]{han2015deep}
Song Han, Huizi Mao, and William~J Dally.
\newblock Deep compression: Compressing deep neural networks with pruning,
  trained quantization and huffman coding.
\newblock {\em arXiv preprint arXiv:1510.00149}, 2015.

\bibitem[\protect\citeauthoryear{He \bgroup \em et al.\egroup
  }{2016a}]{he2016deep}
Kaiming He, Xiangyu Zhang, Shaoqing Ren, and Jian Sun.
\newblock Deep residual learning for image recognition.
\newblock In {\em Proceedings of the IEEE conference on computer vision and
  pattern recognition}, pages 770--778, 2016.

\bibitem[\protect\citeauthoryear{He \bgroup \em et al.\egroup
  }{2016b}]{he2016identity}
Kaiming He, Xiangyu Zhang, Shaoqing Ren, and Jian Sun.
\newblock Identity mappings in deep residual networks.
\newblock In {\em European Conference on Computer Vision}, pages 630--645.
  Springer, 2016.

\bibitem[\protect\citeauthoryear{Hinton \bgroup \em et al.\egroup
  }{2015}]{hinton2015distilling}
Geoffrey Hinton, Oriol Vinyals, and Jeff Dean.
\newblock Distilling the knowledge in a neural network.
\newblock {\em arXiv preprint arXiv:1503.02531}, 2015.

\bibitem[\protect\citeauthoryear{Huang \bgroup \em et al.\egroup
  }{2017}]{huang2017densely}
Gao Huang, Zhuang Liu, Laurens Van Der~Maaten, and Kilian~Q Weinberger.
\newblock Densely connected convolutional networks.
\newblock In {\em CVPR}, volume~1, page~3, 2017.

\bibitem[\protect\citeauthoryear{Hubara \bgroup \em et al.\egroup
  }{2017}]{hubara2017quantized}
Itay Hubara, Matthieu Courbariaux, Daniel Soudry, Ran El-Yaniv, and Yoshua
  Bengio.
\newblock Quantized neural networks: Training neural networks with low
  precision weights and activations.
\newblock {\em Journal of Machine Learning Research}, 18:187--1, 2017.

\bibitem[\protect\citeauthoryear{Jaderberg \bgroup \em et al.\egroup
  }{2014}]{jaderberg2014speeding}
Max Jaderberg, Andrea Vedaldi, and Andrew Zisserman.
\newblock Speeding up convolutional neural networks with low rank expansions.
\newblock {\em arXiv preprint arXiv:1405.3866}, 2014.

\bibitem[\protect\citeauthoryear{Kim \bgroup \em et al.\egroup
  }{2015}]{kim2015compression}
Yong-Deok Kim, Eunhyeok Park, Sungjoo Yoo, Taelim Choi, Lu~Yang, and Dongjun
  Shin.
\newblock Compression of deep convolutional neural networks for fast and low
  power mobile applications.
\newblock {\em arXiv preprint arXiv:1511.06530}, 2015.

\bibitem[\protect\citeauthoryear{Kingma and Ba}{2014}]{kingma2014adam}
Diederik~P Kingma and Jimmy Ba.
\newblock Adam: A method for stochastic optimization.
\newblock {\em arXiv preprint arXiv:1412.6980}, 2014.

\bibitem[\protect\citeauthoryear{Kolda and Bader}{2009}]{kolda2009tensor}
Tamara~G Kolda and Brett~W Bader.
\newblock Tensor decompositions and applications.
\newblock {\em SIAM review}, 51(3):455--500, 2009.

\bibitem[\protect\citeauthoryear{Krizhevsky \bgroup \em et al.\egroup
  }{2012}]{krizhevsky2012imagenet}
Alex Krizhevsky, Ilya Sutskever, and Geoffrey~E Hinton.
\newblock Imagenet classification with deep convolutional neural networks.
\newblock In {\em Advances in neural information processing systems}, pages
  1097--1105, 2012.

\bibitem[\protect\citeauthoryear{Lebedev \bgroup \em et al.\egroup
  }{2014}]{lebedev2014speeding}
Vadim Lebedev, Yaroslav Ganin, Maksim Rakhuba, Ivan Oseledets, and Victor
  Lempitsky.
\newblock Speeding-up convolutional neural networks using fine-tuned
  cp-decomposition.
\newblock {\em arXiv preprint arXiv:1412.6553}, 2014.

\bibitem[\protect\citeauthoryear{LeCun \bgroup \em et al.\egroup
  }{1998}]{lecun1998gradient}
Yann LeCun, L{\'e}on Bottou, Yoshua Bengio, and Patrick Haffner.
\newblock Gradient-based learning applied to document recognition.
\newblock {\em Proceedings of the IEEE}, 86(11):2278--2324, 1998.

\bibitem[\protect\citeauthoryear{Lin \bgroup \em et al.\egroup
  }{2013}]{lin2013network}
Min Lin, Qiang Chen, and Shuicheng Yan.
\newblock Network in network.
\newblock {\em arXiv preprint arXiv:1312.4400}, 2013.

\bibitem[\protect\citeauthoryear{Novikov \bgroup \em et al.\egroup
  }{2015}]{novikov2015tensorizing}
Alexander Novikov, Dmitrii Podoprikhin, Anton Osokin, and Dmitry~P Vetrov.
\newblock Tensorizing neural networks.
\newblock In {\em Advances in Neural Information Processing Systems}, pages
  442--450, 2015.

\bibitem[\protect\citeauthoryear{Or{\'u}s}{2014}]{orus2014practical}
Rom{\'a}n Or{\'u}s.
\newblock A practical introduction to tensor networks: Matrix product states
  and projected entangled pair states.
\newblock {\em Annals of Physics}, 349:117--158, 2014.

\bibitem[\protect\citeauthoryear{Oseledets}{2011}]{oseledets2011tensor}
Ivan~V Oseledets.
\newblock Tensor-train decomposition.
\newblock {\em SIAM Journal on Scientific Computing}, 33(5):2295--2317, 2011.

\bibitem[\protect\citeauthoryear{Rastegari \bgroup \em et al.\egroup
  }{2016}]{rastegari2016xnor}
Mohammad Rastegari, Vicente Ordonez, Joseph Redmon, and Ali Farhadi.
\newblock Xnor-net: Imagenet classification using binary convolutional neural
  networks.
\newblock In {\em European Conference on Computer Vision}, pages 525--542.
  Springer, 2016.

\bibitem[\protect\citeauthoryear{Romero \bgroup \em et al.\egroup
  }{2014}]{romero2014fitnets}
Adriana Romero, Nicolas Ballas, Samira~Ebrahimi Kahou, Antoine Chassang, Carlo
  Gatta, and Yoshua Bengio.
\newblock Fitnets: Hints for thin deep nets.
\newblock {\em arXiv preprint arXiv:1412.6550}, 2014.

\bibitem[\protect\citeauthoryear{Simonyan and
  Zisserman}{2014}]{simonyan2014very}
Karen Simonyan and Andrew Zisserman.
\newblock Very deep convolutional networks for large-scale image recognition.
\newblock {\em arXiv preprint arXiv:1409.1556}, 2014.

\bibitem[\protect\citeauthoryear{Sindhwani \bgroup \em et al.\egroup
  }{2015}]{sindhwani2015structured}
Vikas Sindhwani, Tara Sainath, and Sanjiv Kumar.
\newblock Structured transforms for small-footprint deep learning.
\newblock In {\em Advances in Neural Information Processing Systems}, pages
  3088--3096, 2015.

\bibitem[\protect\citeauthoryear{Szegedy \bgroup \em et al.\egroup
  }{2015}]{szegedy2015going}
Christian Szegedy, Wei Liu, Yangqing Jia, Pierre Sermanet, Scott Reed, Dragomir
  Anguelov, Dumitru Erhan, Vincent Vanhoucke, and Andrew Rabinovich.
\newblock Going deeper with convolutions.
\newblock In {\em Proceedings of the IEEE conference on computer vision and
  pattern recognition}, pages 1--9, 2015.

\bibitem[\protect\citeauthoryear{Szegedy \bgroup \em et al.\egroup
  }{2017}]{szegedy2017inception}
Christian Szegedy, Sergey Ioffe, Vincent Vanhoucke, and Alexander~A Alemi.
\newblock Inception-v4, inception-resnet and the impact of residual connections
  on learning.
\newblock In {\em AAAI}, volume~4, page~12, 2017.

\bibitem[\protect\citeauthoryear{Wang and Lu}{2017}]{wang2017tensor}
Po-An Wang and Chi-Jen Lu.
\newblock Tensor decomposition via simultaneous power iteration.
\newblock In {\em International Conference on Machine Learning}, pages
  3665--3673, 2017.

\bibitem[\protect\citeauthoryear{Wang \bgroup \em et al.\egroup
  }{2018}]{wang2018wide}
Wenqi Wang, Yifan Sun, Brian Eriksson, Wenlin Wang, and Vaneet Aggarwal.
\newblock Wide compression: Tensor ring nets.
\newblock {\em learning}, 14(15):13--31, 2018.

\bibitem[\protect\citeauthoryear{Yang \bgroup \em et al.\egroup
  }{2015}]{yang2015deep}
Zichao Yang, Marcin Moczulski, Misha Denil, Nando de~Freitas, Alex Smola,
  Le~Song, and Ziyu Wang.
\newblock Deep fried convnets.
\newblock In {\em Proceedings of the IEEE International Conference on Computer
  Vision}, pages 1476--1483, 2015.

\bibitem[\protect\citeauthoryear{Zhu \bgroup \em et al.\egroup
  }{2016}]{zhu2016trained}
Chenzhuo Zhu, Song Han, Huizi Mao, and William~J Dally.
\newblock Trained ternary quantization.
\newblock {\em arXiv preprint arXiv:1612.01064}, 2016.

\end{thebibliography}

\onecolumn
\newpage
\appendix
\begin{center}{\Large Appendix: \mytitle}\end{center}

\section{Supplementary experiments}
\label{app:supp_experiments}

\paragraph{Convergence Rate}
Compared to \etoeLong, an ancillary benefit of \seqTrain tuning is
{\em much} faster and leads to more stable convergence.
Figure~\ref{convergence} plots compression error over number of gradient
updates for various methods.  (This experiment is for \ourbaselineshort with 10\% compression rate.)
There are three salient points: first, \seqTrain tuning has very
high error in the beginning while the ''early'' blocks of the
network are being tuned (and the rest of the network is left
unchanged to {tensor decomposition values}).
However, as the final block is tuned
(around $2\times 10^{11}$ gradient updates) in the figure, the errors drop
to nearly minimum immediately. In comparison, \etoeLong tuning
requires 50--100\% more gradient updates to achieve stable
performance. 
Finally, the result also shows that for each block,
\seqTrain tuning achieves convergence very quickly (and nearly
monotonically), which results in the stair-step pattern since extra
tuning of a block does not improve (or appreciably reduce)
performance.  

\paragraph{Performance on Fully-Connected Layers} 
An extra advantage of \ouralgo is that it can apply flexibly to
fully-connected as well as convolutional layers of a neural network.
Table~\ref{table:exp-dense} shows the results of applying \ouralgo to
various tensor decompositions on a variant of LeNet-5
network~\cite{lecun1998gradient}.  The convolutional layers of the
LeNet-5 network were {\em not} compressed, trained or updated in
these experiments.  The uncompressed network achieves 99.31\% accuracy.
Table~\ref{table:exp-dense} shows \textbf{the fully-connected layers can be compressed
to 0.2\% losing only about 2\% accuracy}.  In fact, compressing the
dense layers to 1\% of their original size reduce accuracy by less
than 1\%, demonstrating the extreme efficacy of \ouralgo when
applied to fully-connected neural network layers.

\begin{table}[!htbp]
\centering
\begin{tabular}{ c | c | c | c }
& \multicolumn{3}{c}{Compression rate} \\ 
Architect.  & 0.2\% & 0.5\% & 1\% \\ 
\hline
\tnnCP & 97.21 & 97.92 & 98.65 \\
\tnnTK &  97.71 & 98.56 & 98.52 \\
\tnnTT & 97.69 & 98.43 & 98.63 \\
\end{tabular}
\caption{TNN compression of fully-connected layer in LeNet-5. 
The uncompressed network achieves 99.31\% accuracy.}
\label{table:exp-dense}
\end{table}

\section{Notations}
\label{app:notations}

\paragraph{Symbols:}
Lower case letters (e.g. $\myvector{v}$) are used to denote column vectors, while upper case letters (e.g. $\mymatrix{M}$) are used for matrices, and curled letters (e.g. $\tensor{T}$) for multi-dimensional arrays (tensors). For a tensor $\tensor{T} \in \R^{I_0 \times \cdots \times I_{m-1}}$, we will refer to the number of indices as \textit{order}, each individual index as \textit{mode} and the length at one mode as \textit{dimension}. Therefore, we will say that $\tensor{T} \in \R^{I_0 \times \cdots \times I_{m-1}}$ is an $m$-order tensor which has dimension $I_k$ at mode-$k$. {\em Tensor operations} are extensively used in this paper: The {\em tensor (partial) outer product} is denoted as $\otimes$, {\em tensor convolution} as $\ast$, and finally $\times$ denotes either {\em tensor contraction} or {\em tensor multiplication}. Each of these operators will be equipped with subscript and superscript when used in practice, for example $\times^{m}_{n}$ denotes mode-$(m, n)$ tensor contraction (defined in Appendix~\ref{app:operations}). Furthermore, the symbol $\circ$ is used to construct {\em compound operations}. For example, $(\ast \circ \otimes)$ is a compound operator simultaneously performing tensor convolution and tensor partial outer product between two tensors.     

\paragraph{Indexing:} In this paragraph, we explain the usages of subscripts/superscripts for both multi-dimensional arrays and operators, and further introduce several functions that are used to alter the layout of multi-dimensional arrays. 

\begin{itemize}[leftmargin=*]

\item 
(1) Nature indices start from 0, but reversed indices are used occasionally, which start from $-1$. Therefore the first entry of a vector $\myvector{v}$ is ${v}_{0}$, while the last one is ${v}_{-1}$. (2) For multi-dimensional arrays, the {\em subscript} is used to denote an entry or a subarray within an object, while {\em superscript} is to index among a sequence of arrays. For example, $\mymatrix{M}_{i, j}$ denotes the entry at $i^{\tha}$ row and $j^{\tha}$ column of a matrix $\mymatrix{M}$, and $\matrixSup{M}{k}$ is the $k^{\tha}$ matrix in a set of $N$ matrices $\{ \matrixSup{M}{0}, \matrixSup{M}{1}, \cdots \matrixSup{M}{N-1} \}$. For operators, as we have seen, both subscript and superscript are used to denote the modes involved in the operation. (3) {\em The symbol colon '$:$'} is used to slice a multi-dimensional array. For example, $\tensorSub{M}{:, k}$ denotes the $k^{\tha}$ column of $\mymatrix{M}$, and $\tensorSub{T}{:, :, k}$ denotes the $k^{\tha}$ {\em frontal slice} of a 3-order tensor $\tensor{T}$. (4) {\em Big-endian notation} is adopted in conversion between multi-dimensional array and vectors. Specifically, the function $\vectorize(\cdot)$ flattens (a.k.a. {\em vectorize}) a tensor $\tensor{T} \in \R^{I_0 \times \cdots \times I_{m-1}}$ into a vector $\myvector{v} \in \R^{\prod_{l=0}^{m-1} I_l}$ such that $\tensorSub{T}{i_0, \cdots, i_{m-1}} = v_{i_{m-1} + i_{m-2} I_{m-1} + \cdots + i_0 I_1 \cdots I_{m-1}}$.
 
\item 
(1) The function $\swapaxes(\cdot)$ is used to permute ordering of the modes of a tensor as needed. For example, given two tensors $\tensor{U} \in \R^{I \times J \times K}$ and $\tensor{V} \in \R^{K \times J \times I}$, the operation $\tensor{V} = \swapaxes(\tensor{U})$ convert the tensor $\tensor{U}$ into $\tensor{V}$ such that $\tensorSub{V}{k, j, i} = \tensorSub{U}{i, j, k}$.
(2) The function $\flipaxis(\cdot, \cdot)$ flips a tensor along a given mode. For example, given a tensor $\tensor{U} \in \R^{I \times J \times K}$ and $\tensor{V} = \flipaxis(\tensor{U}, 0)$, the entries in $\tensor{V}$ is defined as $\tensorSub{V}{i, j, k} = \tensorSub{U}{I - 1 - i (mod ~ I), j, k}$.

\end{itemize}

\section{Tensor operations}
\label{app:operations}


\begin{table*}[!htbp]
\centering
\begin{tabular}{ l | l | l }
Operator & Notation & Definition \\
\hline

\hspace{-0.5em}\begin{tabular}{l}mode-($k, l$) \hspace{-2em} \\Tensor\hspace{-1em} \\Contraction\hspace{-1em}\end{tabular} 	\hspace{-1em} & \hspace{-0.5em} $\tensorSup{T}{0} = \tensor{X} \times^{k}_{l} \tensor{Y}$\hspace{-0.5em} & \hspace{-1em} \begin{tabular}{l} $\tensorInd{T}{i_0, \cdots, i_{k-1}, i_{k+1}, \cdots, i_{m-1}, j_0, \cdots, j_{l-1}, j_{l+1}, \cdots, j_{n-1}}{0} $\\$= \big< \tensorSub{X}{i_0, \cdots, i_{k-1},{ :}, i_{k+1}, \cdots, i_{m-1}},  \tensorSub{Y}{j_0, \cdots, j_{l-1}, {:}, j_{l+1}, \cdots, j_{n-1}}\big>$\\ \small{inner product of mode-$k$ fiber of $\tensor{X}$} \\ \small{and mode-$l$ fiber of $\tensor{Y}$} \end{tabular}\\
\hline

\hspace{-0.5em}\begin{tabular}{l}mode-$k$ \hspace{-2em} \\Tensor\hspace{-1em} \\Multiplication\hspace{-1em}\end{tabular} \hspace{-1em} & \hspace{-0.5em} $\tensorSup{T}{1} = \tensor{X} \times_k \mymatrix{M}$\hspace{-0.5em} & \hspace{-1em} \begin{tabular}{l}$ \tensorInd{T}{i_0, \cdots, i_{k-1}, {r}, i_{k+1}, \cdots, i_{m-1}}{1} $\\$= \big< \tensorSub{X}{i_0, \cdots, i_{k-1}, {:}, i_{k+1}, \cdots, i_{m-1}}, \mymatrix{M}_{{:},{ r}}\big>$\\ \small{ inner product of mode-$k$ fiber of $\tensor{X}$} \\ \small{and $r^{\tha}$ column of $\mymatrix{M}$} \end{tabular}\\
\hline

\hspace{-0.5em}\begin{tabular}{l}mode-($k, l$)\hspace{-2em}  \\Tensor\hspace{-1em} \\Convolution\hspace{-1em}\end{tabular}	\hspace{-1em}&\hspace{-0.5em} $\tensorSup{T}{2} = \tensor{X} \ast^{k}_{l} \tensor{Y}$\hspace{-0.5em}	& \hspace{-1em} \begin{tabular}{l}$\tensorInd{T}{i_0, \cdots, i_{k-1}, {:}, i_{k+1}, \cdots, i_{m-1}, j_0, \cdots, j_{l-1}, j_{l+1}, \cdots, j_{n-1}}{2} $\\$= \tensorSub{X}{i_0, \cdots, i_{k-1}, {:}, i_{k+1}, \cdots, i_{m-1}} \ast \tensorSub{Y}{j_0, \cdots, j_{l-1},{ :}, j_{l+1}, \cdots, j_{n-1}}$\\ \small{convolution of mode-$k$ fiber of $\tensor{X}$} \\ \small{and mode-$l$ fiber of $\tensor{Y}$}\end{tabular}\\
\hline

\hspace{-0.5em}\begin{tabular}{l}mode-($k, l$) \hspace{-2em} \\Partial-\hspace{-1em}\\Outer Product\hspace{-1em}\end{tabular} \hspace{-1em}&\hspace{-0.5em}	$\tensorSup{T}{3} = \tensor{X} \otimes^{k}_{l} \tensor{Y}$\hspace{-0.5em}	& \hspace{-1em} \begin{tabular}{l} $\tensorInd{T}{i_0, \cdots, i_{k-1}, {r}, i_{k+1}, \cdots, i_{m-1}, j_0, \cdots,  j_{n-1}}{3} $ \\$=\tensorSub{X}{i_0, \cdots, i_{k-1}, {r}, i_{k+1}, \cdots, i_{m-1}} ~ \tensorSub{Y}{j_0, \cdots, j_{l-1},{ r}, j_{l+1}, \cdots, j_{n-1}}$\\ \footnotesize{Hadamard product of mode-$k$ fiber of $\tensor{X}$} \\ \small{and mode-$l$ fiber of $\tensor{Y}$}\end{tabular} 

\end{tabular}
\caption{\textbf{Summary of tensor operations}. In this table, $\tensor{X} \in \R^{I_0 \times \cdots \times I_{m-1}}$, $\tensor{Y} \in \R^{J_0 \times \cdots \times J_{n-1}}$ and matrix $\mymatrix{M} \in \R^{I_k \times J}$. Mode-$(k, l)$ tensor contraction and mode-($k, l$) tensor partial-outer product are legal only if $I_k = J_l$. $\tensorSup{T}{0}$ is an $(m+n-2)$-order tensor,  $\tensorSup{T}{1}$ is an $m$-order tensor, $\tensorSup{T}{2}$ is an $(m+n-1)$-order tensor and $\tensorSup{T}{3}$ is an $(m+n-1)$-order tensor.}\label{tab:tensorOperations}
\end{table*}

To begin with, we describe several {\em basic tensor operations} that are natural generalization to their vector/matrix counterparts. These basic operations can be further combined to construct {\em compound operations} that serve as building blocks of {\em tensorial neural networks}. 

\paragraph{Tensor contraction} 
Given an $m$-order tensor $\tensorSup{T}{0} \in \R^{I_0 \times \cdots \times I_{m-1}}$ and another $n$-order tensor $\tensorSup{T}{1} \in \R^{J_0 \times \cdots \times J_{n-1}}$, which share the same dimension at mode-$k$ of $\tensorSup{T}{0}$ and mode-$l$ of $\tensorSup{T}{1}$( i.e. $I_k = J_l$), the mode-$(k, l)$ contraction of $\tensorSup{T}{0}$ and $\tensorSup{T}{1}$, denoted as $\tensor{T} \triangleq \tensorSup{T}{0} \times^{k}_{l} \tensorSup{T}{1}$, returns a $(m + n - 2)$-order tensor $\tensor{T}\ \in R^{I_0 \times \cdots \times I_{k-1} \times I_{k+1} \times \cdots \times I_{m-1} \times J_0 \times \cdots \times J_{l-1} \times J_{l+1} \times \cdots \times J_{n-1}}$, whose entries are computed as
\begin{subequations}
\begin{align}
& \tensorSub{T}{i_0, \cdots, i_{k-1}, i_{k+1}, \cdots, i_{m-1}, j_0, \cdots, j_{l-1}, j_{l+1}, \cdots, j_{n-1}} \nonumber \\
= ~ & \sum_{r=0}^{I_k - 1} \tensorInd{T}{i_0, \cdots, i_{k-1}, r, i_{k+1}, \cdots, i_{m-1}}{0} ~ \tensorInd{T}{j_0, \cdots, j_{l-1}, r, j_{l+1}, \cdots, j_{n-1}}{1} \label{def:tensor-contraction-1} \\
= ~ & \langle  \tensorInd{T}{i_0, \cdots, i_{k-1}, :, i_{k+1}, \cdots, i_{m-1}}{0}, \tensorInd{T}{j_0, \cdots, j_{l-1}, :, j_{l+1}, \cdots, j_{n-1}}{1} \rangle \label{def:tensor-contraction-2}
\end{align}
\end{subequations}
Notice that tensor contraction is a direct generalization of matrix multiplication to higher-order tensor, and it reduces to matrix multiplication if both tensors are $2$-order (and therefore matrices). As each entry in $\tensor{T}$ can be computed as inner product of two vectors, which requires $I_k = J_l$ multiplications, the total number of operations to evaluate a tensor contraction is therefore $O( ( \prod_{u = 0}^{m - 1} I_u ) ( \prod_{v = 0, v \neq l}^{n - 1} J_v ) )$, taking additions into account. 

\paragraph{Tensor multiplication (Tensor product)} 
Tensor multiplication (a.k.a. tensor product) is a special case of tensor contraction where the second operant is a matrix. Given a $m$-order tensor $\tensor{U} \in \R^{I_0 \times \cdots \times I_{m-1}}$ and a matrix $\mymatrix{M} \in \R^{I_k \times J}$, where the dimension of $\tensor{U}$ at mode-$k$ agrees with the number of the rows in $\mymatrix{M}$, the mode-$k$ tensor multiplication of $\tensor{U}$ and $\myvector{M}$, denoted as $\tensor{V} \triangleq \tensor{U} \times_k \mymatrix{M}$, yields another $m$-order tensor $\tensor{V} \in \R^{I_0 \times \cdots \times I_{k-1} \times J \times I_{k+1} \times \cdots I_{m-1}}$, whose entries are computed as
\begin{subequations}
\begin{align}
& \tensorSub{V}{i_0, \cdots, i_{k-1}, j, i_{k+1}, \cdots, i_{m-1}} \nonumber \\
= ~ & \sum_{r=0}^{I_k - 1} \tensorSub{U}{i_0, \cdots, i_{k-1}, r, i_{k+1}, \cdots, i_{m-1}} ~ \mymatrix{M}_{r, j} \label{def:tensor-multiplication-1} \\
= ~ & \langle \tensorSub{U}{i_0, \cdots, i_{k-1}, :, i_{k+1}, \cdots, i_{m-1}}, ~ \mymatrix{M}_{:, j} \rangle \label{def:tensor-multiplication-2}
\end{align}
\end{subequations}
Following the convention of multi-linear algebra, the mode for $J$ now substitutes the location originally for $I_k$ (which is different from the definition of tensor contraction). Regardlessly, the number of operations for tensor multiplication follows tensor contraction exactly, that is $O ( ( \prod_{u = 0}^{m - 1} I_u ) J )$.

\paragraph{Tensor convolution}
Given a $m$-order tensor $\tensorSup{T}{0} \in \R^{I_0 \times I_1 \times \cdots \times I_{m-1}}$ and another $n$-order tensor $\tensorSup{T}{1} \in \R^{J_0 \times J_1 \times \cdots \times J_{n-1}}$. The mode-$(k, l)$ convolution of $\tensorSup{T}{0}$ and $\tensorSup{T}{1}$, denoted as $\tensor{T} \triangleq \tensorSup{T}{0} \ast^{k}_{l} \tensorSup{T}{1}$, returns a $(m + n - 1)$-order tensor $\tensor{T} \in \R^{I_0 \times \cdots \times I^{\prime}_k \times \cdots \times I_{m-1} \times J_0 \times \cdots \times J_{l-1} \times J_{l+1} \times \cdots \times J_{n-1}}$. The entries of $\tensor{T}$ can be computed using any convolution operation $\ast$ that is defined for two vectors.
\begin{subequations}
\begin{align}
& \tensorSub{T}{i_0, \cdots, i_{k-1}, :, i_{k+1}, \cdots, i_{m-1}, j_0, \cdots, j_{l-1}, j_{l+1}, \cdots, j_{n-1}} \nonumber \\
= ~ & \tensorInd{T}{i_0, \cdots, i_{k-1}, :, i_{k+1}, \cdots, i_{m-1}}{0} \ast \tensorInd{T}{j_0, \cdots, j_{l-1}, :, j_{l+1}, \cdots, j_{n-1}}{1} \label{def:tensor-convolution} \\
= ~ & \tensorInd{T}{j_0, \cdots, j_{l-1}, :, j_{l+1}, \cdots, j_{n-1}}{1} ~\overline{\ast}~ \tensorInd{T}{i_0, \cdots, i_{k-1}, :, i_{k+1}, \cdots, i_{m-1}}{0} \label{def:tensor-convolution-reversed}
\end{align}
\end{subequations}
Here we deliberately omit the exact definition of vector convolution $\ast$ (and its conjugate $\overline{\ast}$), because it can be defined differently depending on the user case (Interestingly, the "convolution" in convolutional layer indeed computes {\em correlation} instead of convolution).
Correspondingly, the resulted dimension $I^{\prime}_k$ at mode-$k$ is determined by the chosen type of convolution. 
 For example, the "convolution" in convolutional layer typically yields $I^{\prime}_k = I_k$ (with same padding) or $I^{\prime}_k = I_k - J_l + 1$ (with valid padding). 
Without {\em Fast Fourier Transform (FFT)}, the number of operations is $O ( ( \prod_{u = 0}^{m - 1} I_u ) ( \prod_{v = 0}^{n - 1} J_v ) )$.

\paragraph{Tensor outer product}
Given a $m$-order tensor $\tensorSup{T}{0} \in \R^{I_0 \times I_1 \times \cdots \times I_{m-1}}$ and another $n$-order tensor $\tensorSup{T}{1} \in \R^{J_0 \times J_1 \times \cdots \times J_{n-1}}$, the outer product of $\tensorSup{T}{0}$ and $\tensorSup{T}{1}$, denoted $\tensor{T} \triangleq \tensorSup{T}{0} \otimes \tensorSup{T}{1}$, concatenates all the indices of $\tensorSup{T}{0}$ and $\tensorSup{T}{1}$, and returns a $(m + n)$-order tensor $\tensor{T} \in \R^{I_0 \times \cdots \times I_{m-1} \times J_0 \times \cdots \times J_{n-1}}$ whose entries are computed as 
\beq
\label{def:tensor-outer-product}
\tensorSub{T}{i_0, \cdots, i_{m-1}, j_0, \cdots,  j_{n-1}} = \tensorInd{T}{i_0, \cdots, i_{m-1}}{0} ~ \tensorInd{T}{j_0, \cdots, j_{n-1}}{1} 
\eeq
It is not difficult to see that tensor outer product is a direct generalization for outer product for two vectors $\mymatrix{M} = \myvector{u} \otimes \myvector{v} = \myvector{u} ~ \myvector{v}^\top$. Obviously, the number of operations to compute a tensor outer product explicitly is $O ( ( \prod_{u = 0}^{m - 1} I_u ) ( \prod_{v = 0}^{n - 1} J_v ) )$. 

\paragraph{Tensor partial outer product}
Tensor partial outer product is a variant of tensor outer product defined above, which is widely used in conjunction with other operations. Given a $m$-order tensor $\tensorSup{T}{0} \in \R^{I_0 \times I_1 \times \cdots \times I_{m-1}}$ and another $n$-order tensor $\tensorSup{T}{1} \in \R^{J_0 \times J_1 \times \cdots \times J_{n-1}}$, which share the same dimension at mode-$k$ of $\tensorSup{T}{0}$ and mode-$l$ of $\tensorSup{T}{1}$ (i.e. $I_k = J_l$), the mode-$(k, l)$ partial outer product of $\tensorSup{T}{0}$ and $\tensorSup{T}{1}$, denoted as $\tensor{T} \triangleq \tensorSup{T}{0} \otimes^{k}_{l} \tensorSup{T}{1}$, returns a $(m + n - 1)$-order tensor $\tensor{T} \in \R^{I_0 \times \cdots \times I_{m-1} \times J_0 \times \cdots \times J_{l-1} \times J_{l+1} \times \cdots \times J_{n-1}}$, whose entries are computed as
\begin{subequations}
\begin{align}
& \tensorSub{T}{i_0, \cdots, i_{k-1}, r, i_{k+1}, \cdots, i_{m-1}, j_0, \cdots, j_{l-1}, j_{l+1}, \cdots, j_{n-1}} \nonumber \\
= ~ & \tensorInd{T}{i_0, \cdots, i_{k-1}, r, i_{k+1}, \cdots, i_{m-1}}{0} ~ \tensorInd{T}{j_0, \cdots, j_{l-1}, r, j_{l+1}, \cdots, j_{n-1}}{1} 
\label{def:tensor-partial-outer-product} \\
= ~ & \tensorInd{T}{\cdots, r, \cdots}{0} \otimes \tensorInd{T}{\cdots, r, \cdots}{1}
\end{align}
\end{subequations}
The operation bears the name "partial outer product" because it reduces to outer product once we fix the indices at mode-$k$ of $\tensorSup{T}{0}$ and mode-$l$ of $\tensorSup{T}{1}$.
Referring to the computational complexity of tensor outer product, the number of operations for each fixed index is $O ( ( \prod_{u = 0, u \neq k}^{m - 1} I_u ) ( \prod_{v = 0, v \neq l}^{n - 1} J_v ) )$, therefore the total time complexity for the tensor partial outer product is $O ( ( \prod_{u = 0}^{m - 1} I_u ) ( \prod_{v = 0, v \neq l}^{n - 1} J_v ) )$.

%
%
%
%

\paragraph{Compound operations:}
 As building blocks, the basic tensor operations defined above can further combined to construct compound operations that perform multiple operations on multiple tensors simultaneously. We illustrate their usage using two representative examples in this section, and we will see more examples when we derive backpropagation rules for tensor operations in Appendix~\ref{app:derivatives}.
 
\begin{itemize}[leftmargin=*]

\item \textbf{Simultaneous multi-operations between two tensors. } For example, given two $3$-order tensors $\tensorSup{T}{0} \in \R^{R \times X \times S}$ and $\tensorSup{T}{1} \in \R^{R \times H \times S}$, we can define a compound operation $\left( \otimes^0_0 \circ \ast^1_1 \circ \times^2_2 \right)$ between $\tensorSup{T}{0}$ and $\tensorSup{T}{1}$, where mode-$(0,0)$ partial outer product, mode-$(1,1)$ convolution and mode-$(2,2)$ contraction are performed simultaneously, which results in a $2$-order tensor $\tensor{T}$ of $\R^{R \times X^{\prime}}$ (it is indeed a matrix, though denoted as a tensor). The entries of $\tensor{T} \triangleq \tensorSup{T}{0} \left( \otimes^0_0 \circ \ast^1_1 \circ \times^2_2 \right) \tensorSup{T}{1}$ are computed as
\beq
\tensorSub{T}{r, :} = \sum_{s = 0}^{S - 1} \tensorInd{T}{r, :, s}{0} \ast \tensorInd{T}{r, :, s}{1} 
\label{def:compound-1}
\eeq
For commonly used vector convolution, it is not difficult to show that number of operations required to compute the result $\tensor{T}$ is $O\left( R \max(X, H) \log(\max(X, H)) S \right)$ with FFT and $O(RXHS)$ without FFT, as each of the $R$ vectors in $\tensor{T}$ is computed with a sum of $S$ vector convolutions. 

\item \textbf{Simultaneous operations between a tensor and a set of multiple tensors}. For example, given a $3$-order tensor $\tensor{U} \in \R^{R \times X \times S}$ and a set of three tensors $\tensorSup{T}{0} \in \R^{R \times P}$, $\tensorSup{T}{1} \in \R^{K \times Q}$ and $\tensorSup{T}{2} \in \R^{S \times T}$, we can define a compound operation on $\tensor{U}$ as $\tensor{V} \triangleq \tensor{U} ( \otimes^{0}_{0} \tensorSup{T}{0} \ast^{1}_{0} \tensorSup{T}{1} \times^{2}_{0} \tensorSup{T}{2} )$, which performs mode-$(0,0)$ partial outer product with $\tensorSup{T}{0}$, mode-$(1, 0)$ convolution with $\tensorSup{T}{1}$ and mode-$(2, 0)$ contraction with $\tensorSup{T}{2}$ simultaneously. In this case, a $5$-order tensor $\tensor{V} \in \R^{R \times X^{\prime} \times P \times Q \times T}$ is returned, with entries calculated as 
\beq
\tensorSub{V}{r, :, p, q, t} = \sum_{s=0}^{S-1} \tensorInd{T}{r, p}{0} \left( \tensorSub{U}{r, :, s} \ast \tensorInd{T}{:, q}{1} \right) \tensorInd{T}{s, t}{2}
\label{def:compound-2}
\eeq
Identifying the best order to evaluate a compound operation with multiple tensors is in general an NP-hard problem, but for this example we can find it using exhaustive search: 
\beq
\tensorSub{V}{r, :, p, q, t} = \left( \left( \sum_{s=0}^{S-1}  \tensorSub{U}{r, :, s} \tensorInd{T}{s, t}{2} \right) \ast \tensorInd{T}{:, q}{1} \right) \tensorInd{T}{r, p}{0}
\eeq
If we follows the supplied brackets and break the evaluation into three steps, these steps take $O(RXST)$, $O(RXHPT)$ and $O(RX^{\prime}HPT)$ operations respectively, therefore result in a total time complexity of $O( RXST + RXHPT  + RX^{\prime} PQT)$ for the compound operation.

\end{itemize}

Generally, compound operations over multiple tensors are difficult to flatten into mathematical equations, and usually described by {\em tensor diagrams} as in Section~\ref{sec:preliminary}, which are usually called {\em tensor networks}~\cite{cichocki2016low}in the physics literature. 


\section{Backpropagation of tensor operations}
\label{app:derivatives}

In order to use the operations in Appendix~\ref{app:operations} in tensorial neural networks, we derive in this section their backpropagation rules. For brevity, we use tensor contraction as a tutorial example, and omit the derivations for other operations.   

\paragraph{Tensor contraction}
Recall the definition of tensor contraction in Equations~\eqref{def:tensor-contraction-1} and~\eqref{def:tensor-contraction-2} in tensor algebra:
\begin{equation}
\tensor{T} = \tensorSup{T}{1} \times \tensorSup{T}{2}
\label{def:tensor-contraction}
\end{equation}
The partial derivatives of the result $\tensor{T}$ w.r.t. its operants $\tensorSup{T}{0}$, $\tensorSup{T}{1}$ can be computed at the entries level:
\begin{subequations}
\begin{align}
\frac{\partial \tensorSub{T}{i_0, \cdots, i_{k-1}, i_{k+1}, \cdots, i_{m-1}, j_0, \cdots, j_{l-1}, j_{l+1}, \cdots, j_{n-1}}}{\partial \tensorInd{T}{i_0, \cdots, i_{k-1}, r, i_{k+1}, \cdots, i_{m-1}}{0}} & = \tensorInd{T}{j_0, \cdots, j_{l-1}, r, j_{l+1}, \cdots, j_{n-1}}{1}
 \label{eq:derivative-tensor-contraction-1} \\
\frac{\partial \tensorSub{T}{i_0, \cdots, i_{k-1}, i_{k+1}, \cdots, i_{m-1}, j_0, \cdots, j_{l-1}, j_{l+1}, \cdots, j_{n-1}}}{\partial \tensorInd{T}{j_0, \cdots, j_{l-1}, r, j_{l+1}, \cdots, j_{n-1}}{1}} & = \tensorInd{T}{i_0, \cdots, i_{k-1}, r,  i_{k+1}, \cdots, i_{m-1}}{0} 
\label{eq:derivative-tensor-contraction-2}
\end{align}
\end{subequations}
With chain rule, the derivatives of $\mathcal{L}$ w.r.t. $\tensorSup{T}{0}$ and $\tensorSup{T}{1}$ can be obtained through ${\partial \mathcal{L}}/{\partial \tensor{T}}$.
\begin{subequations}
\begin{gather}
\frac{\partial \mathcal{L}}{\partial \tensorInd{T}{i_0, \cdots, i_{k-1}, r, i_{k+1}, \cdots, i_{m-1}}{0}} = \sum_{j_0 = 0}^{J_0 - 1} \cdots \sum_{j_{l-1} = 0}^{J_{l-1} - 1} \sum_{j_{l+1} = 0}^{J_{l+1} - 1} \cdots \sum_{j_{n-1} = 0}^{J_{n-1} - 1} \nonumber \\
\frac{\partial \mathcal{L}}{\partial \tensorSub{T}{i_0, \cdots, i_{k-1}, i_{k+1}, \cdots, i_{m-1}, j_0, \cdots, j_{l-1}, j_{l+1}, \cdots, j_{n-1}}} ~ \tensorInd{T}{j_0, \cdots, j_{l-1}, r, j_{l+1}, \cdots, j_{n-1}}{1} \label{eq:backprop-tensor-contraction-1-1} \\
\frac{\partial \mathcal{L}}{\partial \tensorInd{T}{j_0, \cdots, j_{l-1}, r, j_{l+1}, \cdots, j_{n-1}}{1}} = \sum_{i_0 = 0}^{I_0 - 1} \cdots \sum_{i_{k-1} = 0}^{I_{k-1} - 1} \sum_{i_{k+1} = 0}^{I_{k+1} - 1} \cdots \sum_{i_{m-1} = 0}^{I_{m-1} - 1} \nonumber \\
\frac{\partial \mathcal{L}}{\partial \tensorSub{T}{i_0, \cdots, i_{k-1}, i_{k+1}, \cdots, i_{m-1}, j_0, \cdots, j_{l-1}, j_{l+1}, \cdots, j_{n-1}}} ~ \tensorInd{T}{i_0, \cdots, i_{k-1}, r, i_{k+1}, \cdots, i_{m-1}}{0} \label{eq:backprop-tensor-contraction-2-1} 
\end{gather}
\end{subequations}
These tedious equations can be greatly simplified with tensor notations introduced before.
\begin{subequations}
\begin{align}
\frac{\partial \mathcal{L}}{\partial \tensorSup{T}{0}} & = \swapaxes \left( \frac{\partial \mathcal{L}}{\partial \tensor{T}} \left( \times^{m-1}_{0} \circ \cdots \circ \times^{m+l-2}_{l-1} \circ \times^{m+l-1}_{l+1} \circ \cdots \circ \times^{m+n-3}_{n-1} \right) \tensorSup{T}{1} \right) 
\label{eq:backprop-tensor-contraction-1-2} \\
\frac{\partial \mathcal{L}}{\partial \tensorSup{T}{1}} & = \swapaxes \left( \frac{\partial \mathcal{L}}{\partial \tensor{T}} \left( \times^{0}_{0} \circ \cdots \circ \times^{k-1}_{k-1} \circ \times^{k}_{k+1} \circ \cdots \circ \times^{m-2}_{m-1} \right) \tensorSup{T}{0} \right) 
\label{eq:backprop-tensor-contraction-2-2}
\end{align}
\end{subequations}
where $\swapaxes(\cdot)$ aligns the modes of outputs. Notice that the backpropagation equations are compound operations, even if the original operation is a basic one. The number of operations required for both backpropagation equations are $O ( ( \prod_{u = 0}^{m - 1} I_u ) \allowbreak ( \prod_{v = 0, v \neq l}^{n - 1} J_v ) )$, which are exactly the same as in the forward computation in Equation~\ref{def:tensor-contraction}. 


\paragraph{Tensor multiplication} 
Recall the definition of tensor multiplication in Equations~\eqref{def:tensor-multiplication-1} and~\eqref{def:tensor-multiplication-2} in tensor algebra:
\begin{equation}
\tensor{V} = \tensor{U} \times_{k} \mymatrix{M}
\label{def:tensor-multiplication}
\end{equation}
Backpropagation equations:
\begin{subequations} 
\begin{gather}
\frac{\partial \mathcal{L}}{\partial \tensor{V}} = \frac{\partial \mathcal{L}}{\partial \tensor{U}} \times_{k} \mymatrix{M}^{\top} \label{eq:backprop-tensor-multiplication-1-2} \\
\frac{\partial \mathcal{L}}{\partial \mymatrix{M}} = \tensor{U} \left( \times^{0}_{0} \circ \cdots \circ \times^{k-1}_{k-1} \circ \times^{k+1}_{k+1} \circ \cdots \circ \times^{m-1}_{m-1} \right) \frac{\partial \mathcal{L}}{\partial \tensor{V}} \label{eq:backprop-tensor-multiplication-2-2}
\end{gather}
\end{subequations}
The time complexities for both equations are $O ( ( \prod_{u = 0}^{m - 1} I_u ) J )$, which is identical to the forward computation in Equation~\eqref{def:tensor-multiplication}.

\paragraph{Tensor convolution} 
Recall the definition of tensor convolution in Equation~\eqref{def:tensor-convolution} in tensor algebra:
\begin{equation}
\tensor{T} = \tensorSup{T}{1} \ast^{k}_{l} \tensorSup{T}{2} 
\label{def:tensor-convolution-2}
\end{equation}
Backpropagation equations:
\begin{subequations} 
\begin{align}
\frac{\partial \mathcal{L}}{\partial \tensorSup{T}{0}} 
& = \swapaxes \left( \frac{\partial \mathcal{L}}{\partial \tensor{T}} \big( \times^{m}_{0} \cdots \times^{m+l-1}_{l-1} \circ \ast^{k}_{l} \circ \times^{m+l-2}_{l+1} \circ \cdots \circ \times^{m+n-2}_{n-1} \big) \flipaxis(\tensorSup{T}{1}, l) \right) 
\nonumber \\
& = \swapaxes \left( \frac{\partial \mathcal{L}}{\partial \tensor{T}} \left( \times^{m}_{0} \cdots \times^{m+l-1}_{l-1} \circ (\ast^{k}_{l})^{\top} \circ \times^{m+l-2}_{l+1} \cdots \times^{m+n-2}_{n-1} \right) \tensorSup{T}{1} \right) 
\label{eq:backprop-tensor-convolution-1-2} \\
\frac{\partial \mathcal{L}}{\partial \tensorSup{T}{1}} 
& = \swapaxes \left( \frac{\partial \mathcal{L}}{\partial \tensor{T}} \big( \times^{0}_{0} \circ \cdots \circ \times^{k-1}_{k-1} \circ \ast^{k}_{k} \circ \times^{k+1}_{k+1} \circ \cdots \circ \times^{m-1}_{m-1} \big) \flipaxis(\tensorSup{T}{0}, k) \right) 
\nonumber \\
 & = \swapaxes \left( \frac{\partial \mathcal{L}}{\partial \tensor{T}} \left( \times^{0}_{0} \circ \cdots \circ \times^{k-1}_{k-1} \circ (\ast^{k}_{k})^{\top} \circ \times^{k+1}_{k+1} \circ \cdots \circ \times^{m-1}_{m-1} \right) \tensorSup{T}{0} \right) 
\label{eq:backprop-tensor-convolution-2-2}
\end{align}
\end{subequations}
where $(\ast^{k}_{l})^{\top}$ is the {\em adjoint operator} of $\ast^{k}_{l}$. Without FFT, 
the time complexities for the equations are $O ( ( \prod_{u = 0, u \neq k}^{m - 1} I_u ) \allowbreak ( \prod_{v = 0, v \neq l}^{n - 1} J_v ) I_k^{\prime} J_l  )$ and $O ( ( \prod_{u = 0, u \neq k}^{m - 1} I_u ) \allowbreak ( \prod_{v = 0, v \neq l}^{n - 1} J_v ) I_k^{\prime} I_k )$ respectively. 

\paragraph{Tensor outer product} 
Recall the definition of tensor outer product in Equation~\eqref{def:tensor-outer-product} in tensor alegebra:
\begin{equation}
\tensor{T} = \tensorSup{T}{0} \otimes \tensorSup{T}{1}
\label{def:tensor-outer-product-2}
\end{equation}
Backpropagation equations:
\begin{subequations}
\begin{align}
\frac{\partial \mathcal{L}}{\partial \tensorSup{T}{0}} & = \frac{\partial \mathcal{L}}{\partial \tensor{T}} \left( \times^{m}_{0} \circ \cdots \circ \times^{m+n-1}_{n-1} \right) \tensorSup{T}{1} 
\label{eq:backprop-tensor-outer-product-1-2} \\
\frac{\partial \mathcal{L}}{\partial \tensorSup{T}{1}} & = \frac{\partial \mathcal{L}}{\partial \tensor{T}} \left( \times^{0}_{0} \circ \cdots \circ \times^{m-1}_{m-1} \right) \tensorSup{T}{0} 
\label{eq:backprop-tensor-outer-product-2-2}
\end{align}
\end{subequations}
The number of operations required for both equations are $O ( ( \prod_{u = 0, u \neq k}^{m - 1} I_u ) \allowbreak ( \prod_{v = 0}^{n - 1} J_v ) )$.

\paragraph{Tensor partial outer product}
Recall the definition of tensor partial outer product in Equation~\eqref{def:tensor-partial-outer-product}:
\begin{equation}
\tensor{T} = \tensorSup{T}{0} \otimes^{k}_{l} \tensorSup{T}{1}
\label{def:tensor-partial-outer-product}
\end{equation}
Backpropagation equations:
\begin{subequations}  
\begin{align}
\frac{\partial \mathcal{L}}{\partial \tensorSup{T}{0}} & = \swapaxes \left( \frac{\partial \mathcal{L}}{\partial \tensor{T}} \left( \times^{m}_{0} \circ \cdots \circ \times^{m+l-1}_{l-1} \circ \otimes^{k}_{l} \circ \times^{m+l-2}_{l+1} \circ \cdots \circ \times^{m+n-2}_{n-1} \right) \tensorSup{T}{1} \right) \label{eq:backprop-tensor-partial-outer-product-1-2} \\
\frac{\partial \mathcal{L}}{\partial \tensorSup{T}{1}} & = \swapaxes \left( \frac{\partial \mathcal{L}}{\partial \tensor{T}} \left( \times^{0}_{0} \circ \cdots \circ \times^{k-1}_{k-1} \circ \otimes^{k}_{k} \circ \times^{k+1}_{k+1} \circ \cdots \circ \times^{m-1}_{m-1} \right) \tensorSup{T}{0} \right) \label{eq:backprop-tensor-partial-outer-product-2-2}
\end{align}
\end{subequations}
It is not difficult to show the time complexity for both equations above are $O( ( \prod_{u = 0}^{m - 1} I_u ) \allowbreak ( \prod_{v = 0, v \neq l}^{n - 1} J_v ) )$.

\section{Tensor decompositions}
\label{app:decompositions}

Tensor decompositions are natural extensions of matrix factorizations for multi-dimensional arrays. In this section, we will review three commonly used tensor decompositions, namely {\em \CPlong (\CP) decomposition}~\cite{kolda2009tensor}, {\em \TKlong (\TK) decomposition}~\cite{kolda2009tensor} and {\em \TTlong (\TT) decomposition}~\cite{oseledets2011tensor}. 


\paragraph{\CP decomposition} 
\CP decomposition is a direct generalization of singular value decomposition (\SVD) which decomposes a tensor into additions of rank-1 tensors (outer product of multiple vectors). Specifically, given an $m$-order tensor $\tensor{T} \in \R^{I_0 \times I_1 \times \cdots \times I_{m-1}}$, \CP decomposition factorizes it into $m$ factor matrices $\{ \mymatrix{M}^{(0)} \}_{l = 0}^{m - 1}$, where $\mymatrix{M}^{(l)} \in \R^{R \times I_l}, \forall l \in [m]$, where $R$ is called the \textit{canonical rank} of the \CP decomposition, which is allowed to be larger than the $I_l$'s.
\begin{subequations}
\begin{gather}
\tensorSub{T}{i_0, \cdots, i_{m-1}} \triangleq \sum_{r = 0}^{R - 1} \mymatrix{M}^{(0)}_{r, i_0} \cdots \mymatrix{M}^{(m-1)}_{r, i_{m-1}} 
\label{def:cp-decomposition-1} \\
\tensor{T} \triangleq \sum_{r = 0}^{R - 1} \mymatrix{M}^ {(0)}_{r, :} \otimes \cdots \otimes \mymatrix{M}^{(m-1)}_{r, :} = \myvector{1} \times^{0}_{0} \left( \mymatrix{M}^{(0)} \otimes^{0}_{0} \cdots \otimes^{0}_{0} \mymatrix{M}^{(m-1)} \right)  \label{def:cp-decomposition-2}
\end{gather}
\end{subequations}
where $\myvector{1} \in \R^{R}$ is an all-ones vector of length $R$. With \CP decomposition, $\tensor{T}$ can be represented with only $( \sum_{l = 0}^{m - 1} I_l ) R$ entries instead of $( \prod_{l = 0}^{m - 1} I_l )$ as in the original tensor. 

\paragraph{\TKlong decomposition} \TKlong decomposition provides more general factorization than \CP decomposition. Given an $m$-order tensor $\tensor{T} \in \R^{I_0 \times I_1 \times \cdots \times I_{m-1}}$, \TKlong decomposition factors it into $m$ factor matrices $\{ \matrixSup{M}{l} \}_{l = 0}^{m - 1}$, where $\matrixSup{M}{l} \in \R^{R_l \times I_l}, \forall l \in [m]$ and an additional $m$-order core tensor $\tensor{C} \in \R^{R_0 \times R_1 \times \dots \times R_{m-1}}$, where the {\em \TKlong ranks} $R_l$'s are required to be smaller or equal than the dimensions at their corresponding modes, i.e. $R_{l} \leq I_l, \forall l \in [m]$. 
\begin{subequations}
\begin{gather}
\tensorSub{T}{i_0, \cdots, i_{m-1}} \triangleq \sum_{r_0 = 0}^{R_0 - 1} \dots \sum_{r_{m-1} = 0}^{R_{m-1}  - 1} \tensorSub{C}{r_0, \dots, r_{m-1}} ~ \mymatrix{M}^{(0)}_{r_0, i_0} \cdots \mymatrix{M}^{(m-1)}_{r_{m-1}, i_{m-1}} 
\label{def:tucker-decomposition-1} \\
\tensor{T} \triangleq \tensor{C} \left( \times_{0} \mymatrix{M}^{(0)} \times_{1} \mymatrix{M}^{(1)} \cdots \times_{m-1} \mymatrix{M}^{(m-1)} \right)
\label{def:tucker-decomposition-2}
\end{gather}
\end{subequations}
Notice that when $R_0 = \cdots = R_{m-1} = R$ and $\tensor{C}$ is a super-diagonal tensor with all super-diagonal entries to be ones (a.k.a. identity tensor), \TKlong decomposition reduces to \CP decomposition, and therefore \CP decomposition is a special case of \TKlong decomposition. With \TKlong decomposition, a tensor is approximately by $(\prod_{l=0}^{m-1} R_l + \sum_{l=0}^{m-1} I_l R_l)$ entries. 

\paragraph{\TTlong decomposition} 
\TTlong decomposition factorizes a $m$-order tensor into $m$ interconnected low-order tensors $\{ \tensorSup{T}{l}  \}_{l = 0}^{m - 1}$, where $\tensorSup{T}{l} \in \R^{R_l \times I_l \times R_{l+1}}, l = 1, \cdots, m-2$ with $\tensorSup{T}{0} \in \R^{I_0 \times R_0}$,  and $\tensorSup{T}{m-1} \in \R^{R_{m-1}\times I_{m-1}}$ such that
\begin{subequations}
\begin{gather}
\tensorSub{T}{i_0, \dots, i_{m-1}} \triangleq \sum_{r_0 = 1}^{R_0 - 1} \dots \sum_{r_{m-2} = 1}^{R_{m-2} - 1} \tensorInd{T}{i_0, r_0}{0} ~ \tensorInd{T}{r_0, i_1, r_1}{1} \cdots \tensorInd{T}{r_{m-2}, i_{m-1}}{m-1} 
\label{def:Tensor-train-decomposition-1} \\
\tensor{T} \triangleq \tensorSup{T}{0} \times^{-1}_{0} \tensorSup{T}{1} \times^{-1}_{0} \cdots \times^{-1}_{0} \tensorSup{T}{m-1} 
\label{def:Tensor-train-decomposition-2}
\end{gather}
\end{subequations}
where the $R_l$'s are known as \textit{\TTlong ranks}, which controls the tradeoff between complexity and accuracy of the representation. With \TTlong decomposition, a tensor is represented by $(R_0 I_0 + \sum_{l=1}^{m-2} R_l I_l R_{l+1} + R_{m-1} I_{m-1})$ entries.

\begin{figure*}[!htbp]
\begin{minipage}{\textwidth}
\centering
\begin{subfigure}[c]{0.17\textwidth}
\centering
	\psfrag{n1}[][][0.8]{$\tensor{K}$}
	\psfrag{H}[][][0.8]{$H$}
	\psfrag{W}[][][0.8]{$W$}
	\psfrag{S}[][][0.8]{$S$}
	\psfrag{T}[][][0.8]{$T$}
	\includegraphics[width=0.6\textwidth]{\fighome/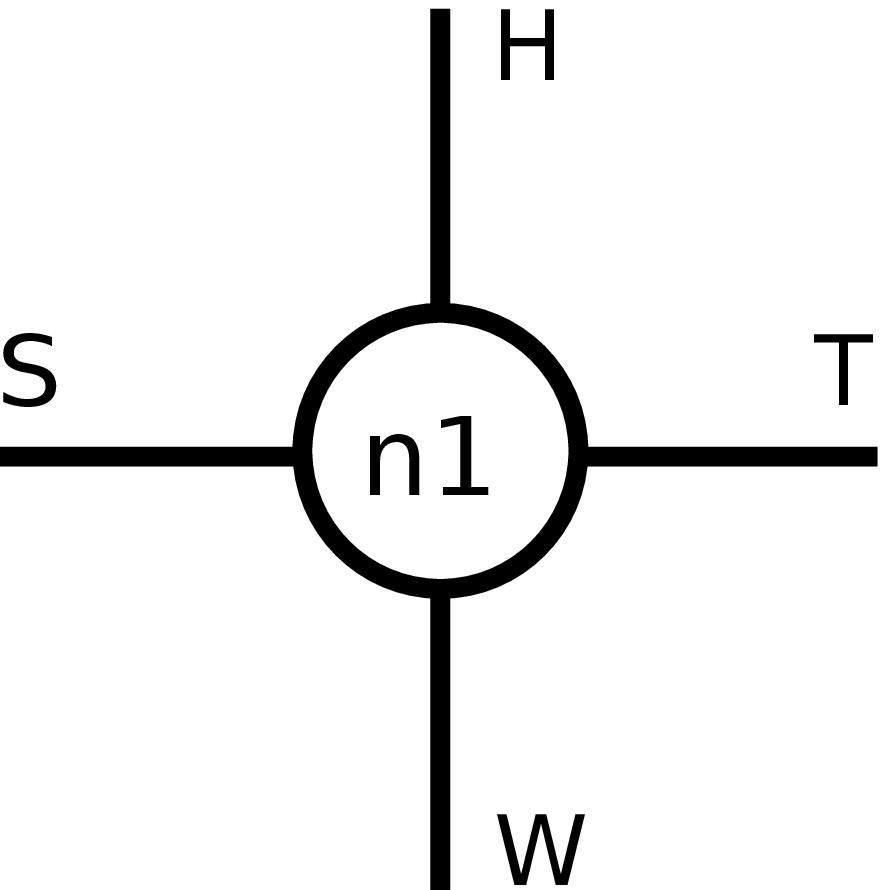}
	\caption{Original kernel}
\label{fig:decomposition-original}
\end{subfigure}
\hfill
\begin{subfigure}[c]{0.23\textwidth}
\centering
	\psfrag{n1}[][][0.5]{$\tensor{K}^{C}$}
	\psfrag{n2}[][][0.5]{$\tensor{K}^{S}$}
	\psfrag{n3}[][][0.5]{$\tensor{K}^{T}$}
	\psfrag{H}[][][0.6]{$H$}
	\psfrag{W}[][][0.6]{$W$}
	\psfrag{S}[][][0.6]{$S$}
	\psfrag{T}[][][0.6]{$T$}
	\psfrag{R}[][][0.6]{$R$}
	\includegraphics[width=0.6\textwidth]{\fighome/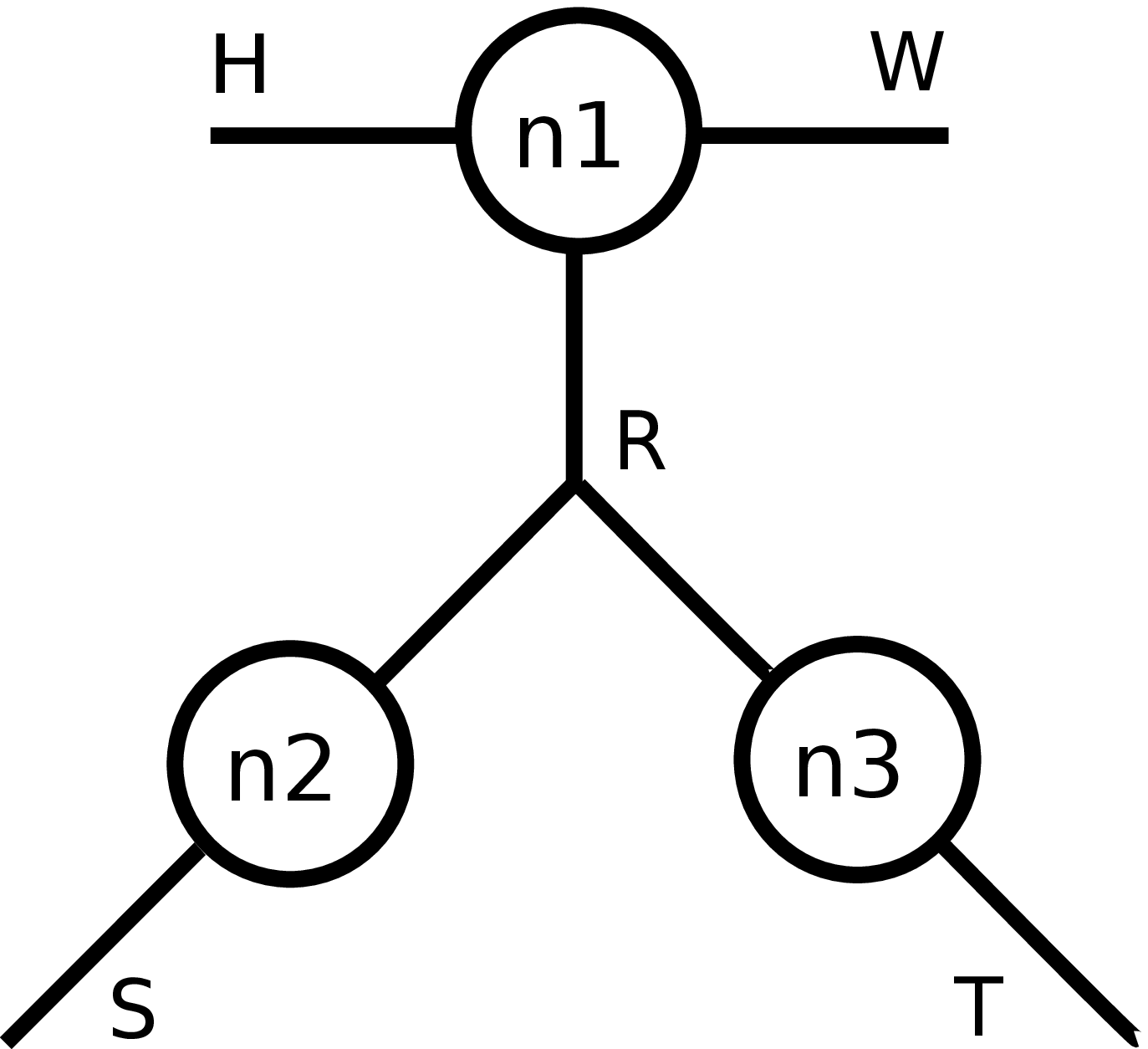}
\caption{\CP}
\label{fig:decomposition-cp}
\end{subfigure}
\begin{subfigure}[c]{0.26\textwidth}
\centering
	\psfrag{n1}[][][0.5]{$\tensor{K}^{S}$}
	\psfrag{n2}[][][0.5]{$\tensor{K}^{C}$}
	\psfrag{n3}[][][0.5]{$\tensor{K}^{T}$}
	\psfrag{H}[][][0.6]{$H$}
	\psfrag{W}[][][0.6]{$W$}
	\psfrag{S}[][][0.6]{$S$}
	\psfrag{T}[][][0.6]{$T$}
	\psfrag{R1}[][][0.6]{$R_s$}
	\psfrag{R2}[][][0.6]{$R_t$}
	\includegraphics[width=\textwidth]{\fighome/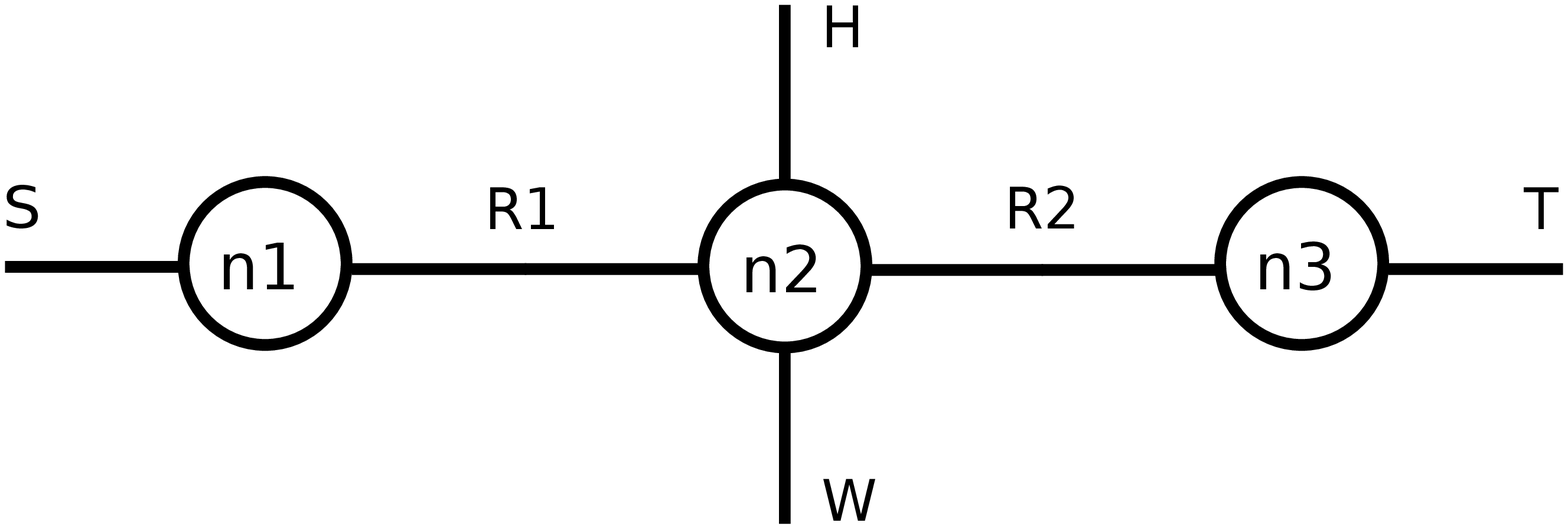}
\caption{\TKlong (\TK)}
\label{fig:decomposition-tk}
\end{subfigure}
\hfill
\begin{subfigure}[c]{0.26\textwidth}
\centering
	\psfrag{n1}[][][0.4]{$\tensor{K}^{S}$}
	\psfrag{n2}[][][0.4]{$\tensor{K}^{H}$}
	\psfrag{n3}[][][0.4]{$\tensor{K}^{W}$}
	\psfrag{n4}[][][0.4]{$\tensor{K}^{T}$}
	\psfrag{H}[][][0.6]{$H$}
	\psfrag{W}[][][0.6]{$W$}
	\psfrag{S}[][][0.6]{$S$}
	\psfrag{T}[][][0.6]{$T$}
	\psfrag{R1}[][][0.6]{$R_s$}
	\psfrag{R2}[][][0.6]{$R$}
	\psfrag{R3}[][][0.6]{$R_t$}
	\includegraphics[width=\textwidth]{\fighome/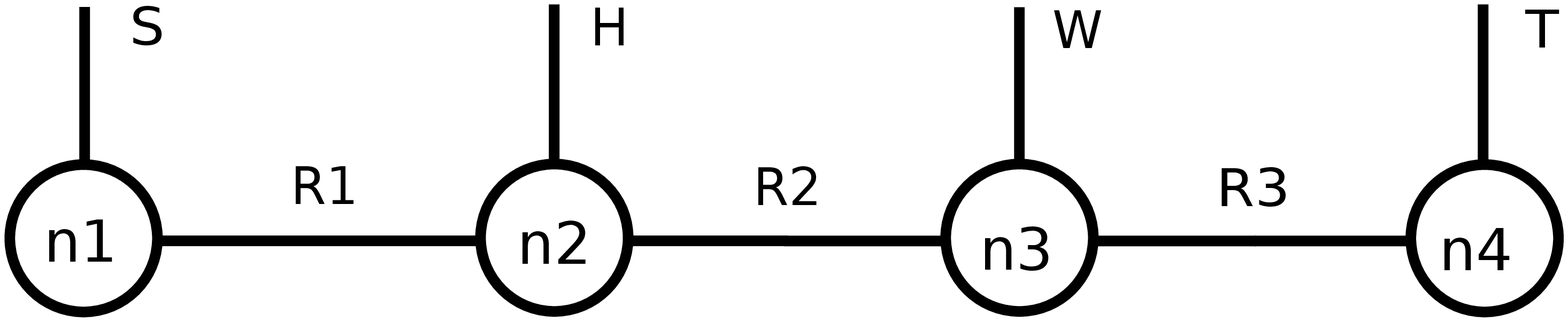}
\caption{\TTlong (\TT)}
\label{fig:decomposition-tt}
\end{subfigure}
\vskip \baselineskip
\begin{subfigure}[c]{0.19\textwidth}
\centering
	\psfrag{n1}[][][0.8]{$\tensor{K}^{\prime}$}
	\psfrag{H}[][][0.8]{$H$}
	\psfrag{W}[][][0.8]{$W$}
	\psfrag{S1}[][][0.8]{$S_0$}
	\psfrag{S2}[][][0.8]{$S_1$}
	\psfrag{S3}[][][0.8]{$S_2$}
	\psfrag{T1}[][][0.8]{$T_0$}
	\psfrag{T2}[][][0.8]{$T_1$}
	\psfrag{T3}[][][0.8]{$T_2$}
	\includegraphics[width=0.6\textwidth]{\fighome/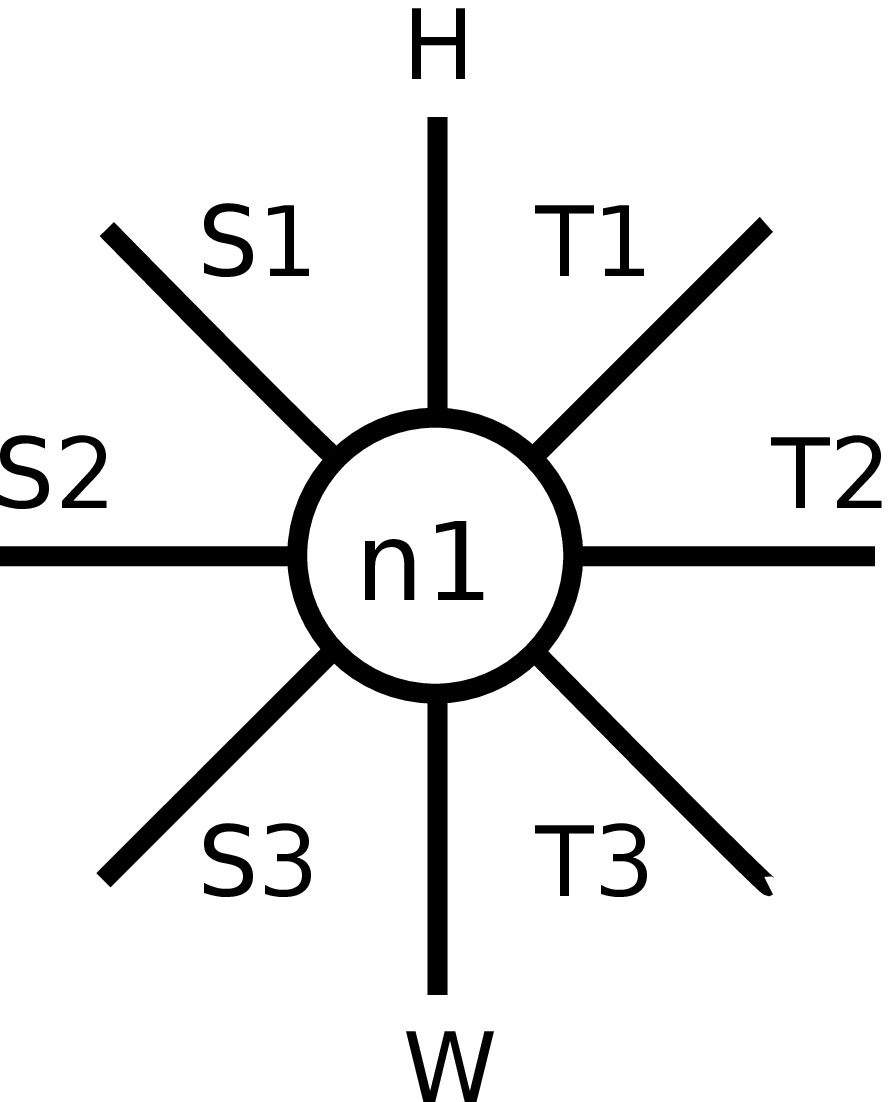}
\caption{{Tensorized kernel}}
\label{fig:decomposition-tensorized}
\end{subfigure}
\hfill
\begin{subfigure}[c]{0.21\textwidth}
\centering
	\psfrag{n1}[][][0.5]{$\tensor{K}^{C}$}
	\psfrag{n2}[][][0.5]{$\tensor{K}^{0}$}
	\psfrag{n3}[][][0.5]{$\tensor{K}^{1}$}
	\psfrag{n4}[][][0.5]{$\tensor{K}^{2}$}
	\psfrag{H}[][][0.6]{$H$}
	\psfrag{W}[][][0.6]{$W$}
	\psfrag{S1}[][][0.6]{$S_0$}
	\psfrag{S2}[][][0.6]{$S_1$}
	\psfrag{S3}[][][0.6]{$S_2$}
	\psfrag{T1}[][][0.6]{$T_0$}
	\psfrag{T2}[][][0.6]{$T_1$}
	\psfrag{T3}[][][0.6]{$T_2$}
	\psfrag{R}[][][0.6]{$R$}
	\includegraphics[width=0.6\textwidth]{\fighome/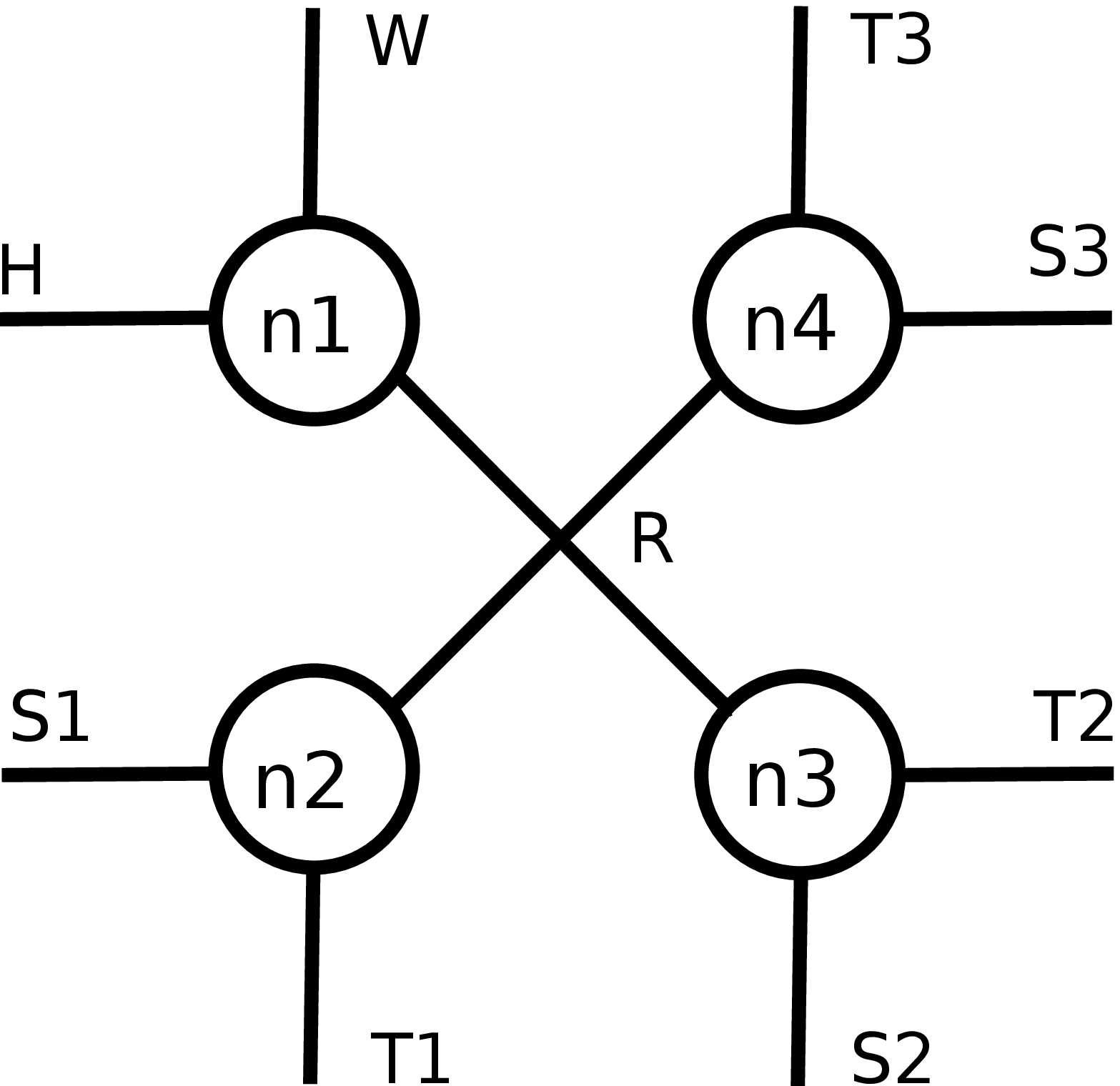}
\caption{\rCP (m=3)}
\label{fig:decomposition-rcp}
\end{subfigure}
\hfill
\begin{subfigure}[c]{0.26\textwidth}
\centering
	\psfrag{c}[][][0.5]{$\tensor{C}$}
	\psfrag{n1}[][][0.5]{$\tensor{P}^{0}$}
	\psfrag{n2}[][][0.5]{$\tensor{P}^{1}$}
	\psfrag{n3}[][][0.5]{$\tensor{P}^{2}$}
	\psfrag{n4}[][][0.5]{$\tensor{Q}^{0}$}
	\psfrag{n5}[][][0.5]{$\tensor{Q}^{1}$}
	\psfrag{n6}[][][0.5]{$\tensor{Q}^{2}$}
	\psfrag{H}[][][0.6]{$H$}
	\psfrag{W}[][][0.6]{$W$}
	\psfrag{S1}[][][0.6]{$S_0$}
	\psfrag{S2}[][][0.6]{$S_1$}
	\psfrag{S3}[][][0.6]{$S_2$}
	\psfrag{T1}[][][0.6]{$T_0$}
	\psfrag{T2}[][][0.6]{$T_1$}
	\psfrag{T3}[][][0.6]{$T_2$}
	\psfrag{R1}[][][0.6]{$R^{s}_0$}
	\psfrag{R2}[][][0.6]{$R^{s}_1$}
	\psfrag{R3}[][][0.6]{$R^{s}_2$}
	\psfrag{R4}[][][0.6]{$R^{t}_0$}
	\psfrag{R5}[][][0.6]{$R^{t}_1$}
	\psfrag{R6}[][][0.6]{$R^{t}_2$}
	\includegraphics[width=\textwidth]{\fighome/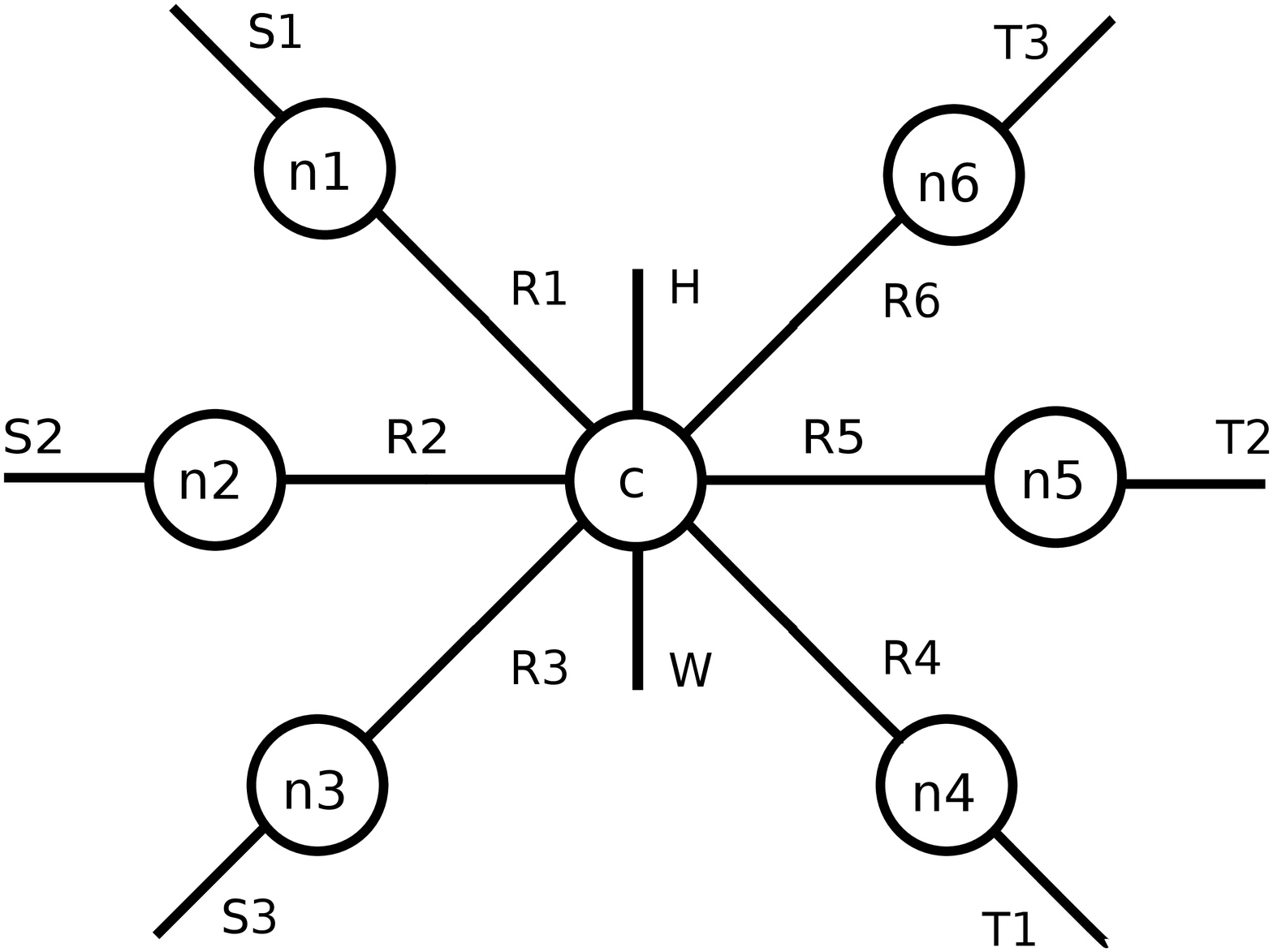}
\caption{\rTK (m=3)}
\label{fig:decomposition-rtk}
\end{subfigure}
\hfill
\begin{subfigure}[c]{0.26\textwidth}
\centering
	\psfrag{n1}[][][0.5]{$\tensor{K}^{0}$}
	\psfrag{n2}[][][0.5]{$\tensor{K}^{1}$}
	\psfrag{n3}[][][0.5]{$\tensor{K}^{2}$}
	\psfrag{n4}[][][0.5]{$\tensor{K}^{C}$}
	\psfrag{H}[][][0.6]{$H$}
	\psfrag{W}[][][0.6]{$W$}
	\psfrag{S1}[][][0.6]{$S_0$}
	\psfrag{S2}[][][0.6]{$S_1$}
	\psfrag{S3}[][][0.6]{$S_2$}
	\psfrag{T1}[][][0.6]{$T_0$}
	\psfrag{T2}[][][0.6]{$T_1$}
	\psfrag{T3}[][][0.6]{$T_2$}
	\psfrag{R1}[][][0.6]{$R_0$}
	\psfrag{R2}[][][0.6]{$R_1$}
	\psfrag{R3}[][][0.6]{$R_2$}
	\includegraphics[width=\textwidth]{\fighome/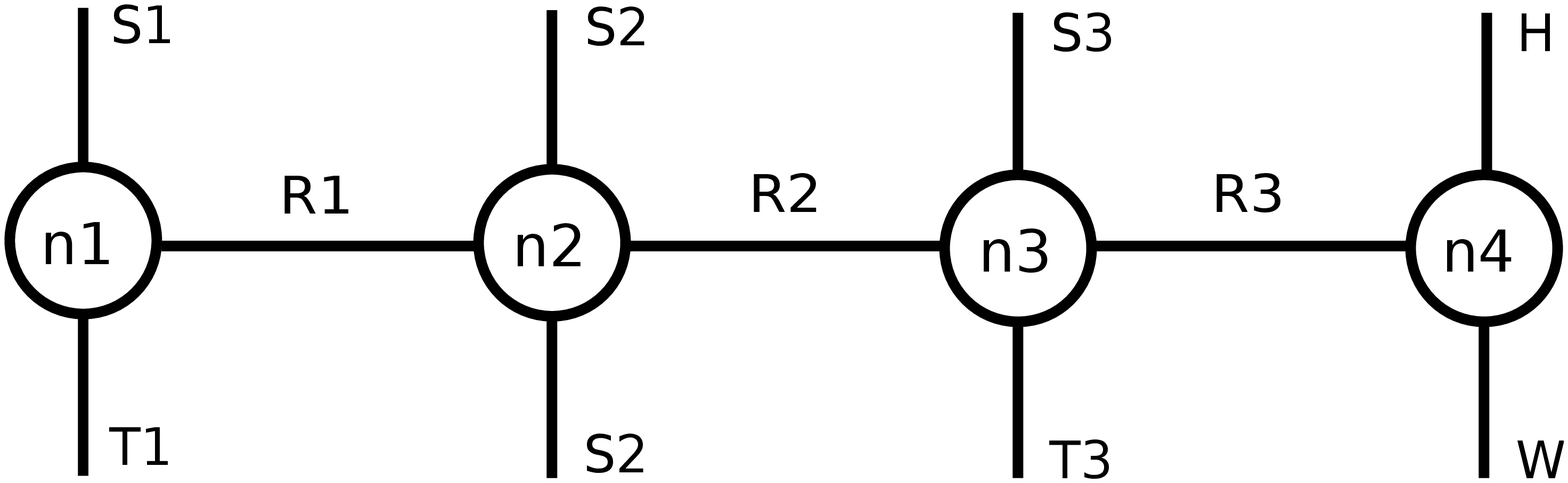}
\caption{\rTT (m=3)}
\label{fig:decomposition-rtt}
\end{subfigure}
\vspace{-0.5em}
\caption{\textbf{Diagrams of general tensor decompositions}. Figures~\textbf{(a)} and~\textbf{(e)} are typical kernels of \convnetshort and \tensornetshort respectively. Figures~\textbf{(b) (c)} and \textbf{(d)} are three types of tensor decompositions on the convolutional kernel $\tensor{K}$ in \textbf{(a)}. Figures~\textbf{(f) (g)} and \textbf{(h)} are three types of general tensor decompositions for our tensorial kernel $\tensor{K}^{\prime}$ in \textbf{(e)}}.
\label{fig:decompositions}
\end{minipage}
\end{figure*}

\paragraph{General tensor decompositions}
\label{general_decomposition}
In this paper, we use the term tensor decomposition in a more general way, i.e. we do not stick to the standard formats defined in the previous paragraphs. Indeed, we consider tensor decomposition as a reverse mapping of tensor operations: consider a general tensor operation $f$ on $m$ input tensors $\{ \tensorSup{T}{l} \}_{l = 0}^{m - 1}$ such that $\hat{\tensor{T}} = f(\tensorSup{T}{0}, \cdots, \tensorSup{T}{m-1})$ (i.e. $\hat{\tensor{T}}$ is linear in each operant $\tensorSup{T}{l}$), the corresponding general tensor decomposition aims to recover the input factors $\{ \tensorSup{T}{l} \}_{l = 0}^{m - 1}$ from a given tensor $\tensor{T}$ such that $\tensor{T} \approx \hat{\tensor{T}} = f(\tensorSup{T}{0}, \cdots, \tensorSup{T}{m-1})$. In Figure~\ref{fig:decompositions}, we demonstrate a few examples of general tensor decomposition using tensor diagrams.

\begin{table*}[!htbp]
\centering
\begin{tabular}{c | c | c | c }
Decomp. & Notations & $O$(\# of params.) & $O$(\# of backprop ops.)  \\ 
\hline
original & $\tensor{T}$ & $I$ & 0 \\
\CP & $\myvector{1} \times^{0}_{0} ( \mymatrix{M}^{(0)} \otimes^{0}_{0} \cdots \otimes^{0}_{0} \mymatrix{M}^{(m-1)} ) $ & $m I^{\frac{1}{m}} R$ & $m I R$ \\
\TK & $\tensor{C} ( \times_{0} \mymatrix{M}^{(0)}  \cdots \times_{m-1} \mymatrix{M}^{(m-1)} )$ & $R^{m} + m I^{\frac{1}{m}} R$ & $m^2 I R$ \\
\TT & $\tensorSup{T}{0} \times^{-1}_{0} \cdots \times^{-1}_{0} \tensorSup{T}{m-1}$ & $m I^{\frac{1}{m}} R^2$ & $m I R^2$ 
\end{tabular}
\caption{\textbf{Summary of tensor decompositions.} In this table, we summarize three types of tensor decompositions in tensor notations, and list their numbers of parameters and time complexities to backpropagate the gradient of a tensor $\tensor{T} \in \R^{I_0 \times I_1 \times \cdots I_{m-1}}$ to its $m$ factors (and an additional core tensor $\tensor{C}$ for \TKlong decomposition). For simplicity, we assume all dimensions $I_l$'s of $\tensor{T}$ are equal, and denote the size of $\tensor{T}$ as the product of all dimensions $I = \prod_{l = 0}^{m-1} I_l$. Furthermore, we assume all ranks $R_l$'s (in \TKlong and \TTlong decompositions) share the same number $R$. }

\label{table:decompositions}
\end{table*}

\section{Low-rank approximations of convolutional layer}
\label{app:convolutional}

In our paper, compact architecture of neural networks can be derived in two steps: (1) proposing an arbitrary architecture; and (2) decomposing the weights kernel at each layer into lower-order factors. When the proposed architecture is a standard convolutional neural network and the kernel is decomposed by traditional tensor decomposition, the derived architectures are known as low-rank approximations, in which each layer in the derived architecture is still a traditional operation (i.e. not general tensor operation).    

\paragraph{Standard convolutional layer} 
We review the standard convolutional neural network (CNN) for reference. In CNN, a convolutional layer is parameterized by a 4-order kernel $\tensor{K} \in \R^{H \times W \times S \times T}$, where $H$ and $W$ are height and width of the filters (which are typically equal), $S$ and $T$ are the number of input and output channels respectively. A convolutional layer maps a 3-order tensor $\tensor{U} \in \R^{X \times Y \times S}$ to another 3-order tensor $\tensor{V} \in \R^{X^{\prime} \times Y^{\prime} \times T}$, where $X$ and $Y$ are the height and width for the input feature map, while $X^{\prime}$ and $Y^\prime$ are the ones for the output feature map, with the following equation:
\beq
\tensor{V} = \tensor{U} \left( \ast_{0}^{0} \circ \ast_{1}^{1} \circ \times_{2}^{2} \right) \tensor{K} 
\label{def:convolutional-2}
\eeq
Notice that the number of parameters in a standard convolutional layer is $HWST$ and the number of operations needed to evaluate the output $\tensor{V}$ is $O(HWSTXY)$.

\paragraph{\SVD-convolutional layer} 
In \SVD-convolutional layer~\cite{jaderberg2014speeding}, the weights kernel $\tensor{K}$ is factorized by SVD.
\begin{equation}
\tensor{K} = \swapaxes \left( \tensorSup{K}{0} \times^{2}_{1} \tensorSup{K}{1} \right) \label{def:convolutional-svd-2}
\end{equation}
where $\tensorSup{K}{0} \in \R^{H \times S \times R}$ and $\tensorSup{K}{1} \in \R^{W \times R \times T}$ are the two factor tensors. With two factors, the forward pass of \SVD-convolutional layer is computed in two steps:
\begin{subequations}
\begin{align}
\tensorSup{U}{0} & = \tensor{U} \left( \ast_{0}^{0} \circ \times^{2}_{1} \right) \tensorSup{K}{0} \\
\tensor{V} & = \tensorSup{U}{0} \left( \ast_{1}^{1} \circ \times^{2}_{1} \right) \tensorSup{K}{1}
\end{align}
\end{subequations}
where $\tensorSup{U}{0}$ is an intermediate tensor after the first step.
The backpropagation equations for these two steps are:
\begin{subequations}
\begin{align}
\frac{\partial \mathcal{L}}{\partial \tensorSup{U}{0}} = \frac{\partial \mathcal{L}}{\partial \tensor{V}} \left( (\ast^{1}_{0})^{\top} \circ \times^{2}_{2} \right) ~ \tensorSup{K}{1}, ~ & \frac{\partial \mathcal{L}}{\partial \tensorSup{K}{1}} = \tensorSup{U}{0} \left( \times^{0}_{0} \circ (\overline{\overline{\ast}^{1}_{1})^{\top}} \right) \frac{\partial \mathcal{L}}{\partial \tensor{V}} \\
\frac{\partial \mathcal{L}}{\partial \tensor{U}} = \frac{\partial \mathcal{L}}{\partial \tensorSup{U}{0}}\left( (\ast^{0}_{0})^{\top} \circ \times^{2}_{2} \right) ~ \tensorSup{K}{0}, ~ & \frac{\partial \mathcal{L}}{\partial \tensorSup{K}{0}} = \tensor{U} \left( \overline{(\overline{\ast}^{0}_{0})^{\top}} \circ \times^{1}_{1} \right) \frac{\partial \mathcal{L}}{\partial \tensorSup{U}{0}}
\end{align}
\end{subequations}

\paragraph{\CP-convolutional layer} 
In \CP-convolutional layer~\cite{lebedev2014speeding, denton2014exploiting}, the weights kernel $\tensor{K}$ is factorized by CP decomposition.
\begin{equation}
\tensor{K} = \myvector{1} \times^{0}_{2} \left( \tensorSup{K}{1} \otimes^{2}_{1} \tensorSup{K}{0} \otimes^{2}_{0} \tensorSup{K}{2} \right) \label{def:convolutional-cp-2}
\end{equation}
where $\tensorSup{K}{0} \in \R^{S \times R}$, $\tensorSup{K}{1} \in \R^{H \times W \times R}$ and $\tensorSup{K}{2} \in \R^{R \times T}$ are three factor tensors.
With three factors, the forward pass of \CP-convolutional consists of three steps:
\begin{subequations}
\begin{align}
\tensorSup{U}{0} & = \tensor{U} \times_2 \tensorSup{K}{0} \\
\tensorSup{U}{1} & = \tensorSup{U}{0} \left( \ast_{0}^{0} \circ \ast_{1}^{1} \circ \otimes_{2}^{2} \right) \tensorSup{K}{1} \\
\tensor{V} & = \tensorSup{U}{1} \times_2 \tensorSup{K}{2}
\end{align}
\end{subequations}
where $\tensorSup{U}{0} \in \R^{X \times Y \times R}$ and $\tensorSup{U}{1} \in \R^{X^{\prime} \times Y^{\prime} \times R}$ are two intermediate tensors, and the corresponding backpropagation equations can then be computed as:
\begin{subequations}
\begin{gather}
\frac{\partial \mathcal{L}}{\partial \tensorSup{U}{1}}  =  \frac{\partial \mathcal{L}}{\partial \tensor{V}} \times_2 (\tensorSup{K}{2})^{\top}, ~ \frac{\partial \mathcal{L}}{\partial \tensorSup{K}{2}} = \tensorSup{U}{1} \left( \times^{0}_{0} \circ \times^{1}_{1} \right) \frac{\partial \mathcal{L}}{\partial \tensor{V}} \\
\frac{\partial \mathcal{L}}{\partial \tensorSup{U}{0}} = \frac{\partial \mathcal{L}}{\partial \tensorSup{U}{1}} \left( (\ast^{0}_{0})^{\top} \circ ({\ast}^{1}_{1})^{\top} \circ \otimes^{2}_{2} \right) \tensorSup{K}{1}, 
\frac{\partial \mathcal{L}}{\partial \tensorSup{K}{1}} = \frac{\partial \mathcal{L}}{\partial \tensorSup{U}{1}} \left( \overline{(\overline{\ast}^{0}_{0})^{\top}} \circ \overline{(\overline{\ast}^{1}_{1})^{\top}} \circ \otimes^{2}_{2} \right) \tensorSup{U}{1} \nonumber \\
\frac{\partial \mathcal{L}}{\partial \tensor{U}} = \frac{\partial \mathcal{L}}{\partial \tensorSup{U}{0}} \times_2 (\tensorSup{K}{0})^{\top}, ~ \frac{\partial \mathcal{L}}{\partial \tensorSup{K}{0}} = \tensor{U} \left( \times^{0}_{0} \circ \times^{1}_{1} \right) \frac{\partial \mathcal{L}}{\partial \tensorSup{U}{0}}
\end{gather}
\end{subequations}

\paragraph{\TK-convolutional layer} 
In \TK-convolutional layer~\cite{kim2015compression}, the kernel $\tensor{K}$ is factorized by a partial Tucker decomposition,
\begin{equation}
\tensor{K} = \tensorSup{K}{1} \left( \times_{2} (\tensorSup{K}{0})^\top \times_{3} \tensorSup{K}{2} \right) 
\label{def:convolutional-tk-2}
\end{equation}
where $\tensorSup{K}{0} \in \R^{S \times R_s}$, $\tensorSup{K}{1} \in \R^{H \times W \times R_s \times R_t}$ and $\tensorSup{K}{2} \in \R^{R_t \times T}$ are three factor tensors.
Again, the forward pass of \TKlong-convolutional layer has three steps:
\begin{subequations}
\begin{align}
\tensorSup{U}{0} & = \tensor{U} \times_{2}  \tensorSup{K}{0} \\
\tensorSup{U}{1} & = \tensorSup{U}{0} \left( \ast_{0}^{0} \circ \ast_{1}^{1} \circ \times_{2}^{2} \right) \tensorSup{K}{1} \\
\tensor{V} & = \tensorSup{U}{1} \times_{2} \tensorSup{K}{2}
\end{align}
\end{subequations}
where $\tensorSup{U}{0} \in \R^{X \times Y \times R_s}$ and $\tensorSup{U}{1} \in \R^{X^{\prime} \times Y^{\prime} \times R_t}$ are two intermediate tensors. Their backpropagation equations are summarized as follows:
\begin{subequations}
\begin{align}
\frac{\partial \mathcal{L}}{\partial \tensorSup{U}{1}}  =  \frac{\partial \mathcal{L}}{\partial \tensor{V}} \times_2 (\tensorSup{K}{2})^{\top}, ~ & \frac{\partial \mathcal{L}}{\partial \tensorSup{K}{2}} = \tensorSup{U}{1} \left( \times^{0}_{0} \circ \times^{1}_{1} \right) \frac{\partial \mathcal{L}}{\partial \tensor{V}} \\
\frac{\partial \mathcal{L}}{\partial \tensorSup{U}{0}} = \frac{\partial \mathcal{L}}{\partial \tensorSup{U}{1}} \left( (\ast^{0}_{0})^{\top} \circ (\ast^{1}_{1})^{\top} \circ \times^{2}_{3} \right) \tensorSup{K}{1}, ~ & \frac{\partial \mathcal{L}}{\partial \tensorSup{K}{1}} = \frac{\partial \mathcal{L}}{\partial \tensorSup{U}{1}} \left( \overline{(\overline{\ast}^{0}_{0})^{\top}} \circ \overline{(\overline{\ast}^{1}_{1})^{\top}} \right) \tensorSup{U}{1} \\
\frac{\partial \mathcal{L}}{\partial \tensor{U}} = \frac{\partial \mathcal{L}}{\partial \tensorSup{U}{0}} \times_2 (\tensorSup{K}{0})^{\top}, ~ & \frac{\partial \mathcal{L}}{\partial \tensorSup{K}{0}} = \tensor{U} \left( \times^{0}_{0} \circ \times^{1}_{1} \right) \frac{\partial \mathcal{L}}{\partial \tensorSup{U}{0}}
\end{align}
\end{subequations}

\paragraph{\TT-convolutional layer}
\TT-convolutional layer is derived by factorizing $\tensor{K}$ using \TTlong decomposition.
\begin{equation}
\tensor{K} = \swapaxes \left( \tensorSup{K}{0} \times^{-1}_{0} \tensorSup{K}{1} \times^{-1}_{0} \tensorSup{K}{2} \times^{-1}_0 \tensorSup{K}{3} \right)
\label{def:convolutional-tt-2}
\end{equation}
where $\tensorSup{K}{0} \in \R^{S \times R_s}$, $\tensorSup{K}{1} \in \R^{R_s \times H \times R}$, $\tensorSup{K}{2} \in \R^{R \times W \times R_t}$ and $\tensorSup{K}{3} \in \R^{R_t \times T}$ are four factor tensors.
The forward pass now contains four steps:
\begin{subequations}
\begin{align}
\tensorSup{U}{0} & = \tensor{U} \times_2 \tensorSup{K}{0} \\
\tensorSup{U}{1} & = \tensorSup{U}{0} ~ (\ast^{0}_{1} \circ \times^{2}_{0}) ~ \tensorSup{K}{1} \\
\tensorSup{U}{2} & = \tensorSup{U}{1} ~ (\ast^{1}_{1} \circ \times^{2}_{0}) ~ \tensorSup{K}{2} \\
\tensor{V} & = \tensorSup{U}{2} \times_2 \tensorSup{K}{3}
\end{align}
\end{subequations}
where $\tensorSup{U}{0} \in \R^{X \times Y \times R_s}$, $\tensorSup{U}{1} \in \R^{X^{\prime} \times Y \times R}$ and $\tensorSup{U}{2} \in \R^{X^{\prime} \times Y^{\prime} \times R_t}$ are three intermediate results.
Corresponding backpropagation equations for these computations are:
\begin{subequations}
\begin{align}
\frac{\partial \mathcal{L}}{\partial \tensorSup{U}{2}}  =  \frac{\partial \mathcal{L}}{\partial \tensor{V}} \times_2 (\tensorSup{K}{3})^{\top}, ~ & \frac{\partial \mathcal{L}}{\partial \tensorSup{K}{3}} = \tensorSup{U}{2} \left( \times^{0}_{0} \circ \times^{1}_{1} \right) \frac{\partial \mathcal{L}}{\partial \tensor{V}} \\
\frac{\partial \mathcal{L}}{\partial \tensorSup{U}{1}} = \frac{\partial \mathcal{L}}{\partial \tensorSup{U}{2}} \left( (\ast^{1}_{0})^{\top} \circ \times^{2}_{2} \right) \tensorSup{K}{2}, ~ & \frac{\partial \mathcal{L}}{\partial \tensorSup{K}{2}} = \tensorSup{U}{1} \left( \times^{0}_{0} \circ \overline{(\overline{\ast}^{1}_{1})^{\top}} \right)\frac{\partial \mathcal{L}}{\partial \tensorSup{U}{2}} \\
\frac{\partial \mathcal{L}}{\partial \tensorSup{U}{0}} = \frac{\partial \mathcal{L}}{\partial \tensorSup{U}{1}}\left( (\ast^{0}_{0})^{\top} \circ \times^{2}_{2} \right) \tensorSup{K}{1}, ~ & \frac{\partial \mathcal{L}}{\partial \tensorSup{K}{2}} = \tensorSup{U}{0} \left(\overline{(\overline{\ast}^{0}_{0})^{\top}} \circ \times^{1}_{1} \right) \frac{\partial \mathcal{L}}{\partial \tensorSup{U}{1}} \\
\frac{\partial \mathcal{L}}{\partial \tensor{U}} = \frac{\partial \mathcal{L}}{\partial \tensorSup{U}{0}} \times_2 (\tensorSup{K}{0})^{\top}, ~ & \frac{\partial \mathcal{L}}{\partial \tensorSup{K}{0}} = \tensor{U} \left( \times^{0}_{0} \circ \times^{1}_{1} \right) \frac{\partial \mathcal{L}}{\partial \tensorSup{U}{0}}
\end{align}
\end{subequations}


\begin{table*}[!htbp]
\centering
\begin{tabular}{c | c | c }
Architect. & \begin{tabular}{c} $O$(\# of params.) \\ \textcolor{blue}{$O$(\# of forward ops.)} \end{tabular} & \begin{tabular}{c} $O$(\# of backward ops. for inputs) \\  \textcolor{blue}{$O$(\# of backward ops. for params.)} \end{tabular} \\ 
\hline
original & 
\begin{tabular}{c} $HWST$ \\ \textcolor{blue}{$HWSTXY$} \end{tabular} & 
\begin{tabular}{c} $HWSTX^{\prime}Y^{\prime}$ \\ \textcolor{blue}{$XYSTX^{\prime}Y^{\prime}$} \end{tabular} \\ 
\hline
\SVD & 
\begin{tabular}{c} $(HS + WT)R$ \\ \textcolor{blue}{$HSXY+WTX^{\prime}Y) R$ }\end{tabular} &
\begin{tabular}{c} $(HSX^{\prime}Y+ WTX^{\prime}Y^{\prime})R$ \\ \textcolor{blue}{$(XSX^{\prime}Y + YTX^{\prime}Y^{\prime})R$} \end{tabular} \\
\hline
\nCP & 
\begin{tabular}{c} $(HW + S + T)R$ \\ \textcolor{blue}{$(SXY + HWXY + TX^{\prime}Y^{\prime}) R$} \end{tabular} & 
\begin{tabular}{c} $(SXY + HWX^{\prime}Y^{\prime} + TX^{\prime}Y^{\prime})R$ \\ \textcolor{blue}{$(SXY + XYX^{\prime}Y^{\prime} + TX^{\prime}Y^{\prime}) R$} \end{tabular} \\
\hline
\nTK & 
\begin{tabular}{c} $(HWR_sR_t + $ \\ $SR_s + R_tT)$ \\ \textcolor{blue}{$(HWR_sR_tXY + $} \\ \textcolor{blue}{$ SR_sXY + R_tTX^{\prime}Y^{\prime})$} \end{tabular} & 
\begin{tabular}{c} $(HWR_sR_tX^{\prime}Y^{\prime} + $ \\ $SR_sXY + R_tTX^{\prime}Y^{\prime})$ \\ \textcolor{blue}{$(XYR_sR_tX^{\prime}Y^{\prime} +$} \\ \textcolor{blue}{$SR_sXY + R_tTX^{\prime}Y^{\prime})$} \end{tabular} \\
\hline
\nTT & \begin{tabular}{c} $(S R_s + H R_s R + $ \\ $ W R_t R + R_t T)$ \\ \textcolor{blue}{$(SR_sXY + HR_sRXY + $} \\ \textcolor{blue}{$WR_tRX^{\prime}Y + R_t TX^{\prime}Y^{\prime} )$} \end{tabular} & 
\begin{tabular}{c}  $(SR_sXY + HR_sRX^{\prime}Y +$ \\ $WR_tTX^{\prime}Y^{\prime} + R_t TX^{\prime}Y^{\prime})$ \\ \textcolor{blue}{$(SR_sXY + XR_sRXY +$} \\ \textcolor{blue}{$YR_tRX^{\prime}Y^{\prime} + R_tTX^{\prime}Y^{\prime})$} \end{tabular}\\
\end{tabular}
\caption{\textbf{Summary of low-rank approximations of convolutional layer.} We list the number of parameters and the number of operations required by forward/backward passes for various low-rank arppoximations that compress convolutional layer. For reference, a standard convolutional layer maps a set of $S$ feature maps with height $X$ and width $Y$, to another set of $T$ feature maps with height $X^{\prime}$ and width $Y^{\prime}$.  All filters in the convolutional layer share the same height $H$ and width $W$.}
\label{table:convolutional}
\end{table*}

\section{TNN based compression on dense layer}
\label{app:dense-tensorized}

When the same technique in Appendix~\ref{app:convolutional} is applied to compress the dense layer (a.k.a fully connected layer), we encounter one difficulty: the kernel in dense layer is a matrix and can not be further factorized by any tensor decomposition. In order to address this difficult, we define an equivalent {\em tensorized dense layer} that maps an $m$-order input tensor $\tensor{U} \in \R^{S_0 \times \cdots \times S_{m-1}}$ to another $m$-order tensor $\tensor{V} \in \R^{T_0 \times \cdots \times T_{m-1}}$ with a $2m$-order kernel $\tensor{K} \in \R^{S_0 \times \cdots S_{m-1} \times T_0 \times \cdots \times T_{m-1}}$. 
\begin{equation}
\tensor{V} = \tensor{U} \left( \times^{0}_{0} \circ \cdots \circ \times^{m-1}_{m-1} \right) \tensor{K} 
\label{def:dense-tensorized-2}
\end{equation}
The layer is equivalent to a standard dense layer if input $\tensor{U}$, kernel $\tensor{K}$ and output $\tensor{V}$ are reshaped versions of their counterparts. Now we are ready to derive compact architectures for compression using general tensor decompositions. 

\paragraph{\rCP-dense layer}
The layer is derived if the kernel $\tensor{K}$ is factorized by a modified \CP decomposition.
\begin{equation}
\tensor{K} = \swapaxes \left( \myvector{1} \times^{0}_{0} (\tensorSup{K}{0} \otimes^{0}_{0} \cdots \otimes^{0}_{0} \tensorSup{K}{m-1} \right)
\label{def:dense-cp-2}
\end{equation}
where $\tensorSup{K}{l} \in \R^{R \times S_l \times T_l}, \forall l \in [m]$ are $m$ factors. It results in a multi-steps procedure for forward pass: 
\begin{subequations}
\begin{align}
\tensorSup{U}{0} & = \myvector{1} \otimes \tensor{U} \\
\tensorSup{U}{l+1} & = \tensorSup{U}{l} \left( \otimes_{0}^{0} \circ \times^{1}_{1} \right) \tensorSup{K}{l} \\
\tensor{V} & = \myvector{1} \times^{0}_{0} \tensorSup{U}{m}
\end{align}
\end{subequations}
where the $m$ intermediate results $\tensorSup{U}{l} \in \R^{R \times S_{l} \times \cdots S_{m-1} \times T_0 \times \cdots \times T_{l-1}}, \forall l \in [m]$ are all $(m+1)$-order tensors. Correspondingly, their backpropagation equations are computed as
\begin{subequations}
\begin{align}
\frac{\partial \mathcal{L}}{\partial \tensorSup{U}{l}} & = \tensorSup{K}{l} \left( \otimes^{0}_{0} \circ \times^{2}_{-1} \right) \frac{\partial \mathcal{L}}{\partial \tensorSup{U}{l+1}} \\
\frac{\partial \mathcal{L}}{\partial \tensorSup{K}{l}} & = \tensorSup{U}{l} \left( \otimes^{0}_{0} \circ \times^{2}_{1} \circ \cdots \circ \times^{m}_{m-1} \right) \frac{\partial \mathcal{L}}{\partial \tensorSup{U}{l+1}} \\ 
\frac{\partial \mathcal{L}}{\partial \tensorSup{U}{m}} & = \myvector{1} \otimes \frac{\partial \mathcal{L}}{\partial \tensor{V}}, ~ \frac{\partial \mathcal{L}}{\partial \tensor{U}} = \myvector{1} \times^{0}_{0} \frac{\partial \mathcal{L}}{\partial \tensorSup{U}{0}} \nonumber
\end{align}
\end{subequations}

\paragraph{\rTK-dense layer} The layer is obtained when a modified \TKlong decomposition is used.
\begin{equation}
\tensor{K} = \tensor{C} \left( \times_0 (\matrixSup{P}{0})^{\top} \cdots \times_{m-1} (\matrixSup{P}{m-1})^{\top} \times_{m} \matrixSup{Q}{0} \cdots \times_{2m - 1} \matrixSup{Q}{m-1} \right) 
\label{def:dense-tensorized-tk-2}
\end{equation}
where $\matrixSup{P}{l} \in \R^{S_l \times R^s_l}, \forall l \in [m]$ are named as input factors, $\tensor{C} \in \R^{R^s_0 \times \cdots \times R^s_{m-1} \times R^t_0 \times \cdots \times R^t_{m-1}}$ as core factor, and lastly $\matrixSup{Q}{l} \in \R^{R^t_l \times T_l}, \forall l \in [m]$ as output factors.
The forward pass of \rTK-dense layer can then be evaluated in three steps
\begin{subequations}
\begin{align}
\tensorSup{U}{0} & = \tensor{U} \left( \times_0 \matrixSup{P}{0} \cdots \times_{m-1} \matrixSup{P}{m-1} \right) \\
\tensorSup{U}{1} & = \tensorSup{U}{0} \left( \times_{0}^{0} \circ \cdots \circ \times_{m-1}^{m-1} \right) \tensor{C} \\
\tensor{V} & = \tensorSup{U}{1} \left( \times_0 \matrixSup{Q}{0} \cdots \times_{m-1} \matrixSup{Q}{m-1} \right) 
\end{align}
\end{subequations}
where $\tensorSup{U}{0}$ and $\tensorSup{U}{1}$ are two intermediate results. 
The backpropagation equations can then be derived accordingly:
\begin{subequations}
\begin{gather}
\frac{\partial \mathcal{L}}{\partial \tensorSup{U}{1}} = \frac{\partial \mathcal{L}}{\partial \tensor{V}} \left( \times_{0} (\matrixSup{Q}{0})^{\top} \cdots \times_{m-1} (\matrixSup{Q}{m-1})^{\top} \right) \\
\left( \frac{\partial \mathcal{L}}{\partial \matrixSup{Q}{l}} \right)^{\top} = \frac{\partial \mathcal{L}}{\partial \tensor{V}} \left( \times^{0}_{0} \circ \cdots \circ \times^{l-1}_{l-1} \circ \times^{l+1}_{l+1} \circ \cdots \circ \times^{m-1}_{m-1} \right) \nonumber \\
\left( \tensorSup{U}{1} \left( \times_{0} \matrixSup{Q}{0} \cdots \times_{l-1} \matrixSup{Q}{l-1} \times_{l+1} \matrixSup{Q}{l+1} \cdots \times_{m-1} \matrixSup{Q}{m-1} \right) \right) \\
\frac{\partial \mathcal{L}}{\partial \tensor{U}} = \frac{\partial \mathcal{L}}{\partial \tensorSup{U}{0}} \left( \times_{0} (\matrixSup{P}{0})^{\top} \cdots \times_{m-1} (\matrixSup{P}{m-1})^{\top} \right) \\
\frac{\partial \mathcal{L}}{\partial \tensorSup{U}{0}} = \tensor{C} \left( \times^{m}_{0} \circ \cdots \circ \times^{2m - 1}_{m - 1} \right) \frac{\partial \mathcal{L}}{\partial \tensorSup{U}{1}} \\
\frac{\partial \mathcal{L}}{\partial \tensor{C}} = \tensorSup{U}{0} \otimes \frac{\partial \mathcal{L}}{\partial \tensorSup{U}{1}} \\
\left( \frac{\partial \mathcal{L}}{\partial \matrixSup{P}{l}} \right)^{\top} = \frac{\partial \mathcal{L}}{\partial \tensorSup{U}{0}} \left( \times^{0}_{0} \circ \cdots \circ \times^{l-1}_{l-1} \circ \times^{l+1}_{l+1} \circ \cdots \circ \times^{m-1}_{m-1} \right) \nonumber \\
\left( \tensor{U} \left( \times_{0} \matrixSup{P}{0} \cdots \times_{l-1} \matrixSup{P}{l-1} \times_{l+1} \matrixSup{P}{l+1} \cdots \times_{m-1} \matrixSup{P}{m-1} \right) \right)
\end{gather}
\end{subequations}

\paragraph{\rTT-dense layer}
The layer~\cite{novikov2015tensorizing} is derived by factorizing $\tensor{K}$ using a modified \TTlong decomposition. 
\begin{equation}
\tensor{K} = \swapaxes \left( \tensorSup{K}{0} \times^{-1}_{0} \tensorSup{K}{1} \times^{-1}_{0} \cdots \times^{-1}_{0} \tensorSup{K}{m-1} \right) \label{def:dense-tensorized-tt-2}
\end{equation}
where the factor tensors are $\tensorSup{K}{l} \in \R^{R_{l-1} \times S_l \times T_l \times R_l},\forall l = 1, \cdots, m-2$, with two corner cases $\tensorSup{K}{0} \in \R^{S_0 \times T_0 \times R_0}$ and $\tensorSup{K}{m-1} \in \R^{R_{m-2} \times S_{m-1} \times T_{m-1}}$.
\beq
\tensorSup{U}{l+1} = \tensorSup{U}{l} \left( \times^{0}_{1} \circ \times^{-1}_{0} \right) \tensorSup{K}{l}
\eeq
where $\tensorSup{U}{0} = \tensor{U}$,  $\tensorSup{U}{m} = \tensor{V}$, and the other $m$ tensors $\tensorSup{U}{l}$'s are intermediate results. Following Appendix~\ref{app:derivatives}, its backpropagation equations can be derived as:
\begin{subequations}
\begin{align}
\frac{\partial \mathcal{L}}{\partial \tensorSup{U}{l}} & = \tensorSup{K}{l} \left( \times^{1}_{-2} \circ \times^{2}_{-1} \right) \frac{\partial \mathcal{L}}{\partial \tensorSup{U}{l+1}} \\
\frac{\partial \mathcal{L}}{\partial \tensorSup{K}{l}} & = \swapaxes \left( \tensorSup{U}{l} \left( \times^{1}_{0} \circ \cdots \circ \times^{m-1}_{m-2} \right) \frac{\partial \mathcal{L}}{\partial \tensorSup{U}{l+1}} \right)
\end{align}
\end{subequations}

 \begin{table*}[!htbp]
\centering
\begin{tabular}{ c | c | c }
Architect. & $O$(\# of params.) & $O$(\# of forward/backprop ops.) \\ 
\hline
original & $ST$ & $ST$ \\ 
SVD & $(S + T) R$ & $(S + T) R$ \\ 
\hline
\rCP & $m (ST)^{\frac{1}{m}} R$ & 
$m \max(S, T)^{1 + \frac{1}{m}} R$ \\
\rTK & $m (S^{\frac{1}{m}} + T^{\frac{1}{m}}) R + R^{2m}$ & 
$m(S + T)R + R^{2m}$ \\
\rTT & $m (ST)^{\frac{1}{m}} R^2$ & 
$m \max(S, T)^{1 + \frac{1}{m}} R^2$ 
\end{tabular}
\caption{\textbf{Summary of TNN compression on dense layer.} In this table, we list the numbers of parameters and time complexities of forward/backward passes required by various TNN architectures that compress dense layer. For simplicity, we assume that the number of input units $S$ and outputs units $T$ are factorized evenly, i.e. $S_l = S^{\frac{1}{m}}, T_l = T^{\frac{1}{m}}, \forall l \in [m]$ and all ranks (in \rTK and \rTT) share the same number $R$, i.e. $R_l = R, \forall l \in [m]$. }
\label{table:dense-tensorized}
\end{table*}


\section{TNN compression on convolutional layer}
\label{app:convolutional-tensorized}




Inspired by the reshaping trick in Appendix~\ref{app:dense-tensorized}, we develop three novel architectures that compress convolution layer by first proposing its higher-order counterpart: a {\em tensorized convolutional layer} maps $(m+2)$-order tensor $\tensor{U} \in \R^{X \times Y \times S_0 \times \cdots \times S_{m-1}}$ to another $(m+2)$-order tensor $\tensor{V} \in \R^{X^{\prime} \times Y^{\prime} \times T_0 \times \cdots \times T_{m-1}}$ with a $(2m + 2)$-order kernel $\tensor{K} \in \R^{H \times W \times S_0 \times \cdots \times S_{m-1} \times T_0 \times  \cdots \times T_{m-1}}$.
\begin{equation}
\tensor{V} = \tensor{U} \left( \ast_{0}^{0} \circ \ast_{1}^{1} \circ \times_{2}^{2} \circ \cdots \circ \times^{m+1}_{m+1} \right) \tensor{K} 
\label{def:convolutional-tensorized-2}
\end{equation}
The tensorized convolutional layer is equivalent to standard convolutional layer if input $\tensor{U}$, kernel $\tensor{K}$ and output $\tensor{V}$ are reshaped from their counterparts.. 


\paragraph{\rCP-convolutional layer}
The layer is derived if the kernel $\tensor{K}$ is factorized by a modified CP decomposition as in Figure~\ref{fig:decomposition-rcp}. 
\begin{equation}
\tensor{K} = \swapaxes \left( \myvector{1} \times^{0}_{0} \left( \tensorSup{K}{0} \otimes^{0}_{0} \cdots \otimes^{0}_{0} \tensorSup{K}{m} \right) \right) \label{def:convolutional-tensorized-cp-2}
\end{equation}
where $\tensorSup{K}{l}  \in \R^{R \times S_l \times T_l}, \forall l \in [m]$ and $\tensorSup{K}{m} \in \R^{R \times H \times W}$ are $(m + 1)$ factors.
Accordingly, the multi-steps procedure to evaluate the output $\tensor{V}$ now has $(m + 2)$-steps:
\begin{subequations}
\begin{align}
\tensorSup{U}{0} & = \myvector{1} \otimes \tensor{U} \\
\tensorSup{U}{l+1} & = \tensorSup{U}{l} \left( \otimes_{0}^{0} \circ \times^{3}_{1} \right) \tensorSup{K}{l} \\
\tensor{V} & = \tensorSup{U}{m} \left( \times_{0}^{0} \circ \ast^{1}_{1} \circ \ast^{2}_{2} \right) \tensorSup{K}{m}
\end{align}
\end{subequations}
where $\tensorSup{U}{l} \in \R^{R \times S_{l} \times \cdots \times S_{m+1} \times T_{0} \times \cdots \times T_{l-1}}, \forall l \in [m]$ are $m$ intermediate tensors. 
The backpropagation equations of these steps above are
\begin{subequations}
\begin{align}
\frac{\partial \mathcal{L}}{\partial \tensorSup{U}{m}} & = \tensorSup{K}{m} \left( (\ast^{1}_{0})^{\top} \circ (\ast^{2}_{1})^{\top} \right) \frac{\partial \mathcal{L}}{\partial \tensor{V}} \\ 
\frac{\partial \mathcal{L}}{\partial \tensorSup{K}{m}} & = \tensorSup{U}{m} \left( (\ast^{1}_{0})^{\top} \circ (\ast^{2}_{1})^{\top} \times^{3}_{2} \circ \cdots \circ \times^{m+2}_{m+1} \right) \frac{\partial \mathcal{L}}{\partial \tensor{V}} \\
\frac{\partial \mathcal{L}}{\partial \tensorSup{U}{l}} & = \swapaxes \left( \tensorSup{K}{l} \left( \otimes^{0}_{0} \circ \times^{2}_{-1} \right) \frac{\partial \mathcal{L}}{\partial \tensorSup{U}{l+1}} \right) \\
\frac{\partial \mathcal{L}}{\partial \tensorSup{K}{l}} & = \tensorSup{U}{l} \left( \otimes^{0}_{0} \circ \otimes^{1}_{1} \circ \otimes^{2}_{2} \circ \times^{4}_{3} \circ \cdots \circ \times^{m+3}_{m+2} \right) \frac{\partial \mathcal{L}}{\partial \tensorSup{U}{l+1}} \\
\frac{\partial \mathcal{L}}{\partial \tensor{U}} & = \myvector{1} \times^{0}_{0} \frac{\partial \mathcal{L}}{\partial \tensorSup{U}{0}}
\end{align}
\end{subequations}

\paragraph{\rTK-convolutional layer} 
The layer is derived if a modified TK decomposition as in Figure~\ref{fig:decomposition-rtk} on kernel $\tensor{K}$. 
\begin{equation}
\tensor{K} = \tensor{C} \left( \times_2 (\matrixSup{P}{0})^\top \cdots \times_{m+1} (\matrixSup{P}{m-1})^\top \times_{m+2} \matrixSup{Q}{0} \cdots \times_{2m + 1} \matrixSup{Q}{m-1} \right) 
\label{def:convolutional-tensorized-tk-2}
\end{equation}
where $\matrixSup{P}{l} \in \R^{S_l \times R^s_l}, \forall l \in [m]$, $\tensor{C} \in \R^{H \times W \times R^s_0 \times \cdots \times R^s_{m-1} \times R^t_{0} \times \cdots \times R^t_{m-1}}$ and $\matrixSup{Q}{l} \in \R^{R^t_l \times T_l}, \forall l \in [m]$ are named as input factors, core factor and output factors respectively. The forward pass of \rTK-convolutional layer takes three steps:
\begin{subequations}
\begin{align}
\tensorSup{U}{0} & = \tensor{U} \left( \times_2 \matrixSup{P}{0} \cdots \times_{m+1} \matrixSup{P}{m-1} \right) \\
\tensorSup{U}{1} & = \tensorSup{U}{0} \left(\ast^{0}_{0} \circ \ast^{1}_{1} \times_{2}^{2} \circ \cdots \circ \times_{m-1}^{m-1} \right) \tensor{C} \\
\tensor{V} & = \tensorSup{U}{1} \left(\times_2 \matrixSup{Q}{0} \cdots \times_{m+1} \matrixSup{Q}{m-1} \right) 
\end{align}
\end{subequations}
where $\tensorSup{U}{0} \in \R^{X \times Y \times R^s_0 \times \cdots \times  R^s_{m-1}}$ and $\tensorSup{U}{1} \in \R^{X^{\prime} \times Y^{\prime} \times R^t_0 \times \cdots \times R^t_{m-1}}$ are two intermediate tensors. The backpropagation equations can be derived similarly:
\begin{subequations}
\begin{gather}
\frac{\partial \mathcal{L}}{\partial \tensorSup{U}{0}} = \tensor{C} \left( (\ast^{0}_{0})^{\top} \circ (\ast^{1}_{1})^{\top} \circ \times^{m+2}_{2} \circ \cdots \circ \times^{2m + 1}_{m + 1} \right) \frac{\partial \mathcal{L}}{\partial \tensorSup{U}{1}} \\
\frac{\partial \mathcal{L}}{\partial \tensor{C}} = \tensorSup{U}{0} \left( (\ast^{0}_{0})^{\top} \circ (\ast^{1}_{1})^{\top} \right) \frac{\partial \mathcal{L}}{\partial \tensorSup{U}{1}}
\end{gather}
\end{subequations}

\paragraph{\rTT-convolutional layer} 
The layer is derived by factoring $\tensor{K}$ according to a modified \TT decomposition as in Figure~\ref{fig:decomposition-rtt}.
\begin{equation}
\tensor{K} = \swapaxes \left( \tensorSup{K}{0} \times^{-1}_{0} \tensorSup{K}{1} \times^{-1}_{0} \cdots \times^{-1}_{0} \tensorSup{K}{m} \right) 
\label{def:convolutional-tensorized-tt-2}
\end{equation}
where $\tensorSup{K}{0} \in \R^{S_0 \times T_0 \times R_0}$, $\tensorSup{K}{l} \in \R^{R_{l-1} \times S_l \times T_l \times R_l}$ and $\tensorSup{K}{m} \in \R^{R_{m-1} \times H \times W}$ are $(m + 1)$ factor tensors. The multi-stages forward pass to evaluate $\tensor{V}$ now has $(m+1)$ steps:
\begin{subequations}
\begin{align}
\tensorSup{U}{l} & = \tensorSup{U}{l} \left( \times^{2}_{1} \circ \times^{-1}_{0} \right) \tensorSup{K}{l} \\
\tensor{V} & = \tensorSup{U}{m} \left( \ast^{0}_{1} \circ \ast^{1}_{2} \circ \times^{-1}_{0} \right) \tensorSup{K}{m}
\end{align}
\end{subequations}
where $\tensorSup{U}{l} \in \R^{X \times Y \times S_{l} \times \cdots \times S_{m-1} \times T_{0} \times \cdots \times T_{l-1} \times R_{l-1}}, \forall l \in [m]$ are the intermediate results. We summarize the corresponding backpropagation equations as follows:
\begin{subequations}
\begin{align}
\frac{\partial \mathcal{L}}{\partial \tensorSup{U}{m}} & = \frac{\partial \mathcal{L}}{\partial \tensor{V}} \left( (\ast^{0}_{1})^{\top} \circ (\ast^{2}_{1})^{\top} \right) \tensorSup{K}{m} \\
\frac{\partial \mathcal{L}}{\partial \tensorSup{K}{m}} & = \frac{\partial \mathcal{L}}{\partial \tensor{V}} \left( (\ast^{0}_{0})^{\top} \circ (\ast^{1}_{1})^{\top} \circ \times^{2}_{2} \cdots \times^{m+1}_{m+1} \right) \tensorSup{U}{m} \\
\frac{\partial \mathcal{L}}{\partial \tensorSup{U}{l}} & = \swapaxes \left( \tensorSup{K}{l} \left( \times^{2}_{-2} \circ \times^{3}_{-1} \right) \frac{\partial \mathcal{L}}{\partial \tensorSup{U}{l+1}} \right) \\
\frac{\partial \mathcal{L}}{\partial \tensorSup{K}{l}} & = \swapaxes \left( \tensorSup{U}{l} \left( \times^{0}_{0} \circ \times^{1}_{1} \circ \times^{3}_{2} \cdots \times^{m+1}_{m} \right) \frac{\partial \mathcal{L}}{\partial \tensorSup{U}{l+1}} \right) 
\end{align}
\end{subequations}


\begin{table*}[!htbp]
\centering
\begin{tabular}{l}
\begin{tabular}{  c | c | c  }
Architect. & 
\begin{tabular}{c} O(\# of params) \\ \textcolor{blue}{O(\# of forward ops.)} \end{tabular} & 
\begin{tabular}{c} O(\# of back ops. for inputs) \\ \textcolor{blue}{O(\# of back ops. for params.)} \end{tabular} \\ 
\hline
original & 
\begin{tabular}{c} $HWST$ \\ \textcolor{blue}{$HWSTXY$} \end{tabular} & 
\begin{tabular}{c} $HWSTX^{\prime}Y^{\prime}$ \\ \textcolor{blue}{$XYSTX^{\prime}Y^{\prime}$} \end{tabular} \\ 
\hline
\tnnCP & 
\begin{tabular}{c} $\mathbf{\#Params_{CP}} + HWR$ \\ \textcolor{blue}{$\mathbf{\#Ops_{CP}}XY + HWTRXY$} \end{tabular} &
\begin{tabular}{c} $\mathbf{\#Ops_{CP}}XY + HWTRX^{\prime}Y^{\prime}$ \\ \textcolor{blue}{$\mathbf{\#Ops_{CP}}XY + XYTRX^{\prime}Y^{\prime}$} \end{tabular} \\
\hline
\tnnTK & 
\begin{tabular}{c} $(HW \mathbf{\# Params_{TK_C}} + $ \\ $\mathbf{\# Params_{TK_I}} + \mathbf{\# Params_{TK_O}}$ \\ \textcolor{blue}{$(HW \mathbf{\# Ops_{TK_C}} XY + $} \\ \textcolor{blue}{$\mathbf{\# Ops_{TK_I}} XY + \mathbf{\# Ops_{TK_O}} X^{\prime}Y^{\prime})$} \end{tabular} & 
\begin{tabular}{c} $(HW \mathbf{\# Ops_{TK_C}} X^{\prime}Y^{\prime} + $ \\ $\mathbf{\# Ops_{TK_I}}XY + \mathbf{\# Ops_{TK_O}}X^{\prime}Y^{\prime})$ \\ \textcolor{blue}{$(XY\mathbf{\# Ops_{TK_C}}X^{\prime}Y^{\prime} + $} \\ \textcolor{blue}{$\mathbf{\# Ops_{TK_I}}XY + \mathbf{\# Ops_{TK_O}}X^{\prime}Y^{\prime})$} \end{tabular} \\
\hline
\tnnTT & 
\begin{tabular}{c} $\mathbf{\#Params_{TT}} + HWR$ \\ \textcolor{blue}{$\mathbf{\#Ops_{TT}}XY + HWTRXY$} \end{tabular} &
\begin{tabular}{c} $\mathbf{\#Ops_{TT}}XY + HWTRX^{\prime}Y^{\prime}$ \\ \textcolor{blue}{$\mathbf{\#Ops_{TT}}XY + XYTRX^{\prime}Y^{\prime}$} \end{tabular} \\
\end{tabular} \\
\rule{0in}{1.2em}$^\dag$\scriptsize In order to compare against \ouralgo\!s on dense layer in Table~\ref{table:dense-tensorized}, we denote the numbers for the dense layers as: \\
\scriptsize \begin{tabular}{l l l}
$\mathbf{\#Params_{CP}} = m(ST)^{\frac{1}{m}}R$ & $\mathbf{\#Params_{TK}} = \sum_{\mathbf{L}}\mathbf{\#Params_{TK_L}}$ & $\mathbf{\#Params_{TT}} = m(ST)^{\frac{1}{m}} R^2$ \\
$\mathbf{\#Ops_{CP}} = m \max(S, T)^{1 + \frac{1}{m}}$ & $\mathbf{\#Ops_{TK}} = \sum_{\mathbf{L}} \mathbf{\#Ops_{TK_L}}$ & $\mathbf{\#Ops_{TT}} = m \max(S, T)^{1 + \frac{1}{m}} R^2$  \\
$\mathbf{\# Params_{TK_I}} = mS^{\frac{1}{m}}R$ & $\mathbf{\# Params_{TK_C}} = R^{2m}$ & $\mathbf{\# Params_{TK_O}} = mT^{\frac{1}{m}}R$ \\
 $\mathbf{\#Ops_{TK_I}} = mSR$ & $\mathbf{\# Ops_{TK_C}} = R^{2m}$ & $\mathbf{\# Ops_{TK_O}} = mTR$ \\
\end{tabular}
\end{tabular}
\caption{\textbf{Summary of TNN compression on convolutional layer.} In this table, we list the number of parameters and time complexities required by various TNN architectures that compress convolutional layer. Recall that a convolutional layer, composed with $ST$ filters of size $H \times W$, maps a set of $S$ feature maps of size $X \times Y$ to another set of $T$ feature maps of size $X^{\prime} \times Y^{\prime}$. For simplicity, we assume the numbers of input/output feature maps $S$, $T$ are factorized evenly, i.e. $S_l = S^{\frac{1}{m}}, T_l = T^{\frac{1}{m}}, \forall l \in [m]$, and all ranks are equal to $R$, i.e $R_l = R, \forall l \in [m]$. } 
\label{table:convolutional-tensorized}
\end{table*}

\end{document}